Federal Office
for Information Security

# Facial Metrics for EES (EESFM)

State of the Art of Quality Assessment of Facial Images

Version 1.2
11.11.2022

Johannes Merkle, Christian Rathgeb, Benjamin Tams, Dhay-Parn Lou, André Dörsch, Pawel Drozdowski

secunet Security Networks AG

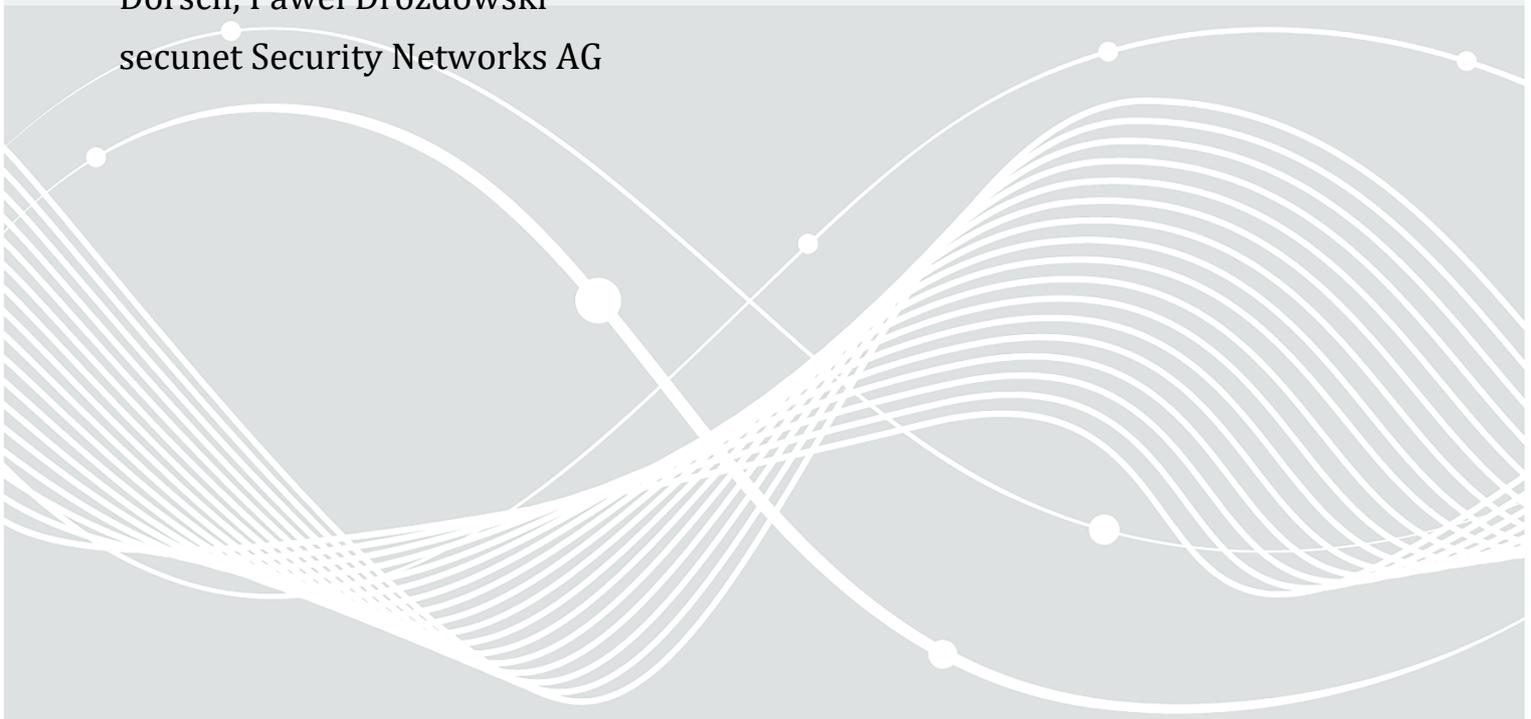

# Document history

| Version | Date | Editor | Description |
|---------|------|--------|-------------|
| 0.1 | 25.02.2022 | D.-P. Lou, J. Merkle, C. Rathgeb | "Expression Neutrality" and "No Occlusion of Face" added. |
| 0.2 | 04.04.2022 | P. Drozdowski, D.-P. Lou | "Unified Quality Score" added. |
| 0.3 | 11.04.2022 | D.-P Lou, J. Merkle, C. Rathgeb | "Face detection, segmentation, landmark localization and alignment", "Inter-eye Distance", "Pose", "No Reflections on Eyeglasses" and Glossary added. |
| 0.4 | 26.04.2022 | D.-P Lou, C. Rathgeb | "De-Focus, Sharpness, Motion blur, and Edge Density" added. |
| 0.5 | 20.05.2022 | A. Dörsch, D.-P Lou, J. Merkle, C. Rathgeb | "Introduction", "Unnatural Colour and Colour Imbalance", "Compression", "Uniform Background" added |
| 0.6 | 23.05.2022 | A. Dörsch, D.-P Lou, J. Merkle, B. Tams | "Mouth Closed", "Camera Subject Distance", "Camera lens Focal Length", "No other faces in the background", "Eyes open", "Commercial Algorithms" added |
| 0.7 | 24.05.2022 | D.-P Lou, B. Tams | Finalization of draft for internal discussion |
| 0.8 | 08.06.2022 | J. Merkle | "No head Coverings" added, converted to BSI document format, minor corrections |
| 0.9 | 04.07.2022 | A. Dörsch, D.-P Lou, J. Merkle, C. Rathgeb | Revision according to comments from BSI and C. Busch, revision of listed Repositories, amendment of algorithms for "Eyes Open", general quality assurance |
| 1.0 | 05.07.2022 | J. Merkle | Correction of headers and footers, modification of contact information |
| 1.1 | 09.08.2022 | J. Merkle | Specification of some file sizes of CNN models |
| 1.2 | 11.11.2022 | J. Merkle | Amendment of data sets for landmarks localization and occlusion, description of approached for face de-occlusion added, citation numbers corrected |

*Table 1: Document history*

# Acknowledgement

This report has been created by secunet Security Networks as part of the Project P 527 - Facial Metrics for EES (EESFM) of the Federal Office for Information Security.

The authors would like to thank Christoph Busch for his valuable comments.

# Table of Contents

# List of Figures

# List of Tables

# 1 Introduction

The goal of the project "Facial Metrics for EES" is to develop, implement and publish an open source algorithm for the quality assessment (QA) of facial images for face recognition, in particular for border control scenarios.[1] In order to stimulate the harmonization of the requirements and practices applied for QA for facial images, the insights gained and algorithms developed in the project will be contributed to the current (2022) revision of the ISO/IEC 29794-5 standard; at the time of writing, the current draft of this revision is Working Draft (WD) 4, henceforth referred to as ISO/IEC WD4 29794-5:2022. Furthermore, the implemented quality metrics and algorithms will consider the recommendations and requirements from other relevant standards, in particular ISO/IEC 19794-5:2011, ISO/IEC 29794-5:2010, ISO/IEC 39794-5:2019 and Version 5.2 of the BSI Technical Guideline TR-03121 Part 3 Volume 1, henceforth referred to as TR-03121-3.1.

## 1.1 Objective and Approach

In order to establish an informed basis for the selection of quality metrics and the development of corresponding quality assessment algorithms, the state of the art of methods and algorithms (defining a metric), implementations and datasets for quality assessment for facial images is surveyed. For all relevant quality aspects, this document summarizes the requirements of the aforementioned standards, known results on their impact on face recognition performance, publicly available datasets, proposed methods and algorithms and open-source software implementations.

For some quality criteria, a large amount of research and open-source implementations has been published (e.g. for expression neutrality, pose, occlusion of mouth and nose). In these cases, only the algorithms and implementations have been documented for which one of the following conditions is satisfied:

- The algorithm has been proposed in a publication either published in a top-rank conference or journal in the field of computer vision or cited by a large number of scientific papers. A ranking of computer vision conferences can, for instance, be found on Google Scholar[2].
- The algorithm or implementation is particularly promising with respect to accuracy, either for classification / regression with respect to the relevant quality criterion or for the prediction of biometric recognition performance, or with respect to computational performance. Note that for some of the implementations listed in this survey, no corresponding scientific publications exist.
- Further, since a continuous improvement of algorithms for facial image quality assessment is observable over time, we pay particular attention to the latest developments in this field of research.

Generally, only software implementations that are completely public, including the trained machine learning (e.g. Convolutional Neural Networks) models, have been considered.

---

[1] Face recognition in border control scenarios does not only comprise the comparison of the live image with the reference image but also supporting algorithms for attack detection, in particular morphing attack detection (MAD) and presentation attack detection (PAD).

[2] https://scholar.google.com/citations?view_op=top_venues&hl=en&vq=eng_computervisionpatte-recognition

Numbers of classification accuracy of algorithms and software implementations have been taken from publicly available sources (e.g. scientific publications or web sites) The assessment of computational performance (processing time and demand for computer resources) have been assessed based on published results and the size of the files containing the trained machine learning models. Note, however, that the file size of convolutional neural networks (CNNs) may heavily vary depending on whether the whole training checkpoint had been saved or just the model weights; thus, the file size is only a rough indicator of the resources for deployment.[3]

Furthermore, for each data set and each implementation, the applicable license is specified;[4] in cases, where no license has been specified, "no license" is sepcified.

## 1.2    Requirements from EES Implementing Decision

The Commission Implementing Decision (EU) 2019/329 lays down the specifications for the quality of facial images for biometric verification and identification in the Entry/Exit System (EES). The annex 1.1.2 "Facial images" requires that the quality of the facial images shall comply with the image requirements of ISO/IEC 19794-5:2011 frontal image type which can be found in the clause 7 "The Frontal Face Image Type".

The "Part 5: Face Image Data" of the ISO/IEC 19794-5:2011 standard specifies face image quality criteria intended to improve face recognition accuracy, in particular – scene constraints (lighting, pose, expression, etc.), photographic properties (positioning, camera focus, etc.) and digital image attributes (image resolution, image size, etc.). Requirements related to Frontal Face image type are given in clause 7 of the standard, the corresponding best practice recommendations can be found in Annex B "Best practices for Face Images".

The other relevant standard is the ISO/IEC 39794-5:2019 "Face Image Data". It provides a generic face image data format for face recognition applications and, in addition to that, it specifies application specific profiles including scene constraints, photographic properties and digital image attributes like image spatial sampling rate, image size, etc. In the context of this report, the relevant application specific profile is the MRTD profile in Annex D.1. The current Edition 8 of the "ICAO Doc 9303, Machine Readable Travel Documents, Part 9" (from 2021) refers to the quality requirements for facial images of Annex D1 of ISO/IEC 39794-5:2019.

The most current of all relevant standards is ISO/IEC WD4 29794-5:2022 "Information technology – Biometric sample quality – Part 5: Face image data". This working draft references the face image requirements specified in ISO/IEC 39794-5:2019 and establishes requirements on implementations that quantify how face image's properties deviate from those specified in ISO/IEC 39794-5.

The implementing decision for Entry/Exit Systems requires that the quality of facial images comply with the ISO/IEC 19794-5:2011, therefore we see those requirements as most important among requirements of all relevant standards and show in the following table in which sections of this report the corresponding requirements are elaborated.

---

[3] In some cases, which are marked by a reference to this footnote, we have determined the actual size of the model contained in the checkpoint file.

[4] For some data sets and software implementations, an individual permission to use it in OFIQ was granted by its owners. These cases are indicated by a reference to this footnote.

| Requirement of ISO/IEC 19794-5:2011 | Subclause | Reference to sections in current report |
|---|---|---|
| Spatial Sampling Rate | 5.7.6, 8.4.1 | 12 Inter-Eye Distance |
| Image Data Encoding Requirements<br>Image Data Compression Requirements | 6.2<br>6.3 | 7 Compression |
| Pose | 7.2.2 | 14 Pose<br>13 Location and Coverage of the Head |
| Expression | 7.2.3 | 15 Expression Neutrality<br>16 Mouth Closed |
| Assistance in Positioning the Face | 7.2.4 | 19 No Other Faces in the Background |
| Shoulders | 7.2.5 | 13 Pose |
| Backgrounds | 7.2.6 | 4 Background Uniformity |
| Subject and Scene Lighting | 7.2.7 | 4 Illumination |
| Hot Spots and Specular Reflections | 7.2.8 | 4 Illumination<br>18 No Reflections on Eyeglasses |
| Eye Glasses | 7.2.9 | 11 No Occlusion of the Face |
| Head Coverings | 7.2.10 | 20 No Head Coverings |
| Visibility of Pupils and Irises | 7.2.11 | 17 Eyes Open<br>11 No Occlusion of the Face |
| Lighting Artefacts | 7.2.12 | 18 No Reflections on Eyeglasses |
| Eye Patches | 7.2.13 | 11 No Occlusion of the Face |
| Contrast and Saturation | 7.3.2 | 4 Illumination |
| Focus and Depth of Field | 7.3.3 | 6 Image Sharpness Aspects |
| Unnatural Colour | 7.3.4 | 8 Unnatural Colour and Colour Balance |
| Colour or Greyscale Enhancement | 7.3.5 | 4 Illumination |
| Radial Distortion of the Camera Lens | 7.3.6 | 9 Camera Subject Distance<br>9 Camera Lens Focal Length |
| Colour Profile | 7.4.2 | 7 Compression |
| Pixel Aspect Ratio | 7.4.1.1 | Out of scope, as it cannot be checked |
| Head entirely visible in the image<br>Horizontal Position of Face<br>Vertical Position of Face | 8.3.1<br>8.3.2<br>8.3.3 | 13 Location and Coverage of the Head |
| Width of Head<br>Length of Head | 8.3.4<br>8.3.5 | 12 Inter-Eye Distance<br>13 Location and Coverage of the Head |
| Post-Acquisition Processing | 8.4.2, 9.2.6 | Out of scope, as it cannot be checked |

| Requirement of ISO/IEC 19794-5:2011 | Subclause | Reference to sections in current report |
|---|---|---|
| Geometric Characteristics<br>Minimum width of the Token Image Type | 9.2.3<br>9.2.4 | 12 Inter-Eye Distance<br>13 Location and Coverage of the Head |
| Padding | 9.2.5 | Out of scope, as it cannot be checked |

*Table 1. Requirements of ISO/IEC 19794-5:2011 and corresponding chapters in current report*

# 2 Face Detection, Segmentation, Landmark Localization and Alignment

Many quality criteria refer to properties of the face (e.g. pose, expression, occlusion) or the face area in the image (e.g. illumination, sharpness) and hence, require the detection (localization) and segmentation of the face(s) in the image or the localization of certain parts of the face. Some algorithms (in particular CNNs) even require that, for all images, the facial features (eyes, mouth, nose) are roughly located at the same position. For these tasks, face detection, face segmentation, facial landmark localization, and face alignment is used.

- *Face detection* refers the localization of all faces visible in the image, and is typically done by determining (ideally minimal) bounding boxes for each of the detected faces. This step is a pre-requisite for facial landmark localization, face segmentation and face alignment, and is, thus, required for the assessment of most quality criteria. A vast number of algorithms and implementations have been proposed for face detection, the state of the art of which is described in detail in Section 2.1.
- *Face segmentation* means splitting the image into a foreground, covering only the area of the face, and the background. This can be accomplished coarsely by cropping to the bounding box of the face, more concise by computing facial landmarks of the face and using the convex hull of the landmarks as foreground, or by using software for the direct segmentation of the face (potentially among other parts of the subject like hair, neck, etc.). The state of the art for the latter approach (direct segmentation of the face) is described in Section 2.2.
- *Facial landmarks* are supposed to specify the location of pre-defined feature points like corners or boundaries of the eyes, mouth, nose, eye brows, face boundary. Several (indexed) sets of feature points can be defined, depending on the demands of the applications (see Figure 3). A vast number of algorithms and implementations have been proposed for facial landmark localization, the state of the art of which is described in detail in Section 2.3. Note, that in many publications, the localization of facial landmarks is denoted as face alignment [1]. However, in the context of this document, the term face alignment exclusively denotes the pre-processing described in the following paragraph.
- In this document, *face alignment* refers to a pre-processing step of CNNs processing facial images, where an affine transformation (translation, rotation, scaling) and cropping is applied to all input images to ensure that they have the required dimension and that eyes, nose and mouth are always located roughly at the same position. The dimension of the input image must fit to the dimension of the input layer of the CNN, while the uniform positioning of the eyes and mouth in the input images (both in the training set and for prediction) can considerably improve the accuracy.
  Using facial landmarks, the face alignment is quite straightforward: First, an affine transformation of the image is computed minimizing the deviation of resulting positions of eyes, mouth and nose from the target positions, then this transformation is applied, and finally, the image is cropped to the required dimensions. Therefore, face alignment is not further discussed in this document.
  Note that, according to ISO/IEC 19794-5:2011, besides compression, the processing steps of face alignment (in-plane rotation, cropping, scaling) are the only post-processing operations allowed after the capture of fully frontal face images.

## 2.1  Face Detection

### 2.1.1  Datasets

The following table summarizes several popular datasets containing face images and labels (coordinates of the bounding boxes) for the position of faces within the images.

| Name | Subjects | Images | Constrained? | License |
|------|----------|--------|--------------|---------|
| Wider-Face[5] | 393,703 | 32,203 | No | No license |
| AFW[6] | 437 | 205 | No | No license |
| FDDB[7] | 5,171 | 2,845 | No | No license |
| MALF[8] | 11,931 | 5,250 | No | No license |
| PASCAL Face[9] | 1,335 | 851 | No | No license |
| IJB-A[10] | 49,759 | 24,327 | No | IJB-A LICENSING |
| UMDFaces[11] | 8,277 | 367,888 | No | No license |
| UFFD[12] | 10,897 | 6,425 | No | Academic use only |

*Table 2: Datasets with bounding box labels for face detection*

With the publication of the WIDER FACE Dataset in 2016, its test and validation subsets became the most common benchmarks, both of which are split into subsets Easy, Medium, and Hard according to the difficulty posed to face detection (the difficulty is mainly influenced by the size of the face, illumination, pose, and occlusion). Typically, precision-recall curves are reported along with the average precision.

---

[5] http://shuoyang1213.me/WIDERFACE/
[6] https://www.ics.uci.edu/~xzhu/face/
[7] http://vis-www.cs.umass.edu/fddb/
[8] http://www.cbsr.ia.ac.cn/faceevaluation/
[9] http://host.robots.ox.ac.uk/pascal/VOC/datasets.html
[10] https://www.nist.gov/itl/iad/image-group/ijb-Dataset-request-form
[11] http://umdfaces.io/
[12] https://ufdd.info/

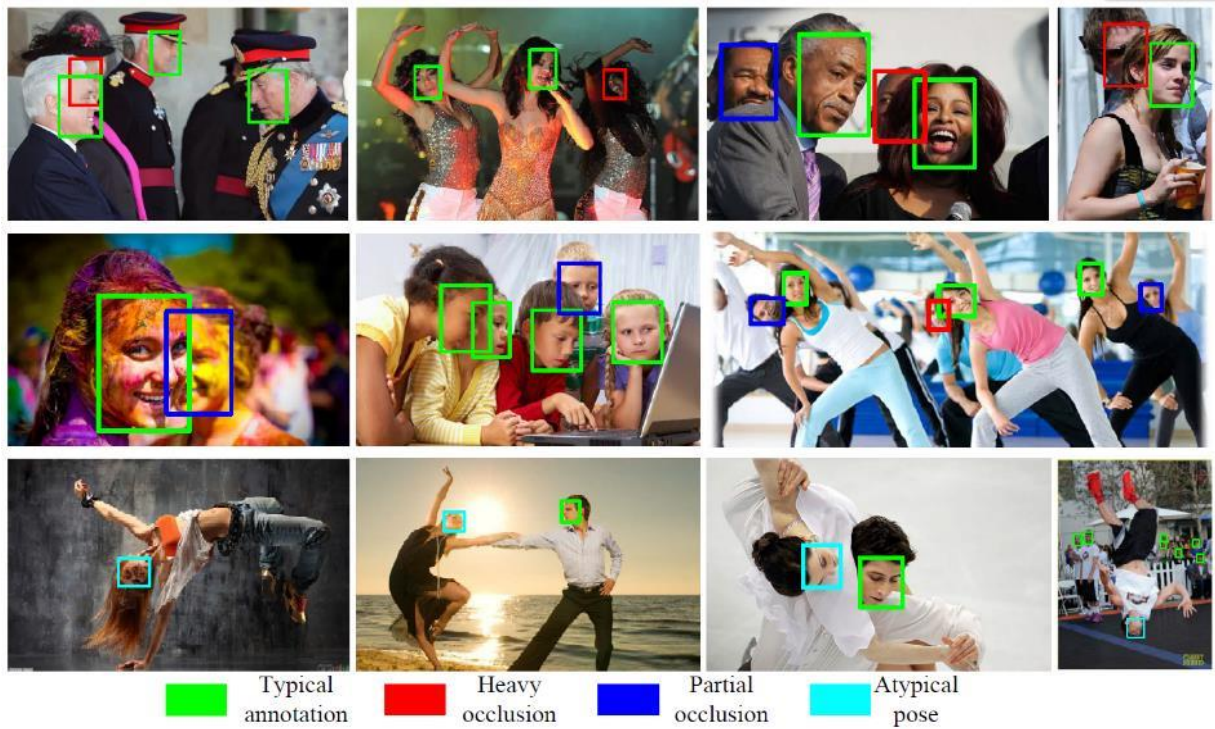

| Typical annotation | Heavy occlusion | Partial occlusion | Atypical pose |

*Figure 1: Example images from WIDER FACE (taken from [2])*

## 2.1.2 Methods and Algorithms

Many approaches exist for face detecting, some of which also output some (often 5) landmarks. Most face detection algorithms can detect several faces in a single image and output bounding boxes and landmarks for all detected faces. Note, that if a face is located close to the image boundary or even only partly depicted in the image, the bounding box can be partly outside the image area.

Generally, approaches for face detection can be classified in

- Approaches using hand-crafted features
- Approaches using deep CNNs.

The most prominent hand-crafted method deployed for face recognition is the Viola-Jones Face Detector [2]. This algorithm uses Haar-like filters (filters similar to Haar wavelets) to detect typical facial features and an AdaBoost classifier to decide whether a considered sub-window contains a face. In order to efficiently eliminate irrelevant sub-windows, a cascade of gradually more complex classifiers (with an increasing number of features) is applied. The Viola-Jones detector is quite old and requires (more or less) a frontal pose. Nevertheless, it is still used in many publications and implementations.

In [3], Histogram of Oriented Gradients (HOG) is deployed with a Support Vector Machine (SVM) as a classifier to detect humans (not just faces) in images. An adaptation of this approach for face detection is still widely in use due to its implementation in the popular library dlib (see Section 2.1.3), but it is much less robust than modern approaches.

In 2014, the HeadHunter face detector [4] applied a so-called *Integral Channels Detector*, which generalizes the feature computation approach used in the Viola-Jones algorithm, on the LUV[13] colour space channels and HOG features. By training a Gradient Boosting classifier on these features for 22 different face templates (with various poses), it achieved, at the time of publication, state of the art performance.

A vast number of CNN-based face detectors has been proposed [5]. One of the first, was the Multitask Cascaded Convolutional Networks (MTCNN) [6] published 2016, which is still widely used in publications and implementations. In order to detect faces of different sizes, this approach first resizes the image to various scales resulting in an image pyramid which is then fed to a cascade of three CNNs, successively refining the proposals of bounding boxes.

Other deep learning-based face detectors were based on regional CNNs (R-CNN, see Glossary).

- Face R-CNN [7], published in 2017, deployed the Faster R-CNN approach.
- Face R-FNC [8], published in 2017, proposed to use a region-based fully convolutional network (ResNet) for face detection, and improved on the accuracy of Face R-CNN.

After 2017, very few face detectors based on R-CNN have been proposed and, hence, it can be considered outdated. Nevertheless, building blocks of R-CNN have been used in the more recent (and very accurate) MogFace face detector described below.

In the period between 2016 and 2019, Single Shot Multibox Detectors (SSD) [9] have been popular for face detection.

- FaceBoxes [10], published 2017, which applies so-called Rapidly Digested Convolutional Layers (RDCL) to achieve real-time face detection.
- S3FD [11], published 2017, combines the SSD approach with region proposal networks (RPN) of R-CNNs.
- BlazeFace [12], published 2019, a lightweight face detector developed by Google Research. It combines a feature extraction CNN related to MobileNet with a modified anchor scheme of SSD as the second part.
- MogFace [13], published 2021, introduced several improvements applied to the architecture of S3FD, and using ResNet50 as a backbone, it achieved the best accuracy of all face detectors to date.

Another popular approach for face detection, which is currently state of the art, are Feature Pyramid Networks (FPN, see Glossary), which combine different levels of hierarchical features to detect faces at different scales. Face detection approaches based on FPN are:

- The Feature Agglomeration Networks for Single Stage Face Detection (FANet) [14], published 2017, which uses VGG16 as its backbone CNN architecture.
- PyramidBox [15], published 2018, which also deploys the VGG16 network as backbone CNN.
- The Selective Refinement Network (SRN) [16], published 2018, which uses a ResNet-50 backbone.
- The Dual Shot Face Detector (DSFD) [17], published 2018, which extends the approach of the FPN by adding so-called Feature Enhance Module (FEM) to create a second set of feature maps from the original ones, and uses an extended VGG16 network as backbone architecture.

---

[13] Actually, LUV is not an abbreviation, see https://en.wikipedia.org/wiki/CIELUV.

- RetinaFace [18], published 2019, which is based on a ResNet-50 backbone, considerably improved on the detection accuracy of the previous face detectors.
- TinaFace [19], published 2020, which deploys ResNet-50 as backbone architecture. At the time of publication, it was the face detector with the highest detection accuracy.
- SCRFD [20], published 2021, which has been proposed for various backbone CNN architectures. In order to reduce the computational costs, it targets small images (with accordingly small faces), and for the WIDER FACE images resized to VGA resolution (640x460), the version with ResNet-50 (coined SCRFD-34GF) improved on the detection accuracy of TinaFace.

YOLO5-Face [21], published 2021, is based on the YOLO5v5 object detector and, for some configurations, it achieves even better performance on small images than SCRFD.

State of the art face detectors already achieve very high precision rates for unconstrained facial images that are not small, occluded or poorly illuminated. For images captured in constrained environments, even better detection performance can be expected, but no corresponding evaluation has been published so far.

| Publication/Detector | Year | Method | WIDER FACE val. set (Easy/Medium/Hard) | WIDER FACE val. set resized to VGA (Easy/Medium/Hard) |
|---|---|---|---|---|
| [2] Viola-Jones | 2001 | Hand-crafted | 41.2/33.3/13.7 [22] | n. a. |
| [4] HeadHunter | 2014 | Hand-crafted | 71.2/58.4/24.9 [23] | n. a. |
| [6] MTCNN | 2016 | Cascaded CNNs | 84.8/82.5/59.8 | n. a. |
| [7] Face R-CNN | 2017 | R-CNN | 93.8/92.2/82.9 | n. a. |
| [8] Face R-FNC | 2017 | R-CNN | 94.7/93.5/87.4 | n. a. |
| [10] FaceBoxes | 2017 | SSD | n. a. | 76.2/57.2/24.2 |
| [11] S³FD | 2017 | SSD | 93.7/92.5/85.9 | n. a. |
| [14] FANet | 2017 | FPN | 95.6/94.7/89.5 | n. a. |
| [15] PyramidBox | 2018 | FPN | 96.1/95.0/88.9 | n. a. |
| [17] DSFD | 2018 | FPN | 96.6/95.7/90.4 | 94.3/91.5/71.4 |
| [16] SRN | 2018 | FPN | 96.4/95.2/90.1 | n. a. |
| [18] RetinaFace | 2019 | FPN | 96.9/96.1/91.8 | 94.9/91.9/64.2 |
| [12] BlazeFace | 2019 | Modified SSD | n. a. | 91.5/89.9/79.7 |
| [21] TinaFace | 2020 | FPN | 97.0/96.3/93.4 | 95.6/94.3/81.4 |
| [13] MogFace-E | 2021 | SSD | 97.7/96.9/93.8 | n. a. |

| Publication/Detector | Year | Method | WIDER FACE val. set (Easy/Medium/Hard) | WIDER FACE val. set resized to VGA (Easy/Medium/Hard) |
|---|---|---|---|---|
| [20] SCRFD-34GF | 2021 | FPN | n. a. | 96.1/94.9/85.3 |
| [21] YOLO5Facev5x6 | 2021 | YOLO | n. a. | 96.7/95.2/86.6 |

*Table 3: Approaches and detection performance (Average Precision) of the described face detectors.*

## 2.1.3 Software

| Name | Method | Framework | Model size | AP (WIDER FACE) | License |
|---|---|---|---|---|---|
| OpenCV | Haar[14] | OpenCV | n. a. | 30.7[15] | Apache 2.0 |
| dlib[16] | HOG+SVM | Own-Library | n. a. | 25.3[15] | Boost Software |
|  | CNN | Own-Library | 0.7 MB | 28.6[15] |  |
| Cheng Chi[17] | SRN | PyTorch | 407 MB | 96.5/95.3/90.2 | Apache 2.0 |
| Tencent[18] | DSFD | PyTorch | 459 MB | 96.6/95.7/90.4 | Academic use only |
| Sr6033[19] | SSD | OpenCV (Caffe) | 10 MB | n. a. | MIT |
| I. de Paz Centeno[20] | MTCNN |  | 3 MB | n. a. | MIT |
| yuanyang[21] | MTCNN |  | 6 MB | n. a. | MIT |
| Zhao Jian[22] | MTCNN |  | 3 MB | n. a. | MIT |
| Google mediapipe[23] | BlazeFace | Own-Library (TFLite) | < 1MB | n. a. | Apache 2.0 |
| hollance[24] | BlazeFace | PyTorch | < 1MB | n. a. | Apache 2.0 |
| peteryuX[25] | RetinaFace | Tensorflow | 12 MB | 93.2/91.7/80.0 | Code: MIT |
|  | RetinaFace | Tensorflow | 194 MB | 94.2/93.3/83.5 | Models: Non-commercial use |

[14] https://docs.opencv.org/3.4.1/d7/d8b/tutorial_py_face_detection.html
[15] https://github.com/nodefluxio/face-detector-benchmark
[16] http://dlib.net/
[17] https://github.com/ChiCheng123/SRN
[18] https://github.com/Tencent/FaceDetection-DSFD
[19] https://github.com/sr6033/face-detection-with-OpenCV-and-DNN
[20] https://github.com/ipazc/mtcnn
[21] https://github.com/YYuanAnyVision/mxnet_mtcnn_face_detection
[22] https://github.com/ZhaoJ9014/face.evoLVe
[23] https://google.github.io/mediapipe/
[24] https://github.com/hollance/BlazeFace-PyTorch
[25] https://github.com/peteryuX/retinaface-tf2

| Name | Method | Framework | Model size | AP (WIDER FACE) | License |
|---|---|---|---|---|---|
| Serengil[26] | RetinaFace | Tensorflow | 113 MB | n. a. | MIT |
| Ternaus[27] | RetinaFace | PyTorch | 209 MB | n. a. | MIT |
| Charrin[28] | RetinaFace | Caffe | 113 MB | 96.5/95.6/90.4 | No license |
| | RetinaFace | Caffe, ncnn | 1.7 MB | n. a./n. a./82.5 | No license |
| FacePose_pytorch[29] | RetinaFace | OpenCV (Caffe) | 1.8 MB | n. a. | MIT |
| biubug6[30] | RetinaFace | PyTorch | 1.7 MB | 90.7/88.2/73.8 | MIT |
| | | | 104 MB | 95.5/94.0/84.4 | |
| Deep Insight[31] | RetinaFace | MXNET | 105 MB | 96.5/95.6/ 90.4 | Code: MIT Models and Data: Non-commercial use |
| | RetinaFace | MXNET | 1.7 MB | n. a./n. a./82.5 | No license |
| | SCRFD | PyTorch | 2.7 MB | 93.8/92.2/77.8 | No license |
| | SCRFD | PyTorch | 15 MB | 95.2/93.9/83.1 | No license |
| | SCRFD | PyTorch | 38 MB | 96.1/94.9/85.3 | No license |
| | BlazeFace | PaddlePaddle | 0.6 MB | 91.9/89.8/81.7 | No license |
| Media-Smart[32] | TinaFace | PyTorch | 145 MB | 96.3/95.7/93.0 | Apache 2.0 |
| Deepcam[33] | BlazeFace | PyTorch | 0.5 MB | 90.4/88.7/78.0 | GPL-3.0 |
| | Yolo5-Face | PyTorch | 14 MB | 93.6/91.5/80.5 | No license |
| | Yolo5-Face | PyTorch | 54 MB | 94.3/92.6/83.2 | No license |
| | Yolo5-Face | PyTorch | 161 MB | 95.3/93.8/85.3 | No license |
| | Yolo5-Face | PyTorch | 356 MB | 95.8/94.3/86.1 | No license |
| DamoCV[34] | MogFace | PyTorch | 160 MB | 95.1/94.2/87.4 | No license |
| | | PyTorch | 327 MB | 97.7/96.9/92.0 | No license |

*Table 4: Publicly available implementations of face detectors which are state of the art*

---

[26] https://github.com/serengil/retinaface
[27] https://github.com/ternaus/retinaface
[28] https://github.com/Charrin/RetinaFace-Cpp
[29] https://github.com/WIKI2020/FacePose_pytorch
[30] https://github.com/biubug6/Pytorch_Retinaface
[31] https://github.com/deepinsight/insightface/tree/master/detection/
[32] https://github.com/Media-Smart/vedadet/
[33] https://github.com/deepcam-cn/yolov5-face
[34] https://github.com/damo-cv/MogFace

## 2.2   Face Segmentation

### 2.2.1   Datasets

The following table summarizes the most popular datasets containing face images and ground truth annotations of the face area.

| Name | Images | Remark | License |
|------|--------|--------|---------|
| CelebAMask-HQ[35] | 30.000 | Hair, face, eyes, eyebrows, ears, nose, lips, mouth, hat, eyeglass, earring, necklace, neck, and cloth | Non-commercial use |
| LFW - Part Labels Dataset[36] | 2.927 | Hair, face (incl. neck) | No license |
| Helen with segmenetation annotations[37] | 2.330 | Hair, face, eyes, eyebrows, nose, lips | No license |
| iBugMask[38] | 1.000 | Hair, face, eyes, eyebrows, nose, lips | No license |
| LaPa-Dataset[39] | 22.000 | Hair, face, eyes, eyebrows, nose, lips, mouth | Non-commercial use |
| FaceOcc[40] | 30.000 | Un-occluded parts of face | No license |

*Table 5: Publicly available datasets for face segmentation*

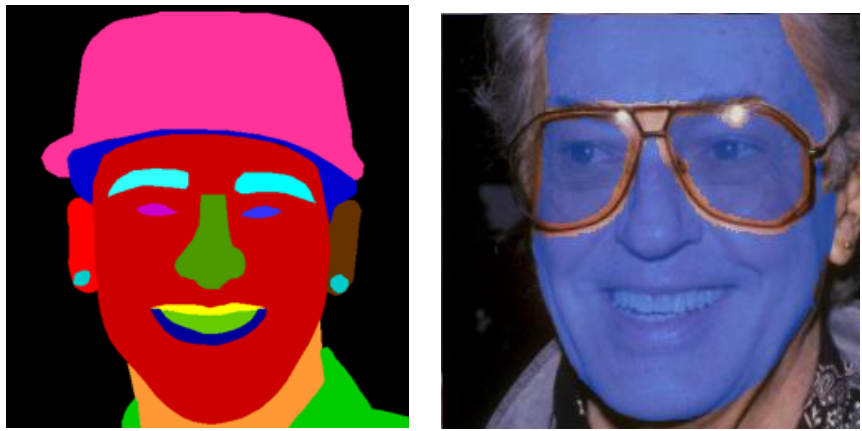

*Figure 2: Example images from CelebAMask-HQ and FaceOcc showing the ground truth segmentation maps*

---

[35] https://github.com/switchablenorms/CelebAMask-HQ
[36] http://vis-www.cs.umass.edu/lfw/part_labels/
[37] https://pages.cs.wisc.edu/~lizhang/projects/face-parsing/
[38] https://ibug.doc.ic.ac.uk/resources/ibugmask/
[39] https://github.com/lucia123/lapa-dataset
[40] https://github.com/face3d0725/FaceExtraction

## 2.2.2 Methods and Algorithms

Many methods have been proposed for face segmentation, also referred to as face parsing. Some publications limit on the segmentation of just the visible parts of the face without any regions occluded by hair, sunglasses, hands or objects, e.g. Saito et al. [24], Nirkin et al. [25], Masi et al. [26] or Yin et al. in [27]; the method of Masi also excludes beards, while the method of Yin has been trained on the FaceOcc dataset, in which even frames of eyeglasses and the tongue are labelled as background (see Figure 2 and Figure 24). Other publications focus on the segmentation of face and face parts (eyes, mouth, nose, eyebrows) without hairs, e.g. Smith et al. [28] or Te et al. [29]. However, most publications also perform segmentation of the hairs, e.g. AGRNet proposed by Te et al. [30], Liu et al. in [31] and [32], or Lin et al. [33]. There are even methods that also perform segmentation of the neck, hats and the clothing, e.g. EHANet proposed by Luo [34], which has been trained on CelebA-Mask-HQ (see Figure 2).

## 2.2.3 Software

| Name | Framework | Segmentation | Model Size | License |
|------|-----------|--------------|------------|---------|
| CelebAMask-HQ Face Parsing[41] | PyTorch | Face, parts of face, hair, hat, neck, clothing | 7.5 MB | Non-commercial |
| face-parsing.PyTorch[42] | PyTorch | Face, parts of face, hair, hat, neck, clothing | 51 MB | MIT |
| FaceParsing.PyTorch[43] [34] | PyTorch | Face, parts of face, hair, hat, neck, clothing | 96 MB | MIT |
| face-seg[44] | PyTorch | Face incl. neck, hair | 14 MB | No license |
| Deep face segmentation in extremely hard conditions[45] [25] | Caffe | Non-occluded parts of face (incl. frames of eyeglasses) | 475 MB | Apache 2.0 |
| face-segmentation-keras[46] [25] | Tensorflow | Non-occluded parts of face (incl. frames of eyeglasses) | 513 MB | MIT |
| FaceExtraction[40] [27] | PyTorch | Non-occluded parts of face (excl. frames of eyeglasses) | 55 MB | Non-commercial |
| RoI Tanh-polar Transformer Network for Face Parsing in the Wild[47] | PyTorch | Face, parts of face | 105 MB  158 MB | MIT |

[41] https://github.com/switchablenorms/CelebAMask-HQ/tree/master/face_parsing
[42] https://github.com/zllrunning/face-parsing.PyTorch
[43] https://github.com/TracelessLe/FaceParsing.PyTorch
[44] https://github.com/kampta/face-seg
[45] https://github.com/YuvalNirkin/face_segmentation
[46] https://github.com/shaoanlu/face-segmentation-keras
[47] https://github.com/hhj1897/face_parsing#roi-tanh-polar-transformer-network-for-face-parsing-in-the-wild

| Name | Framework | Segmentation | Model Size | License |
|------|-----------|--------------|------------|---------|
| Decoupled Multi-task Learning with Cyclical Self-Regulation for Face Parsing[48] | PyTorch | Model trained on CelebAMask-HQ: Face, parts of face, hair, hat, neck, clothing | 272 MB | Non-commercial use |
| Towards Learning Structure via Consensus for Face Segmentation and Parsing[49] [26] | PyTorch | Non-occluded parts of face (incl. frames of eyeglasses, excl. beard), occluded parts incl. hair and beard | 17 MB | Academic or non-commercial |

*Table 6: Examples of software available for face segmentation.*

## 2.3    Facial Landmark Localization

### 2.3.1    Datasets

The following table summarizes the most popular Datasets containing face images and ground truth landmark annotations. The column "variations" specifies the type of image variants comprised in the dataset: expression (e), illumination (i), occlusion (o) and pose(p).

| Name | Images | Variations | Landmarks | Remark | Constrained? | License |
|------|--------|------------|-----------|--------|--------------|---------|
| BioID[50] | 1,521 | e, i | 20 | - | Yes | No license |
| AR Face[51] | 4,000 | e, i, o, | 22 | - | Yes | Royalty-free, Non-commercial use |
| CMU Multi-PIE[52] | 750,000 | e, i, p | 68 or 39 | - | Yes | Royalty-bearing |
| XM2VTS[53,54] | 2.360 | p | 68 | - | Yes | Royalty-bearing |
| FRGCv2[55,56] | 45,000 | e, i | 68 | - | Yes | No license |
| BU-4DFE[57] | 3000 | e | 68 | - | Yes | Royalty-free, Non-commercial use |
| AFLW[58] | 25,933 | e, i, o, p | 21 | - | No | Royalty-free, Non-commercial use |

---

[48] https://github.com/deepinsight/insightface/tree/master/parsing/dml_csr
[49] https://github.com/isi-vista/structure_via_consensus
[50] https://www.bioid.com/facedb
[51] https://www2.ece.ohio-state.edu/~aleix/ARdataset.html
[52] https://cmu.flintbox.com/technologies/67027840-27d5-4570-86dd-ad4715ef3c09
[53] http://www.ee.surrey.ac.uk/CVSSP/xm2vtsdb/
[54] ttps://personalpages.manchester.ac.uk/staff/timothy.f.cootes/data/xm2vts/xm2vts_markup.html
[55] https://cvrl.nd.edu/projects/data/#face-recognition-grand-challenge-frgc-v20-data-collection
[56] https://ibug.doc.ic.ac.uk/resources/facial-point-annotations/
[57] https://www.cs.binghamton.edu/~lijun/Research/3DFE/3DFE_Analysis.html
[58] https://www.tugraz.at/institute/icg/research/team-bischof/lrs/downloads/aflw/

| Name | Images | Variations | Landmarks | Remark | Constrained? | License |
|------|--------|-----------|-----------|--------|--------------|---------|
| Helen[59,56] | 2,330 | e, i, o, p | 194 & 68 | - | No | No license |
| AFW[60,56] | 205 | e, i, o, p | 68 | - | No | No license |
| 300-W[61,56] | 399 | e, i, o, p | 68 | - | No | Royalty-free, Non-commercial use |
| ibug300[61,56] | 135 | e, i, o, p | 68 | Subset of 300-W | No | Royalty-free, Non-commercial use |
| COFW[62] | 1,007 | e, i, o, p | 29 | Occluded faces | No | CC-BY |
| WFLW[63] | 10,000 | e, i, o, p | 98 | - | No | No license |
| JD-landmark[64] | 24,000 | e, i, o, p | 106 | - | No | Royalty-free, Non-commercial use |
| MERL-RAV[65] | 19.000 | e, i, o, p | 68 | Annotations for AFLW | No | No license |
| PUT Face Database[66] | 9,971 | p | 30 | 194 landmarks provided for 2,971 images | Yes | Royalty-free, research purposes |
| SCFace[67,68] | 4.160 | p | 21 | | Yes | Royalty-free, research purposes |
| MUCT[69] | 3,755 | p | 76 | - | Yes | No license |
| LS3D-W[70] | ~230,000 | e, i, o, p | 68 | 2D and 3D landmark annotations for AFLW, 300-W, 300VW, FDDB | No | Royalty-free |
| SCUT-FBP550[71] | 5.500 | e, i, o, p | 86 | - | No | Royalty-free, academic research |

| Name | Images | Variations | Landmarks | Remark | Constrained? | License |
|------|--------|-----------|-----------|--------|--------------|---------|
| IMM Frontal Face Database | 120 | e | 73 | | Yes | Royalty-free, academic research |
| FEI[72] | 400 | e, p | 46 | | Yes | Royalty-free, research purposes |

*Table 7: Datasets with facial landmark annotations*

The landmarks annotated in the listed datasets differ in their number and defined locations, see Figure 3. The most commonly used landmark definition system is that of ibug300 with 68 landmarks.

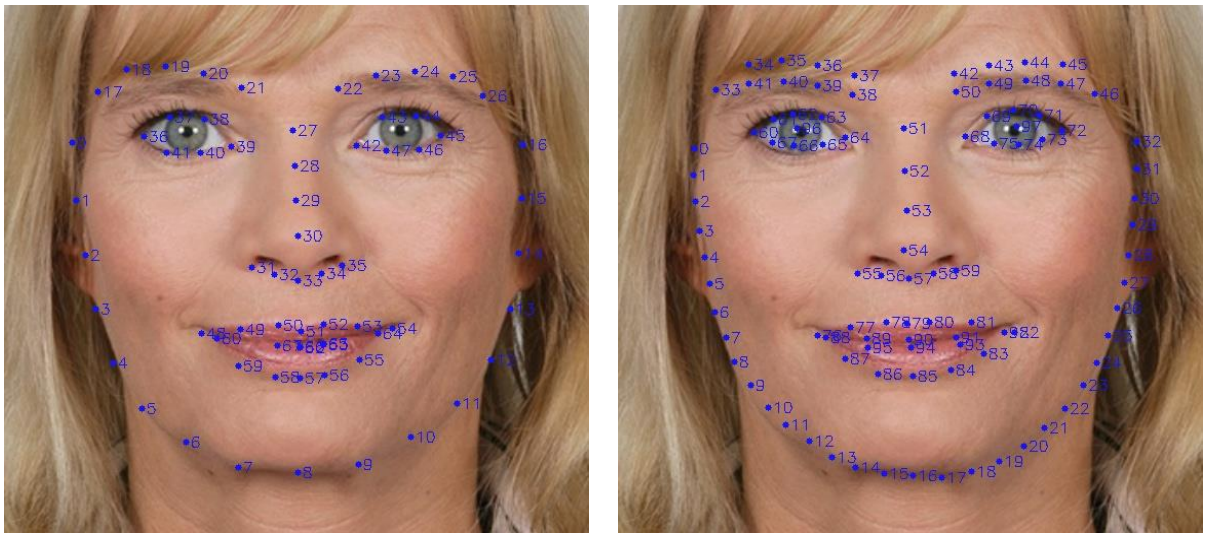

*Figure 3: Application of the landmark definitions of ibug300 (left) and WFLW (right) to the same image (original images taken from http://jbusse.de).*

The accuracy of facial landmark localization algorithms is typically measured by the *Normalized Mean Error (NME)*, where

- the *Error* is defined by the Euclidean distance between the computed landmark and the ground truth position,
- the *Mean* is taken over all landmarks and images, and
- the *Normalization* is a division by a specific face-dependent distance, e.g. the Inter-Ocular-Distance (IOD) or the Inter-Pupil-Distance (IPD), to ensure that the error does not depend on the scale of the face.

## 2.3.2 Methods and Algorithms

Since in the current project, facial landmarks are needed for assessing criteria related to specific properties of certain face parts, e.g. mouth close, eyes closed, algorithms are needed that determine a sufficient number of landmarks per face part. Therefore, face detectors like MTCNN, RetinaFace, BlazeFace or YOLO5Face, that output just 5 or 6 landmarks per face are not discussed.

---

[72] https://fei.edu.br/~cet/facedatabase.html

Early methods for facial landmark detection were mainly based on fitting a deformable face mesh to the face image. One of the most prominent algorithms is the *Active Shape Model (ASM)* [35], which is implemented in the library stasm. These algorithms provide sufficient accuracy for images captured under controlled conditions, but fail for rotated poses, poor illumination, or partially occluded faces.

A better robustness and accuracy at real-time speed was achieved by the approach of [36], published in 2014, which applies gradient boosting on an *Ensemble of Regression Trees (ERT)* using the pixel intensities as features. This approach in implemented in the popular library dlib and it is among the fastest landmark localization algorithms.

After 2015, almost all proposed methods have been based on CNNs. Almost all of the CNN-based methods are not bound to a fixed number of landmarks, but rather can be (and typically have been) trained on various Datasets resulting in different numbers of landmarks.

While some approaches train a CNN to predict the landmark locations directly, others use CNNs that output a set of heatmaps, one per landmark, where each entry in the heatmap specifies the probability of the landmark to be located at the corresponding position. In [37], this approach was applied to a CNN architecture called *Stacked Hourglass*, which, several times, successively reduces the feature dimensions and then expands them back to the input dimensions, an approach reminiscent of auto-encoders.

Using a modified loss functions, *Wing Loss*, for the training, a more accurate CNN for (direct) facial landmark regression was obtained in [38]. This loss function limits the influence of outliers, so that samples where the landmarks estimations are completely wrong do not dominate the error values.

The *Style Aggregated Network (SAN)* [39] is an approach, where a Generative Adversarial Network (GAN) is applied to "normalize" the image with respect to its "style" (referring to contrast, illumination, and saturation) before a CNN is applied on both images (normalized and original) to predict the landmarks (directly). This algorithm achieved a higher accuracy than previous methods.

In the *Look at Boundary (LAB)* approach [40], a CNN is applied to compute a boundary heatmap of the facial features (mouth, nose, eyes, eyebrows, contour) which is then fed into another CNN to predict the actual facial landmark locations. This approach provides considerably improved accuracy as compared to previous approaches.

Another algorithm for landmark localization is the Practical Facial Landmark Detector (PFLD) [41], in which MobileNetV2 is used as a backbone for feature extraction and the samples are weighted in the error function according to their difficulty. The PFLD algorithm is very lightweight and accurate.

The *AWing* algorithm [42] is trained on a modified version of the Wing Loss (the algorithms name is an acronym of Adaptive Wing Loss), uses ideas of the LAB and the heatmap approach, and computes landmarks via corresponding heatmaps. Using this approach, the authors achieved highly accurate predictions.

The approach Deep Adaptive Graph (DAG) [43] represents the landmarks as a graph and applies so-called Graph Convolutional Networks (GCNs) to predict them. Their algorithm is currently the most accurate landmark localization algorithm.

There are also approaches that compute facial landmarks by fitting 3-dimensional model, a face mesh consisting of ten thousands of points, to the face. An early approach was 3D Dense Face Alignment (3DDFA) [44], in which a dense 3D face model was fitted to the image by a CNN. SynergyNet [45] achieved an improved accuracy by exploiting the synergy between a 3D Morphable Models and 3D facial landmarks. In [46], the approach of 3DDFA was further improved by dynamically choosing one of two loss functions during training.

In [47], researchers from Google have trained a CNN on predicting a 3D morphable model with 468 landmarks, resulting in 468 3-Dimensional landmarks. This CNN is used in conjunction with a model for predicting 8 additional landmarks for the irises [48] in Google's library mediapipe to compute 476 3-dimensional landmarks. In contrast to other landmark semantics (e.g. iBug, WFLW), not all landmarks of Googles 3DMM mark boundaries of facial features, i.e. many landmarks are in located between or outside contours (e.g. on the cheeks, forehead, around the nose, on the lips), see Figure 4.

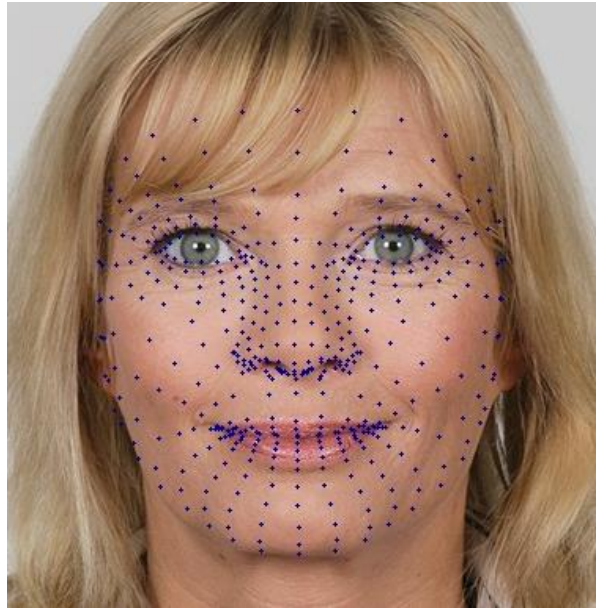

*Figure 4: the landmarks detected by mediapipe do not (all) correspond to specific face features but constitute a mesh (original image taken from http://jbusse.de)*

The described methods are summarized in the following table. Note that the accuracy values all refer to datasets of images captured in unconstrained environments. In fact, there are no public evaluation results of facial landmark localization algorithms for constrained images.

| Name | Year | Method | NME | | | |
|------|------|--------|-----|---|---|---|
| | | | Full 300-W (IPD/IOD) | COFW (IPD/IOD) | HELEN | AFLW2000-3D |
| ASM [35] | 1995 | Deformable Mesh | n. a. | n. a. | 11.1 | n. a. |
| ERT [35] | 2014 | Machine Learning | 6.40/n. a. | n. a. | 4.9 | n. a. |
| Stacked Hourglass [37] | 2017 | CNN (heatmap) | n. a. | 5.6/4.0 | n. a. | n. a. |
| Wing Loss [38] | 2017 | CNN | 3.60/n. a. | 5.44/n. a. | n. a. | n. a. |

| Name | Year | Method | NME | | | |
|------|------|--------|-----|-----|-----|-----|
| | | | **Full 300-W (IPD/IOD)** | **COFW (IPD/IOD)** | **HELEN** | **AFLW2000-3D** |
| SAN [39] | 2018 | CNN | n. a./3.98 | n. a. | n. a. | n. a. |
| LAB [40] | 2018 | CNN | 4.12/3.49 | n. a./3.92 | n. a. | n. a. |
| PFLD [41] | 2019 | CNN | 3.76/3.37 | n. a. | n. a. | n. a. |
| AWing [42] | 2019 | CNN (heatmap) | 4.31/3.07 | 4.94/n. a. | n. a. | n. a. |
| DAG [43] | 2020 | CNN (GCN) | 4.27/3.04 | n. a. | n. a. | n. a. |
| 3DDFA [44] | 2015 | CNN (3D Landm.) | n. a. | n. a. | n. a. | 4.94 |
| SynergyNet [45] | 2021 | CNN (3D Landm.) | n. a. | n. a. | n. a. | 4.06 |
| 3DDFA_V2 [46] | 2020 | CNN (3D Landm.) | n. a. | n. a. | n. a. | 3.51 |

*Table 8: Summary of methods and their accuracy on different Datasets expressed by the Normalized Mean Error (NME) using normalization by Inter-Ocular-Distance (IOD) or Inter-Pupil-Distance (IPD). Results on Helen refer to labels for 194 landmarks and results on AFLW2000-3D refer to labels for 68 landmarks.*

### 2.3.3 Software

The following table lists the freely available implementations for landmark localization. In addition, many commercial face recognition and computer vision SDKs, e.g. of Neurotechnology, Dermalog, Eyedea, Luxand, Apple, Google, provide functions to compute facial landmarks (the number of which depends on the SDK).

| Name | Method | Framework | Model size | Number of Landmarks | License |
|------|--------|-----------|------------|---------------------|---------|
| stasm[73] | ASM | Own-Library, OpenCV | n. a. | 77 | Royalty-free |
| dlib[16] | ERT | Own Library | 97 MB | 68 | Boost Software |
| mediapipe[74] | Google's 3DMM | Own Library (TFLite) | 6 MB | 476 (3D) | Apache 2.0 |
| 3DDFA_V2[75] | 3DDFA_V2 | PyTorch / ONNX | 13 MB | 68 (3D) | MIT |
| | | PyTorch / ONNX | 4 MB | 68 (3D) | |
| Dense-Head-Pose-Estimation[76] | 3DDFA_V2 | TFLite | 13 MB | 68 (3D) | MIT |

---

[73] http://www.milbo.users.sonic.net/stasm/
[74] https://mediapipe.dev/
[75] https://github.com/cleardusk/3DDFA_V2
[76] https://github.com/1996scarlet/Dense-Head-Pose-Estimation

| Name | Method | Framework | Model size | Number of Landmarks | License |
|---|---|---|---|---|---|
| SynergyNet[77] | SynergyNet | PyTorch | 72 MB | 68 (3D) | MIT |
| D-X-Y[78] | SAN | PyTorch | 262 MB | 68 | MIT |
| Wayne Wu[79] | LAB | Caffe | 44 | 98 | Royalty-free |
| protossw512[80] | AWing | PyTorch | 95 MB[3] | 98 | Apache 2.0 |
| yxqAIxp[81] | PFLD [41] | PyTorch | 17 MB | 68 | No license |
|  |  |  | 1.1 MB |  |  |
|  |  |  | 7 MB |  |  |
| zhaozhichao[82] | PFLD | PyTorch | 7 MB | 98 | No license |
| zuoqing1988[83] | (unknown) | TVM | 1 MB | 106 | MIT |
|  |  |  | 2 MB |  |  |
|  |  |  | 8 MB |  |  |
| FacePose_pytorch[29] | PFLD | PyTorch | 5 MB[3] | 98 | MIT |
| Xintao[84] | PFLD | PyTorch | 7 MB | 106 | No license |
|  |  |  | 2 MB |  |  |
| AmrElsersy[85] | PFLD | PyTorch | 7 MB | 98 | No license |
| AnthonyF333[86] | PFLD | NNCN | 3 MB | 98 | Apache 2.0 |
| samuelyu2002[87] | PFLD | PyTorch | 7 MB | 68 | MIT |
| Insightface[88] | (unknown) | MXNET | 5 MB | 106 | Non-commercial use |
|  | (unknown) | PyTorch | 182 MB | 68 (3D) |  |

*Table 9: Publicly available implementations of state of the art algorithms for facial landmark localization.*

---

[77] https://github.com/choyingw/SynergyNet
[78] https://github.com/D-X-Y/landmark-detection
[79] https://github.com/wywu/LAB
[80] https://github.com/protossw512/AdaptiveWingLoss
[81] https://github.com/github-luffy/PFLD_68points_Pytorch
[82] https://github.com/polarisZhao/PFLD-pytorch
[83] https://github.com/MirrorYuChen/tvm_106landmarks
[84] https://github.com/Hsintao/pfld_106_face_landmarks
[85] https://github.com/AmrElsersy/PFLD-Pytorch-Landmarks
[86] https://github.com/AnthonyF333/FaceLandmark_PFLD_UltraLight/issues/1
[87] https://github.com/samuelyu2002/PFLD
[88] https://github.com/deepinsight/insightface

# 3    Unified Quality Score

## 3.1    Impact on Face Recognition

Most sections in this report describe individual factors (e.g. head pose) potentially affecting the utility of an image in the context of facial recognition (and hence the accuracy). This section, on the other hand, describes end-to-end methods aimed at determining a holistic ("unified") measure of quality of a facial image. Consequently, the impact on accuracy is entirely due to the use of the resulting quality scores in practice, e.g. for rejection of low-quality images. Experimental evaluations of the surveyed methods rely on the so-called "error-versus-reject characteristic" (ERC) methodology, see Grother et al [49]. An ERC plot shows an error rate (typically the false non-match rate) of a recognition algorithm in relation to the fraction of images rejected as a result of insufficient quality indicated by the quality assessment algorithm. An ERC curve begins with a "starting error" for the case where no images are rejected, whereupon the error rate usually decreases as more images of poor quality are being rejected. Multiple quality assessment methods can be benchmarked by simultaneously plotting ERC curves in a single figure. An example for ERC-based benchmark is shown in Figure 5.

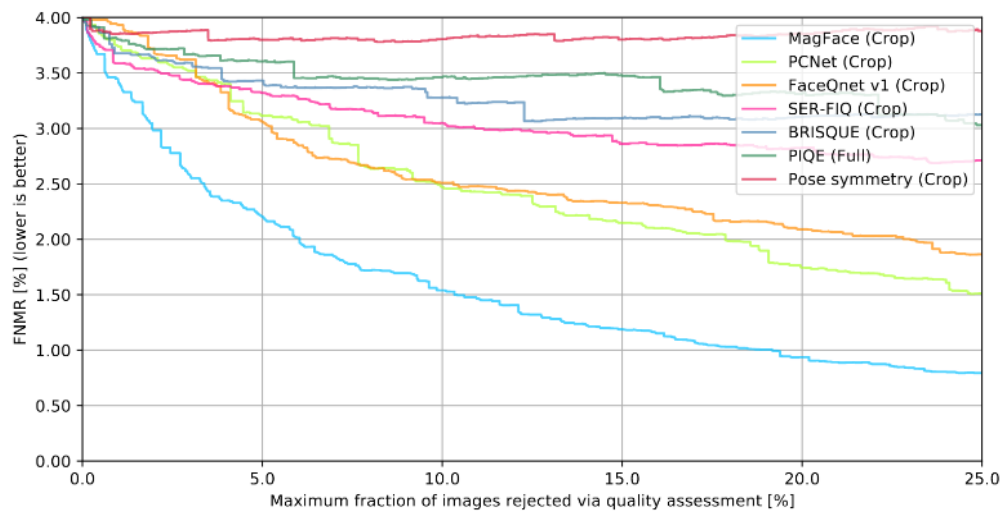

*Figure 5: Example benchmark of quality assessment methods using ERC [43]*

In addition to the graphical representation, a quantitative benchmark can be conducted by computing a (partial) area under curve, for instance between 0 and 10% images being rejected. It should be noted, that in a deployment of a quality assessment algorithm in an operational scenario, the algorithm would naturally be configured to reject images below a certain quality threshold and not a fraction of images.

Additionally, some of the surveyed methods incorporate the image quality directly in the extracted feature vectors or in computation of biometric comparison scores. In those cases (see e.g. [50] and [51]), the authors reported a small increase of face recognition accuracy.

## 3.2    Datasets

Since there is no defined semantic/metric for uniform quality scores, there are no datasets containing corresponding ground truth labels. Instead, publications on uniform quality scores use ERC for evaluation. The following table list the publicly available datasets commonly used in such evaluations.

| Name | Subjects | Images | Remark | Constrained? | License |
|------|----------|--------|--------|--------------|---------|
| LFW[89] | 5,749 | 13,233 | Including derivates such as CALFW[90], CPLFW[91], and XQLFW[92]. | No | No license |
| CFP[93] | 500 | 7,000 | - | No | No license |
| Adience[94] | 2,284 | 26,580 | - | No | CC |
| IJB-B[95] | 1,845 | 76,800 | - | No | Mostly CC BY 3.0 and public domain, also other CC variants |
| IJB-C[95] | 3,531 | 148,800 | - | No | Mostly CC BY 3.0 and public domain, also other CC variants |
| AgeDB[96] | 568 | 16,488 | - | No | Non-commercial use |
| VGGFace2[97] | 9,131 | 3.31 million | Currently offline | No | CC BY-NC 4.0 |

*Table 10: Summary of available datasets regarding Unified Quality Score*

## 3.3    Methods and Algorithms

A Unified Quality Score can be computed as a fusion of several quality metrics described in this report or as a separate quality metric that simply returns a scalar value as a measure of image quality. The vast majority of the surveyed methods can operate in a "no-reference" (also called "blind") mode, i.e. they are capable of outputting a quality score for a single input image. Training, however, often requires reference images, for example to compute comparison scores or as examples

---

---

   

of (almost) optimal facial image quality. Very broadly, two types of methods for unified quality assessment can be distinguished: face specific and general purpose, as summarised below.

### 3.3.1 General Purpose

These methods are not limited to facial images; instead they aim to assess the broadly understood quality of any arbitrary image. The proposed methods often aimed at learning to detect and quantify factors such as compression artefacts, noise, and other synthetic distortions, as well as distortions resulting from the image acquisition itself such as poor illumination. This can be achieved by focussing on measurable deviations from statistical regularities observed in natural images. The following figure shows an example image with different types of distortions applied to it.

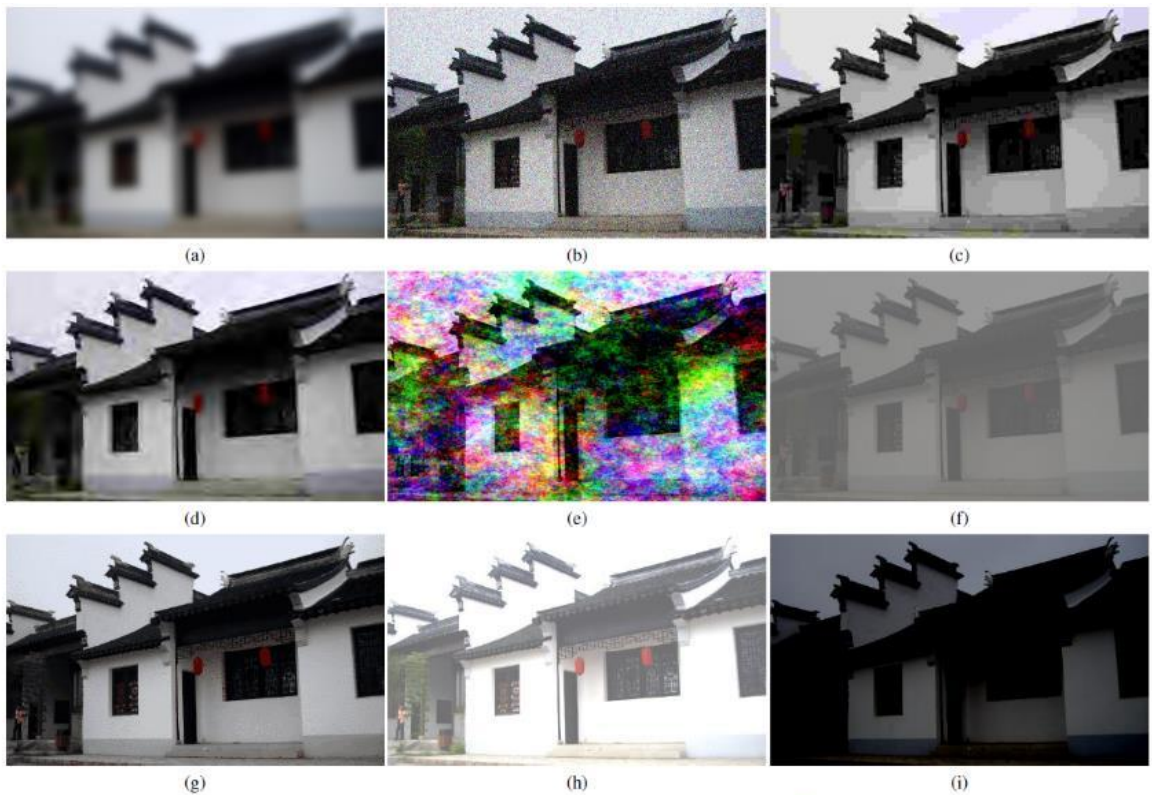

Fig. 1. Sample distorted images synthesized from a reference image in the Waterloo Exploration Database [19]. (a) Gaussian blur. (b) White Gaussian noise. (c) JPEG compression. (d) JPEG2000 compression. (e) Pink noise. (f) Contrast stretching. (g) Image color quantization with dithering. (h) Over-exposure. (i) Under-exposure.

*Figure 6: Training images for general-purpose image quality assessment algorithms [55]*

Furthermore, methods have been proposed aiming at or additionally including the general assessment the aesthetics of the scenery. Such methods are trained using datasets with human assessment "ground truth", i.e. the human perception of the quality and/or aesthetics of an image. While there are obvious limitations to such datasets which are strongly depending on the opinions of the human annotators, pooling multiple human assessments for each image in a dataset allows to approximate a "consensus" quality assessment value based on the human perception of the images. This report focuses on face-specific methods, but several well-known general-purpose methods which have also been applied in the context of face quality assessment are briefly described.

One of the most well-known methods of this type has been proposed by Mittal et al. [52]. BRISQUE is a holistic, no-reference image quality measure based on statistics of luminance coefficients. It aims to quantify deviations from so-called "normalness" of an image scene, which indicate the presence of distortions. Although this method has since been superseded by newer methods, often based on deep-learning, it is included here as it has been extremely widely cited in the scientific literature and is present as a baseline in many benchmarks.

The work of Liu et al. [53] generated a large set of training data by synthetically degrading (compression, noise, blur etc.) images. Since the magnitudes of the degradations are known (i.e. for a pair of images with a given distortion, it is known which has a higher quality), the authors proposed to train a Siamese network to rank pairs of images. Subsequently, they applied knowledge transfer to a CNN in order to assess the quality of single images. Specifically, a single branch from the Siamese network is taken to extract the image representations and fine-tuned using human-labelled image data.

The method proposed in Talebi et al. [54] attempts to predict the distribution (instead of a single value) of human opinion (aesthetics) regarding an image quality. A single score can be output as an average of the distribution, possibly with further statistical moments such as standard deviation. The neural network architecture is strongly inspired by existing image classification networks and has been trained on image datasets with human ratings. The method seems to implicitly detect (i.e. the output scores which correlate with) typical general image distortions such as JPEG compression, poor contrast, blur, etc. In Zhang et al. [55], the use of separate networks for detecting synthetic (e.g. JPEG compression, Gaussian noise) and authentic (i.e. due to variations in image acquisition) distortions has been proposed. A simple CNN for synthetic distortions and a VGG-16-based network for authentic distortions are trained and pooled together to obtain a holistic image quality measure.

The above methods have been included in several publications proposing and benchmarking face-specific quality assessment algorithms. While they were mostly outperformed by the face-specific algorithms, they nevertheless exhibited substantial predictive power w. r. t. face quality assessment. As such, an interesting avenue of research might be coupling the general-purpose methods aimed at detecting broadly understood image distortions with face-specific methods which have been fine-tuned to this specific application domain.

### 3.3.2 Face Specific

Numerous quality assessment algorithms for facial images relying on neural networks have been proposed in the last years. Although the methods exhibit considerable variation and creativity in architecture, protocol design, and training of the networks, they are effectively quite similar. They often rely on similar assumptions, most prominent of which are:

- The compliance according to ICAO 9303 and ISO/IEC 19794-5[98] being a good approximation of actual image quality for the purpose of face recognition.

---

[98] The papers often refer to previous editions of the ICAO 9303 specifications, which contained explicit requirements for facial image quality, while the current (eighth) edition just refers to ISO/IEC 39794-5:2019. Likewise, the papers often refer to previous versions of ISO/IEC 19794-5.

- Comparison scores between mated face samples being strongly correlated with the quality of the samples. (i.e. comparison scores being predictive of quality, and conversely, quality being predictive of comparison scores)
- Variations and/or uncertainties in mated embeddings extracted by face recognition DNNs strongly correlating with differences in image quality.
- Parts of these assumptions, albeit intuitive, remain to be rigorously proven in practice.

One of the first neural-network based methods for facial quality assessment was presented by Hernandez-Ortega et al., who proposed FaceQNet [56]. Their measure of quality can be thought of as a distance between an image and a hypothetical, perfectly ICAO-compliant image. They obtained training data by comparing images against perfectly ICAO compliant images of corresponding subjects. Feature extraction and template comparison was performed with FaceNet [57], whereupon a pre-trained face recognition network (ResNet-50-based) was fine-tuned using the generated data. The resulting network takes a single image as input and produces an output of a single floating-point value representing the quality of the input image, see Figure 7.

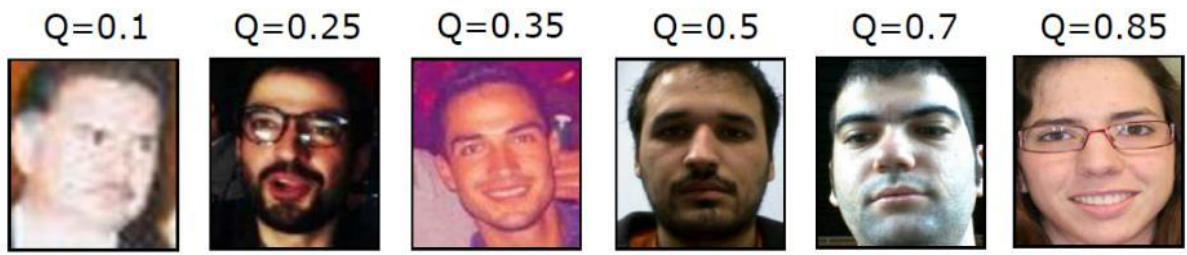

<p align="center"><em>Figure 7: Example images with FaceQNet quality scores [56]</em></p>

In a subsequent follow-up publication [58], slight modifications in the network architecture and training protocol were introduced to improve the ground truth data generation and prevent overfitting. The approach introduced by FaceQNet can be trained using an arbitrary face recognition model or normalised score-level fusion of multiple arbitrary face recognition models, but it operates under the assumption that the degree of ICAO-compliance is indeed a suitable image quality metric. An implementation of FaceQNet has been submitted to and tested in NIST FRVT Quality Assessment.

Several other methods utilised the idea of generating training data for quality assessment using face recognition comparison scores. Xie et al. [59] computed millions of mated comparison scores are using a face recognition network (ResNet-34-based). It is assumed that imperfect comparison scores are due to missing information (i.e. imperfect quality) in the image pair. A further assumption is that the comparison score is determined entirely by the image with the lower quality within each pair. Using the scores, a new network is trained (ResNet-18 architecture) to estimate the predictive confidence (quality) for single images. Ou et al. [60] go beyond using merely mated samples and intra-class distances and includes non-mated samples and inter-class distances in the training of the quality assessment network. Their method relies on the assumption that high-quality samples must exhibit low intra-class variation as well as high inter-class variation. Boutros et al. [61] utilised a similar idea by learning to predict the allocation of samples in feature space w. r. t. their own cluster centres, as well as the centres of their respective nearest negative class centres. The approaches are theoretically conceptually agnostic of the used face recognition system and network architecture. In this context, methods proposed by Chen et al. [62] and Peng et al. [63] are worth mentioning. They rely on conceptually similar training data generation and quality estimation network training setup, but utilise much smaller networks and other optimisations (e.g. concentrating on "difficult" training samples, i.e. samples near the decision boundary) to reduce the computational and memory footprint by up to two orders of magnitude w. r. t. other methods described in this section.

Several published methods concentrated on the uncertainty in the feature embeddings representing the facial images. Terhörst et al. [64] presented a method based on an arbitrary face recognition model. In the paper, pre-trained models made available by authors of FaceNet [57] and ArcFace [65] were used. The proposed method repeatedly extracts feature embeddings from random subnetworks of a face recognition model and computes variation across those using a simple distance measure. The magnitude of variation can be thought of as robustness of the underlying feature representation and indicates the image quality in an inverse relation. Chen et al. [50] and Chang et al. [66] utilise the concept of probabilistic face embeddings (PFE) first introduced by Shi et al. [67]. The concept allows to represent the facial images as (Gaussian) distributions in the latent space. Thus, the traditional deep-learning based feature representations are extended – the probabilistic embeddings use the means of the aforementioned distributions to represent the feature values, while their variances represent their uncertainties. The uncertainty values can be thought of as a measure of image quality. In fact, the experimental evaluations have shown that the image quality is directly correlated with said uncertainty measure, i.e. images with more degradations exhibit a higher degree of uncertainty in the probabilistic embeddings. A conceptually similar idea was presented by Meng et al. [68], who extended the ArcFace [65] loss function with additional parameters and constraints with an explicit goal to encode the face quality within the face feature representation magnitude. Under the new loss function, the high-quality samples tend to position themselves close to their respective cluster centres, as opposed to the standard ArcFace function where the position of samples within a cluster is more arbitrary, especially in unconstrained scenarios. In other words, the feature magnitudes correlate strongly with the cosine similarities with their respective cluster centres and can be thus directly used as an image quality measure. Liu et al. [69] presented a new loss function, which includes a parameter strongly correlated with image quality. They changed the network architecture and training protocol to enable output of both the face embedding and quality measure simultaneously.

In a very recent publication [70], Babnik et al. develop an algorithm FaceQAN for quality score computation based on the assumption that the difficulty of adversarial example generation is correlated with the quality of the image, i.e., that "high quality images are expected to produce stable and robust representations that are difficult to perturb using adversarial noise". Precisely, their method generates a set of adversarial examples for a given face recognition CNN and measures the distance of the embeddings (deep features) of these perturbed images with that of the original image. In order to penalize non-frontal poses, they also use the difference between the embeddings of the image and the horizontally flipped image. Their analysis shows that their approach, when applied to state of the art face recognition CNNs, outperforms other methods.

Explainability and interpretability (by humans) of automated algorithms has been an increasingly debated and researched topic in face recognition and other field where automated machine learning algorithms and models are used. Terhörst et al. [71] presented a method which assesses the utility of the images at pixel-level, thereby allowing to visualise its assessment using a heatmap overlaid on a facial image. Some of the surveyed methods can also support explainability. For instance, the computed uncertainty (with probabilistic face embeddings) has been visualised in [50]. In [72], the feature embeddings produced by several networks surveyed above were visualised using attention maps. Those examples notwithstanding, more research is needed in the field of explainable and interpretable deep learning-based facial image quality assessment algorithms, e.g. w. r. t. providing visualisations and actionable feedback during image acquisition.

### 3.3.3 Benchmarks

Among the face-specific methods, the relative performance of the methods varies depending on the combination of the test datasets and face recognition method, but currently three methods seem to generally outperform the others: MagFace [68], SDD-FIQA [60], PFE [67]. In the benchmarks presented within their respective publications, these three methods have generally outperformed other methods, notably SER-FIQ [64], PCNet [59], and FaceQNet [58], as well as various general-purpose methods. These findings were partially reproduced by [73], where MagFace clearly outperformed other methods, albeit SDD-FIQA and PFE were not included in the experiments. The benchmark of [73] is shown in the figure at the beginning of this section. For example, it can be seen, that the starting error rate (i.e. no images rejected) of the used face recognition algorithm on the test dataset is around 4%. For this specific combination of face recognition and test dataset, using MagFace for quality assessment, this error rate can be nearly halved once approximately 5% of the worst quality images have been rejected. It should be noted that the datasets used in the benchmarks contain different numbers of samples per subject. This fact has a non-trivial impact on the interpretation of the results, as with the exception of a simple experimental protocol where exactly one mated comparison is performed per subject yielding an optimum described by a linear relation between FNMR and rejection fraction, the achievable optimum performance improvement at a given fraction of rejected images is not clearly determinable. Furthermore, there currently exists no methodology to quantitatively ascertain the generalisation capabilities of quality assessment algorithms across multiple face recognition systems and test datasets.

The aforementioned three highly performant methods have not been benchmarked against each other within their respective publications. The benchmark conducted by authors of CR-FIQA [61] confirms the efficacy of the three aforementioned methods; the new method proposed in that paper appears to be on par with or even outperform these in most experimental settings. To date, the most comprehensive benchmark of face image quality assessment methods has been published by Fu et al. [72]. They evaluated 25 quality assessment algorithms used in conjunction with 3 face recognition algorithms on 3 well-known facial image datasets. They evaluated handcrafted and deep learning-based general-purpose and face-specific image quality algorithms. In the benchmark, the deep learning-based face-specific methods generally outperformed the general-purpose methods, albeit, as mentioned previously, the general-purpose methods have proven quite capable of face image quality assessment. In the benchmarks, the performance of MagFace, SDD-FIQA, and PFE have obtained very similar results in terms of ERC and generally outperformed other methods. This benchmark does not, however, include the CR-FIQA method.

First evaluations of the very recent FaceQAN indicate that this method performs even better than the aforementioned algorithms [70] [74]. However, the generation of the adversarial examples is very time consuming because each of them is optimized in an iterative process consisting of five forward passes through the face recognition CNN followed by a gradient computation. Experiments conducted without CUDA on a Core i9 10900X 3.7GHz computer revealed in running times of over two minutes per input image.

It should be noted that the benchmarks described above have been conducted on highly unconstrained datasets containing in-the-wild and automatically scraped images. Although the surveyed methods have been shown to be capable of assessing the quality of such images (some of which do not even contain faces), their usefulness in a more constrained context (e.g. nearly ICAO-compliant images) is yet to be rigorously demonstrated. More concretely, the surveyed quality assessment algorithms appear to perform very well in the domain they were trained on, i.e. unconstrained images, and are capable of reliably sorting out the images of poorest quality (e.g. no faces at all, very strong occlusions, massive blur or lack of sharpness etc.). On the other hand, it remains to be demonstrated whether or not they would be able to reliably distinguish between and quantify small imperfections present in nearly ICAO-compliant images.

In addition to neural network-based approaches, methods of fusing handcrafted image quality features can be considered in the context of unified quality score. Several handcrafted features (according to ISO/IEC 29794-5:2010, e.g. image sharpness and blur, lighting symmetry, inter-eye distance, etc.) were considered in the benchmark of [72]. The results indicate that many of those features are strongly correlated with each other, and that individual features possess only limited predictive power w. r. t. general facial image quality assessment; a finding also present in the benchmark of individual handcrafted features conducted by Henniger et al. [75] on a different dataset of facial images. Unfortunately, these two benchmarks did not consider a fusion of these features. In Khodabakhsh et al. [76], a similar benchmark is conducted on a dataset of images captured with a mobile device and includes a fusion of the features, which is shown to perform slightly better than any of the individual handcrafted quality metrics. The results of these benchmarks of handcrafted features cannot be compared with those of general-purpose and face-specific methods described earlier, as different facial image datasets were used. Generally speaking, the fusion of handcrafted features for the purpose of facial quality assessment appears to be an insufficiently researched topic. Recently, an implementation of the handcrafted features considered by ISO/IEC 19794-5:2011 has been published by Hernandez-Ortega et al. [77]; however, a benchmark in terms of ERC is yet to be conducted.

## 3.4 Software

| Name | Model size | License |
|------|------------|---------|
| MagFace[99] | 255 MB / 171 MB / 94 MB (IResNet100, IResNet50, IResNet18) | Apache 2.0 |
| EQFace[100] | 290 MB | MIT |
| Ex-FIQ[101] | 373 MB | CC BY-NC 4.0 |
| SDD-FIQA[102] | 118 MB | Apache 2.0 |
| LightQNet[103] | 2 MB | MIT |
| ProbFace[104] | 179 MB | MIT |

---

[99] https://github.com/IrvingMeng/MagFace
[100] https://github.com/deepcam-cn/FaceQuality
[101] https://github.com/pterhoer/ExplainableFaceImageQuality
[102] https://github.com/Tencent/TFace/tree/quality
[103] https://github.com/KaenChan/lightqnet
[104] https://github.com/KaenChan/ProbFace

| Name | Model size | License |
|---|---|---|
| CR-FIQA[105] | 166/249 MB | CC BY-NC 4.0 |
| SER-FIQ[106] | 232 MB | CC BY-NC-SA 4.0 |
| FaceQNet[107] | 91 MB | No license, non-profit use mentioned in GitHub issues |
| PFE[108] | 211 MB | MIT |
| FaceQAN[109] | 260 MB for IResNet100 face recognition model from InsightFace | No license |

*Table 11: Recent face-specific image quality assessment methods*

| Name | Model size | License |
|---|---|---|
| UNIQUE[110] | 85 MB | Apache 2.0 |
| DBCNN[111] | 1 GB | MIT |
| PIQ Toolbox[112] | n. a. | Apache 2.0 |
| NIMA[113] | 13 MB | Apache 2.0 |
| RankIQA[114] | 512 MB | MIT |

*Table 12: Recent general-purpose image quality assessment methods*

---

[105] https://github.com/fdbtrs/CR-FIQA
[106] https://github.com/pterhoer/FaceImageQuality
[107] https://github.com/uam-biometrics/FaceQnet
[108] https://github.com/seasonSH/Probabilistic-Face-Embeddings
[109] https://github.com/LSIbabnikz/FaceQAN
[110] https://github.com/zwx8981/UNIQUE
[111] https://github.com/zwx8981/DBCNN
[112] https://github.com/photosynthesis-team/piq
[113] https://github.com/idealo/image-quality-assessment
[114] https://github.com/xialeiliu/RankIQA

# 4   Background Uniformity

The aspect background uniformity belongs to the capture related quality elements of an image. According to ISO/IEC WD4 29794-5:2022, the background of an image is the scenery or the area in the image that is located behind the subject. Thus, hair, neck, parts of the upper body and clothing visible in the image, or head coverings are not part of the background, and consequently, the background is not obtained by a segmentation of the face. Uniformity means that the brightness and colours in the background do not vary much, in particular, on small spatial scales; however, small and gradual changes of brightness and colour in one direction over the entire image (e.g. from left to right) do not necessarily conflict with the uniformity requirement.

In ISO/IEC WD4 29794-5:2022, background uniformity is only mandatory for reference images for ID documents. Since reference images for ID documents are typically captured in highly controlled environments like photographic studios or booths, the uniformity of the background can be easily achieved by a capture environment where the subject is placed in front of a uniformly coloured surface (e.g. a screen or wall). For other enrolment scenarios and for the collection of probes, the quality aspect is optional in ISO/IEC WD4 29794-5:2022, presumably due to the fact that, there, it is much harder to influence the background. For instance, for self-service kiosks at borders (for instance, for EES enrolment), constructional constraints and performance requirements often make it impossible to place the kiosk vis-à-vis a screen or wall.

Likewise, ISO/IEC 39794-5:2019 requires a uniform background only for reference face images for *machine readable travel documents* (MRTD). For such images, it is required that the background should have a smooth, uniform texture/colour and the boundary between the head and the background should be clearly visible around the entire subject.

On the other hand, ISO/IEC 19794-5:2011 recommends for all frontal face images a plain background without texture containing lines or curves.

Examples of a uniform background in images captured under controlled environments can be found in in the following figure.

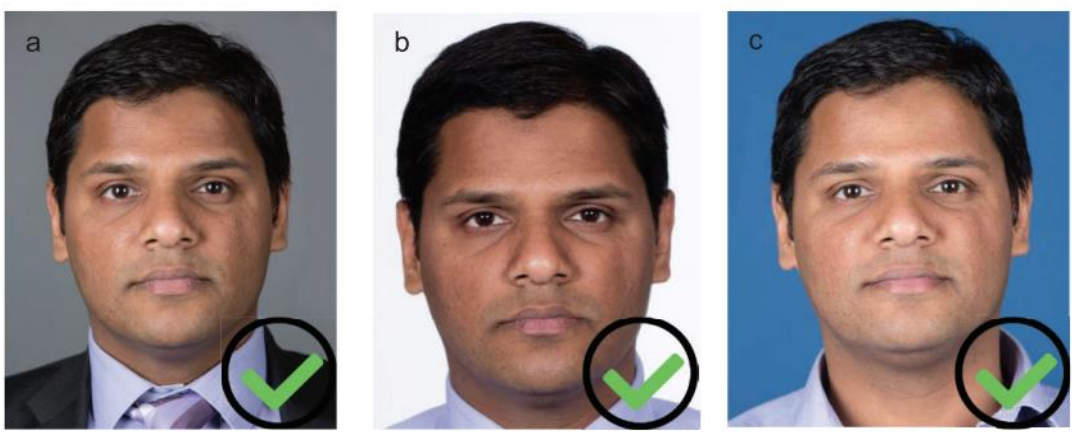

*Figure 8: Example of uniformity backgrounds from ISO IEC 39794-5:2019 D.1.4.2.5*

## 4.1 Impact on Face Recognition

State of the art CNNs for face recognition take as input images cropped closely to the face area, so that eyes, mouth and nose are located at pre-defined positions. Therefore, the background does not have any significant influence on the recognition performance.

A potential issue of a non-uniform background could be that the detection and segmentation of the face can fail. In fact, ISO/IEC 19794-5:2011 motivates its recommendation of a uniform background for frontal face images with potential confusions of face detectors by lines and curves in a non-uniform background. However, since 2001, the robustness and accuracy of face detectors has greatly improved. Current state of the art face detectors reliably detect all faces of sufficient size even in challenging in-the-wild images. Furthermore, even if methods mistake background textures for a face, the subject's face can still be reliably located by choosing the largest of the detected face boxes. Therefore, the reasoning of ISO/IEC 19794-5:2011 seems outdated, at least, if state of the art face detectors are used; nevertheless, since conformance to ISO/IEC 19794-5:2011 is required by the EU Commission Implementing Decision 2019/329, this quality element is still relevant for EES.

On the other hand, a uniform background can facilitate the visual inspection of a reference image and its comparison to the subject by officers.

## 4.2 Datasets

There are no existing datasets with face images and labels specifying the uniformity of the background. However, there are datasets with labels for the capture environment, where some environments ensure uniform background while others result in non-uniform backgrounds. In these datasets, however, the variance of the background textures can be limited due to the defined environments.

| Name | Subjects | Images | Remark | License |
|------|----------|--------|--------|---------|
| FRGCv2[55] | 569 | 45,000 | The background uniformity can be inferred from the environment ID. | Royalty-free |

*Table 13: Summary of available datasets eligible for evaluating the assessment of background uniformity*

## 4.3 Methods and Algorithms

In ISO/IEC WD4 29794-5:2022 a method for evaluating the uniformity of the background is specified as follows: First a face segmentation is performed, by detecting the face and determining the face area. Then, the region extending horizontally from the left side (border) of the image to the subject's right ear[115] and vertically from the top to the centre of the image of the image is computed. Similarly, the region extending horizontally from the right side to the subject's left ear and vertically from the top to the centre of the image of the image is computed. On the join of these two regions, the entropy of the luminance histogram is computed as a measure of uniformity.

---

[115] Here, "right" and "left" refer to the subject's perspective. In the image, the right ear is on the left side of the face, and vice versa.

However, this method has some limitations: Firstly, it assumes that the ear's location can be determined. However, there is no facial landmark localization algorithm (see Section 2.2) that outputs landmarks for the ears, which is likely due to the fact that the ears are often occluded by hair. Therefore, the location of the ears would have to be interpolated from other facial features, e.g. the eyes and the face contour. Secondly, the defined regions can also contain parts of the subject's hair, e.g. in case of voluminous hairstyles, or head coverings. This may result in wrong assessments of the algorithm. Thirdly, the regions specified in the algorithm cover only the upper half of the image, so that non-uniform textures in the background below the ears cannot be detected.

A more reliable and accurate way to assess the uniformity of the background would make use of a method for segmentation of all visible parts of the subject and not just of face. For instance, in [78], Lee et al. proposed a CNN that performs semantic segmentation of the subject if facial images and distinguished various parts of the subject including hair, neck and clothing. Alternatively, a method for so-called *matting*, the segmentation of salient parts in the image, can be applied. A vast number of methods have been proposed for this task, surveys of which are given in [79] and [80].

Given a reliable segmentation of the subject, the uniformity of the background could be estimated by the entropy of luminance or colour intensity values, similarly to the method specified in ISO/IEC WD4 29794-5:2022. Alternatively, the variance or the range of the values could be used.

## 4.4    Software

No implementations could be found that assesses the uniformity of the background in facial images. However, as pointed out at the end of the previous section, the main challenge of implementing such an algorithm is the segmentation of the subject. The following table lists the most relevant free software implementation available for subject segmentation or matting/background removal.

| Name | Method | Model Size | License |
|---|---|---|---|
| ldbm-image-background-remover[116] | Matting | n. a. | MIT |
| Rembg[117] | Matting | 4 MB (general purpose model) 168 MB (human segmentation) | MIT |
| PyMatting[118] | Matting | n. a. | MIT |
| CelebAMask-HQ Face Parsing[41] | Subject Segmentation | 352 MB | Non-commercial |
| face-parsing.PyTorch[42] | Subject Segmentation | 51 MB | MIT |
| face-seg[44] | Face and hair Segmentation | 14 MB | No license |

*Table 14: Examples of software available for the segmentation of fore/background.*

---

[116] https://github.com/whitelok/ldbm-image-background-remover
[117] https://github.com/danielgatis/rembg
[118] https://github.com/pymatting/pymatting

# 5     Illumination

In terms of face recognition performance, a well-lit face in an image is necessary to achieve good results. For this purpose, one should not only aim for a suitable distribution of the brightness values, but sufficient colour differences should also be present as well, because only then certain facial properties, such as pigment spots on the skin become clearly visible. In addition, the distribution of brightness values not only depends on the illumination but also the amount of light the face is exposed to.

Different effects can result in images with poorly-lit faces. For example, the illumination of the face can be too dark or too bright for the duration of exposure. In such a case, the resulting image will be under- or overexposed, respectively, and even after adjusting the brightness using an image editing software, the contrast would be low so that certain contours of the face in the image could be lost and thus no longer be recognizable. Insufficient contrast may also result from inappropriate devices, material or settings used in printing or scanning of the facial images. Note, that increasing the contrast in a low-contrast image does not remedy the problem, as it increases the noise and results in random textures that could be mistaken as facial properties by face recognition software. Another potential problem can be shadows on parts of the face or very bright areas in the face region, e.g. resulting from a single light source on one side of the face or a light source that is too close to the face.

The distribution of the brightness values within the face area, and hence the visibility of the facial properties, can also depend on the skin colour of the subject. For instance, textures like wrinkles or moles are typically less visible on images of faces with very dark skin as compared to light skin. However, the objective of face image quality assessment[119] is not to evaluate the subject's eligibility for face recognition technology, but to determine if the image has been appropriately captured and, potentially, to ensure that another, improved capture is taken in case of insufficient quality. Furthermore, biometric systems deployed in border control scenarios must not be discriminatory with respect to ethnicity or other genetic factors influencing the skin tone. Thus, the quality metrics used for assessing the face image quality with respect to illumination must ensure that, for all skin colours, images can be captured that give satisfactory outputs.

For the assessment of the illumination of facial images, many metrics have been proposed, many of which are based on similar concepts. ISO/IEC WD4 29794-5:2022 tries to divide those into individual metrics, which are:

- Moments of the luminance distribution
- Exposure
- Dynamic Range
- Spatial uniformity of the illumination

---

[119] Note that in Clause 4.2 of ISO/IEC 29794-1 the facial biometric characteristic is defined as "contributor to quality of a sample attributable to inherent properties of the source"

All these metrics are based only on the brightness or luminance values of the pixels in the face region of the image. Pixels outside the face region are not relevant for the assessment of illumination aspects of facial images.

**Statistical moments of the luminance distribution**

Typical measurements of general image illumination (not restricted to facial images) are the brightness and the contrast of an image. These do not have a uniform definition, but often relate to the terms "mean" and "variance" of the pixel intensity values [81]. Therefore, one way to measure the illumination of an image is the concept of moments. In mathematics, the moments of a function are quantitative measurements that relate to the shape of the graph of a statistical (probability or frequency) distribution function.

These concepts can be applied to the luminance histogram of a given image. This requires converting the image from the typical RGB format to a format that represents the luminance channel, e.g. to YUV, where Y represents the luminance channel.

Therefore, the first moment (mean) relates to the brightness and the second moment (variance) to the contrast of an image. However, the mean and variance alone provide only limited information on the shape of histogram. For this reason, higher moments such as the third and fourth moment can also be considered. The third moment describes the skewness of the function, which can reveal the asymmetry of the distribution. An example for asymmetry in an image is a homogenous illumination of the face, but poor exposure, resulting in higher frequencies of dark values. The fourth moment deals with the kurtosis of the function, which quantifies how tail-heavy the distribution is and can be considered as a measure of the "peakedness" of the curve.

ISO/IEC WD4 29794-5:2022 specifies all 4 moments (mean, variance, skewness and kurtosis) of the luminance histogram of the face region as quality metrics.

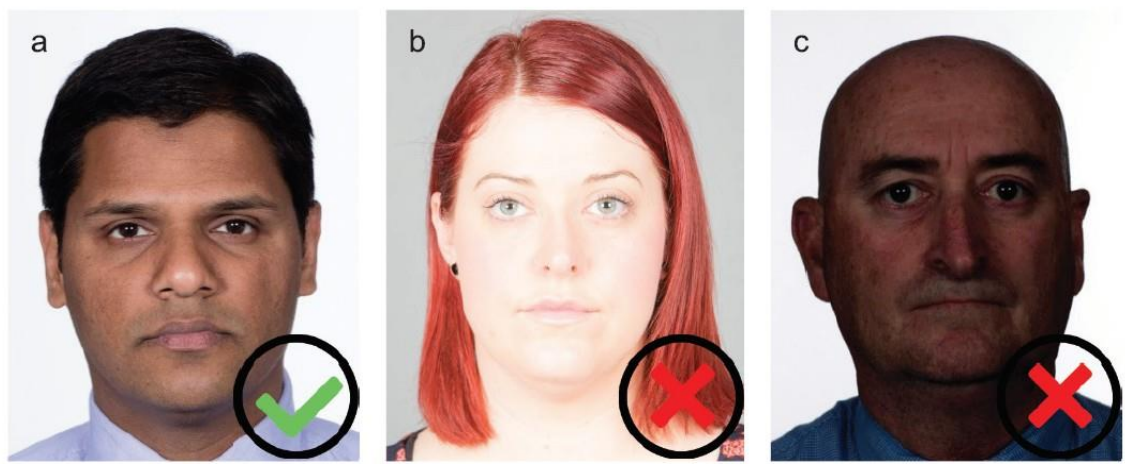

**a) Correct exposure**     **b) Over exposure**     **c) Under exposure**

*Figure 9: Examples of correct (a) and incorrect (b and c) exposure (taken from ISO/IEC 39794-5:2019)*

**Over-Exposure, Under-Exposure**

ISO/IEC WD4 29794-5:2022 defines two metrics, Over-Exposure and Under-Exposure, both of which are based on the luminance histogram. Precisely, these metrics simply compute the fraction of pixels in the face region that have a very low luminance value (smaller than 8 on a scale from 0 to 255) or a very high luminance value (higher than 246 on a scale from 0 to 255), respectively. Thus, only the leftmost (darkest) and rightmost (brightest) parts of the luminance histogram are considered to determine the exposure of the face. Figure 9 displays example images from ISO/IEC 39794-5:2019 for under-exposure and over-exposure as well as correct exposure.

**Dynamic Range**

ISO/IEC WD4 29794-5:2022 also lists the metric "Dynamic Range", which, in general, is a way to describe the range of luminance values of an image; this range can be restricted by limitations of the digital camera or film used for capture. Compared to the metric "Exposure", where only the leftmost and rightmost ends of the luminance histogram are considered, this metric considers the whole range of the histogram and its variation.

This metric and its proposed variations (see Section 5.3) have similarities with the concept of contrast of an image. This demonstrates the overlapping nature of the various metrics for the assessment of the illumination of an image. ISO/IEC 39745-5:2019 and ISO/IEC 19794-5:2011 recommend that the dynamic range, therein called *Grayscale Density*, of the image should comprise at least 7 bits of intensity variation (i.e. to span a range of at least 128 unique values) in the face region of the image. TR-03121-3 Vol. 1 requires, under *Grey scale density and colour saturation*, to measure the number of intensity values existing within the image.

As seen in the Figure 10, a face with low dynamic range can have similar traits as an under-exposed face but also an over-exposed face, whereas a face with high dynamic range is considered well illuminated.

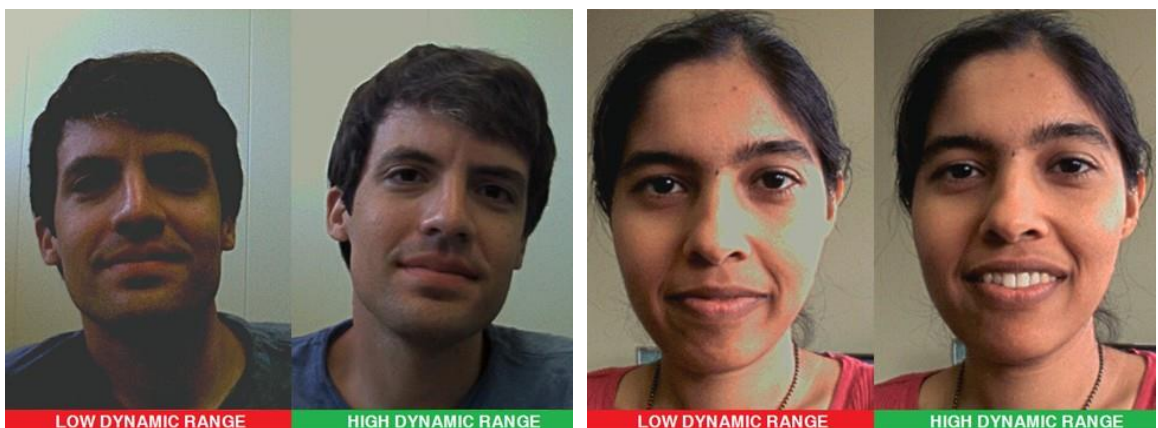

*Figure 10: Examples of faces with low and high dynamic ranges (taken from https://vivonets.ece.ucsb.edu/realtimeHDR.html )*

**Non-uniform illumination**

All of the aforementioned requirements focus only on the distribution of the intensity or luminance values of the face region, but do not explicitly take the spatial distribution into account. However, spatially non-uniform distributions of the illumination, i.e. too bright or too dark areas (e.g. shadows), can result in an obscuration of that the facial biometric characteristic in partial areas.

ISO/IEC WD4 29794-5:2022, but also in ISO/IEC 39794-5:2019, ISO/IEC 19794-5:2011 and TR-03121-3 Vol. 1 require a spatially uniform illumination. While ISO/IEC WD4 29794-5:2022 only requires to measure if the lighting is horizontally uniform, i.e. the difference of the illumination between the left and the right side of the face, ISO/IEC 19794-5:2011 also requires that the lighting is vertically uniform, and ISO/IEC 39794-5:2019 requires uniform lighting between 4 pre-defined regions of the face (left and right cheek, chin and forehead). TR-03121-3 Vol. 1 requires a generally uniform lighting, but also that there are no shadows over the face nor in the eye sockets. ISO/IEC 39794-5:2019 gives more details on the latter aspect by requiring that *there shall not be extreme dark shadow visible on the face, especially around the nose, in the eye sockets, around the mouth, and between mouth and chin that obscure face details important for inspection*. ISO/IEC 19794-5:2011 requires as well that there are *no shadows in the eye-sockets due to the brow*. Examples of inadequately illuminated images are shown in

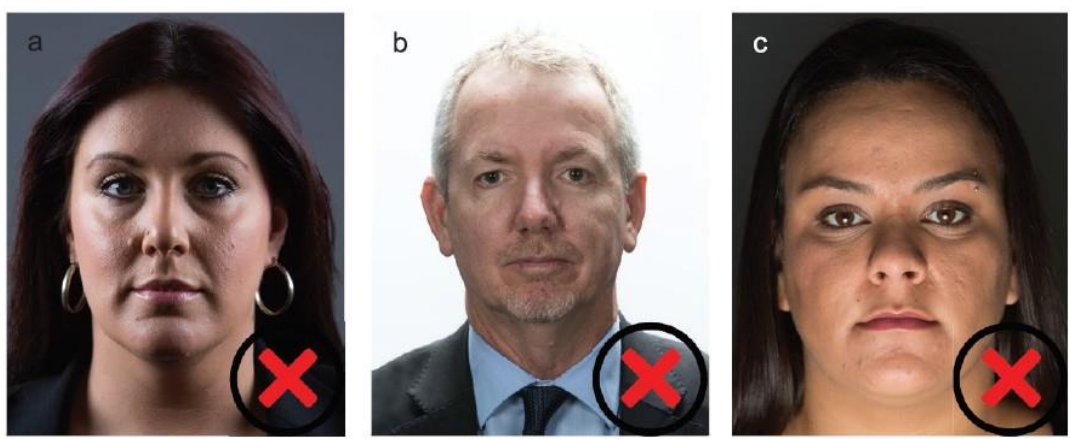

a) Side illumination     b) Top illumination     c) Bottom illumination

*Figure 11: Examples of spatially non-uniform illumination (taken from ISO/IEC 39794-5:2019)*

The definitions of Over-Exposure and Under-Exposure in ISO/IEC WD4 29794-5:2022 can also detect images where only some areas are too bright (hot spots) or too dark. ISO/IEC 39794-5:2019, ISO/IEC 19794-5:2011 and TR-03121-3 Vol. 1 explicitly require that there are no hot spots, i.e. bright regions that results from light shining directly on the face.

## 5.1    Impact on Face Recognition

It is obvious, that poor lightning or bad exposure can have negative effects on face recognition. A face too bright or with too strong contrast can lead to bad exposure. With an under-exposed face, many elements of the facial biometric characteristic (eyes, nose, mouth) can be obscured by the shadows on the face. As a result, trying to extract features from those elements can be quite difficult. Over-exposure creates a similar problem if the lighting creates very bright areas on the face, that obscure the facial properties or contours describing the properties. The same detection problems occur if the dynamic range of the face is too low.

The impact of illumination has been mentioned in several studies as a cause of poor facial image recognition. For instance, in [82], Beveridge et al. mention that changes in illumination rank high on most researcher's list of factors making face recognition difficult. Datasets such as CMU Multi-PIE[120] and YaleB[121] were created in 2008 and 2005, respectively, to study changes in illumination and also pose. Since then, the advent of CNNs have resulted in great improvements of the ability of face recognition algorithms to deal with poor illumination conditions.

In [83], Gazi et al. investigate the influence of illumination (among other factors) on the accuracy of two CNN-based face recognition algorithms. On the Extended Yale Dataset they found a significant drop of accuracy already for moderately impaired illumination (subset 3) and a drastic decrease for poor illumination (subsets 4 and 5). For the poorly illuminated images, a pre-processing by illumination normalization and contrast enhancement resulted in a considerable improvement, but the accuracy was still much lower than for good illumination conditions.

In [84], Grm et al. found that both an increase of the brightness and a decrease of the contrast considerably decreases the recognition performance of CNNs trained on face recognition.

In [85], Dooley et al. conducted an evaluation of three recent commercial face detection algorithms (from Amazon, Microsoft and Google) and found that images captured under poor illumination result in significantly higher error rates. Since face detection is typically a necessary pre-processing for face recognition, this result also implies an impact of illumination on face recognition.

---

[120] https://cmu.flintbox.com/technologies/67027840-27d5-4570-86dd-ad4715ef3c09
[121] http://vision.ucsd.edu/~iskwak/ExtYaleDatabase/ExtYaleB.html

                                          

## 5.2 Datasets

The following table summarizes several available datasets containing face images with different illumination conditions.

| Name | Subjects | Images | Remark | Constrained? | License |
|------|----------|--------|--------|--------------|---------|
| FIIQD[122] | unknown | 224,733 | Various illumination patterns with 3 labels for the associated illumination quality (good, mediocre, poor). | Yes | No license |
| Exended Yale-B[121] | 28 | 16,128 | Includes pose and illumination. 64 illumination conditions. Various azimuth and elevation angles of a single light source direction. | Yes | Research purposes |
| CMU Multi-PIE[120] | 337 | 750,000+ | Includes pose, expression and illumination. 19 illumination conditions, 17 different azimuth angles of a single light source direction. | Yes | Royalty-bearing |
| IMFDB[123] | 100 | 34,512 | Includes pose, expression, illumination, age, occlusion, and makeup. 3 illumination labels (good, mediocre, poor). | No | No license |
| Wider-Face[124] | 393,703 | 32,203 | Includes pose, occlusion, expression, illumination and makeup. 2 illumination labels (0 normal and 1 extreme). | No | No license |
| AR Face[125] | 126 | 4,000+ | Includes 3 illumination conditions (left side light, right side light, all sides light), along with occlusions and various expression. | Yes | Academic use only |
| SoF[126] | 112 | 42,592 | Includes Illumination and (synthetic) Occlusion. The label "illuminationQuality" indicates well-illuminated (all non-occluded facial properties are recognizable by naked-eye) or poor illumination conditions (at least one facial point, i.e. landmark, is invisible due to bad illumination conditions) | No | Royalty-free for any educational purpose |

---

[122] https://github.com/zhanglijun95/FIIQA
[123] https://cvit.iiit.ac.in/projects/IMFDB/
[124] http://shuoyang1213.me/WIDERFACE/
[125] https://www2.ece.ohio-state.edu/~aleix/ARdatabase.html
[126] https://sites.google.com/view/sof-dataset

| Name | Subjects | Images | Remark | Constrained? | License |
|------|----------|--------|--------|--------------|---------|
| EURECOM Visible and Thermal Paired Face Database[127] | 50 | 2,100 | Includes Expression, Pose, Occlusion and Illumination. 5 pairs captured with frontal pose and different illuminations: Ambient light, rim light, key light, fill light, all lights on, all lights off | Yes | Non-commercial use |
| CAS-PEAL-R1[128] | 1040 | 99,594 | Includes Expression, Pose, Illumination, Occlusion. Different illumination through positioning of lamps at 5 azimuths (-90°, -45°, 0°, +45°, +90°) and 3 elevations (-45°, 0°, +45°) | Yes | Non-commercial use |
| FRGCv2[55] | 568 | 49,228 | The illumination conditions can be inferred from the ID of the capture environment. Some environments ensured virtually perfect illumination while others were others used ambient indoor lighting. | Yes (mainly) | Royalty-free |

Table 15: Summary of available datasets including illumination and exposure

---

[127] http://vis-th.eurecom.fr/
[128] http://www.jdl.link/peal/index.html

                                          

## 5.3    Methods and Algorithms

ISO/IEC WD4 29794-5:2022 defines several metrics for assessing the illumination/exposure of the facial image. All of these methods are based on the computation of the luminance histogram in the region of the face. Thus, the algorithms for the computation of these metrics share the first processing steps and differ only in the computations performed on the histogram:

1. Localization and segmentation of the face region of an image.
2. Extraction of the luminance channel of the face region.
   a. Non-uniform illumination: Find the line through the mid-point between the eye centres and perpendicular to the line through the eye centres.
3. Calculation of the luminance histogram.
   a. Non-uniform illumination: Calculate two histograms for the left and right halves of the face.
4. Computation of the individual metric.
   a. Mean (Brightness): First moment of the luminance distribution curve.
   b. Variance (Contrast): Second moment of the luminance distribution curve.
   c. Skewness: Third moment of the luminance distribution curve.
   d. Kurtosis: Fourth moment of the luminance distribution curve.
   e. Under-exposure: Sum of the pixels of the histogram in the interval [0;7].
   f. Over-exposure: Sum of the pixels of the histogram in the interval [247;255].
   g. Dynamic Range: Entropy of the histogram.
   h. Non-uniform illumination: Histogram intersection of the grey-level histograms of the left and right halves of the face region

Besides the methods specified in ISO/IEC WD4 29794-5:2022, other metrics and algorithms for the assessment of illumination of facial images have been proposed in the scientific literature.  These approaches are not completely disjoint, but they overlap with each other, as they aim to detect similar deficits of illumination.

A method to assess the exposure of the face is used by Wasnik et al. in [86]. An exposure of an image is calculated with the use of the *Absolute Central Moment (ACM)* as a basis. The ACM can be described as a matrix in the shape of the given image. After converting the image to grey scale, the mean pixels value is computed, and the absolute difference between the grey value of each pixel and the image mean defines each entry of the ACM matrix. Afterwards, the mean of the ACM matrix is calculated, which defines the exposure of the image. It is important to mention, that this exposure implementation is adopted from the published ISO/IEC TR 29794-5:2010, and is, thus, a previous version of the metric from the current WD described above. Furthermore, they deal with "lighting symmetry", which can be viewed as non-uniform illumination.  The proposed calculation method is also adopted from the published ISO/IEC 29794-5:2010, where the asymmetry can be expressed based on the histogram of pixel values in each sub-area (left and right).

In [87], Hernandez-Ortega proposes a method to detect if the facial image is too dark or too bright, solely "based on the mean pixel value". This approach bears resemblance to the brightness evaluation in ISO/IEC WD4 29794-5:2022, but the authors provide no details, in particular, if the "pixel value" refers to all colour (RGB) channels, the greyscale intensity or the luminance, and over which region the mean is taken. Also, their referenced Github project page[129] is empty at the time of this writing.

In [88], Fourney and Laganiere describe a method, which makes use of the available dynamic range of an image to measure the quality with respect to illumination. At first, the *utilization* U is computed by determining the smallest range of grey scale intensities to which at least 95% of an image's (not just the face region's) pixels can be attributed. Afterwards, the score of quality is computed as the percentage of the total dynamic range. In an 8-bit grey scale image this would be U/256. In addition, Fourney and Laganiere define a measure for the left-right uniformity of the illumination by computing the integral (area) of the intersection $L \cap R$, where L and R are the grey-scale intensity histograms of the left and right halves of the face, respectively. A similar method is specified in ISO/IEC WD4 29794-5:2022, where the intersection of the grey-level histograms of the left and right halves of the face is computed to assess the (spatial) uniformity of the illumination.

---

[129] https://github.com/uam-biometrics/FaceQvec

In [89], Kim et al. propose a new automated face image quality assessment with three quality factors, one of which is *Brightness* ($d_3$). Their method to measure the brightness makes use of the observation that too bright or too dark faces may have one-sided histograms and faces with moderate brightness may have relatively evenly distributed histograms. Thus, they compute brightness as the relative entropy between the histogram of the given face image $h_j$ and a reference histogram $h_{ref}$, where the brightness is considered optimal. The relative entropy is calculated as follows:

$$d_3 = \sum_k h_j(k) \log(h_j(k)/h_{ref}(k))$$

This algorithm is similar to the method specified for the metric "dynamic range" in ISO/IEC WD4 29794-5:2022.

In [90], Zhang et al. propose a CNN *FIIQA_DCNN*, (an abbreviation for *Face Image Illumination Quality Assessment*) to predict the illumination quality of face images. Their model was trained on their own established "Face Image Illumination Quality Dataset", in short FIIQD, containing 224,733 face images with various illumination patterns. As target values for training, the manually assigned to each image a subjective illumination quality score. The performance was evaluated by means of the correlation between the predicted values and the subjective scores using both the Spearman rank-order correlation coefficient (SROCC) and the Kendall rank-order correlation coefficient (KROCC). The model achieved a SROCC of 0.9477 and a KROCC of 0.8915.

## 5.4    Software

Open-Source image processing software, such as Imagemagick[130] or OpenCV[131] provide functions to compute the histogram of certain channels of the given image, preferably the luminance channel, with a few lines of code. In order to apply a face segmentation beforehand, software for facial landmark localization can be deployed (see Section 2.3.3), and the face region can then be defined as the convex hull of the landmarks.

However, there are already complete implementations for the assessment of face illumination or exposure freely available. Two kinds of implementation were found, those based on algorithmic approaches and those who rely on deep CNNs. Therefore, these will be listed in separate tables.

The following table provides information about repositories found on GitHub which implement face illumination or exposure through algorithmic approaches.

---

                                                        

| Name | Method | Remark | License |
|------|--------|--------|---------|
| image-checker[132] | Gives a score for under exposure, over exposure and low contrast via histogram. | | MIT |
| face-quality-metrics[133] | Estimates the quality of illumination by determining the length of available dynamic range of grey intensities excluding 5% of the darkest and brightest pixels. | Paper: "Face Quality Assessment for Face Verification in Video" [91] | MIT |
| AutoExposureChecker[134] | Lower bound threshold of the average pixel value of the image (from 0 - 1). Anything below this value will be considered underexposed (Default threshold: 0.25). Upper bound threshold of the average pixel value of the image (from 0 - 1). Anything above this value will be considered overexposed (Default threshold: 0.75). | | MIT |
| face-image-quality-29794-5-2010[135] [86] | Exposure estimation: The implementation measures the exposure by averaging the absolute value of each image pixel subtracted by the image's pixel mean.<br><br>Illumination uniformity: The implementation estimates the illumination uniformity from the difference of the left and right side of the facial image.<br><br>The implementation also provides further functions for estimating image properties, for example *image brightness, image contrast, perceived contrast, etc.* | | Commercial use |

*Table 16: Summary of relevant GitHub Repositories regarding illumination (algorithmic)*

In ISO/IEC WD4 29794-5:2022, a Python code to compute the (spatial) uniformity of the illumination is included in Annex C. This implementation is based on the face detector and the facial landmark localization of dlib (see Sections 2.1.3 and 2.3.3).

The following table provides information about repositories found on Github which implement face illumination or exposure through the use of CNNs.

| Name | Dataset | Model size | Metrics | License |
|------|---------|-----------|---------|---------|
| FIIQA[122] | FIIQD | 91 MB | SROCC: 0.9477 KROCC: 0.8915 | No license |
| FIIQA-PyTorch[136] | FIIQD | 6 MB | SROCC: 0.9477 KROCC: 0.8915 | No license |
| Expression_Recognition[137] | IMFDB | 0.3 MB | IMFDB Split: Train (80%), Test (10%), Valid (10%) Accuracy: 0.77 | No license |

*Table 17: Summary of relevant Github Repositories regarding illumination/exposure (CNN)*

---

[132] https://github.com/mrseanryan/image-checker
[133] https://github.com/Alireza-Akhavan/face-quality-metrics
[134] https://github.com/amcolash/AutoExposureChecker
[135] https://share.nbl.nislab.no/g03-03-sample-quality/face-image-quality
[136] https://github.com/yangyuke001/FIIQA-PyTorch (essentially an implementation of FIIQA in PyTorch)
[137] https://github.com/Moado/Expression_Recognition

# 6    Image Sharpness Aspects

This section summarizes different interrelated quality aspects listed in ISO/IEC WD4 29794-5:2022 which are related to *image blur*. Generally speaking, there exist two types of blur which can occur when capturing a (facial) image, examples of which are shown in Figure 12:

- De-focus: blurring of an image due to incorrect focus adjustment of the imaging system.
- Motion blur: blurring of an image due to movement of the subject or imaging system.

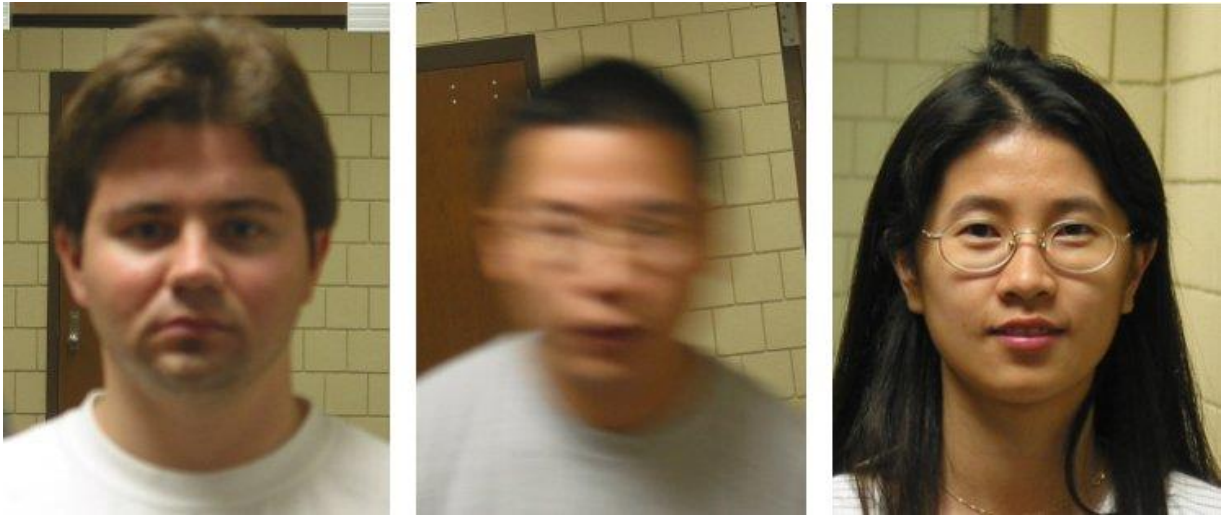

*Figure 12: Examples images (from the dataset FRGCv2) with de-focus (left), motion blur (middle), and a sharp face image (right).*

Of course, images can also be blurred after capture using an image editing software, e.g. using median filter of Gaussian blur filter.

*Sharpness* refers to an image's overall clarity in terms of both focus and contrast. In the presence of blur, the sharpness of (parts of) the image is reduced. Further, edge density may be defined as the amount of edges in an image which may show the concentration of high frequency information. Blur is fading the edges present in an image which reduces its *edge density*. It is important to note that edge density is also affected by illumination [92]. Hence, the state of the art in detecting de-focus, sharpness, motion blur, and edge density are jointly surveyed in this section.

## 6.1    Impact on Face Recognition

The presence of blur in facial images can negatively impact face recognition systems since facial properties are vanishing depending on the blur intensity. Hence, face detection becomes more challenging and the extraction of discriminative facial features becomes more difficult. As a consequence, the biometric performance of face recognition systems is expected to degrade. This effect has been showcased in different scientific works.

For instance, Punnappurath et al. [93], Hua et al. [94] and Grm et al. [84] demonstrated that face recognition performance significantly decreases with high blur intensity. This has been confirmed in a comprehensive benchmark recently proposed by [72]. Also, it has been shown that the aforementioned quality factors relating to blur strongly correlate, e.g. focus and edge density [95].

Said findings motivated researchers to attempt to develop blur-invariant face recognition systems. Increased robustness to blur may be achieved by adapting feature extractors to extract coarse features which are (to a certain extent) more robust to blur, e.g. in [93], or by re-sharpening blurred face images prior to the feature extraction step, e.g. in [84]. However, these concepts have achieved only limited success. In particular, the latter concept which is commonly referred to as face hallucination has been shown to be ineffective for face recognition purposes [84]. Therefore, it is expected that a detection and subsequent rejection of blurred face images improves the biometric performance of a face recognition system.

## 6.2    Datasets

For evaluating the impact of blur on face recognition performance, datasets containing facial images with ground truth labels, i.e. type of blur and blur intensities, are required. Different datasets have been collected which contain annotated blurred face images, see Table 18.

| Name | Subjects | Images | Remark | Constrained? | License |
|---|---|---|---|---|---|
| FDDB[7] | n. a. | 2,845 | environmental influences like snow, fog, rain | No | Academic use only |
| UCCS[138] | 1,085 | n. a. | - | No | No license[139] |
| UFFD[12] | n. a. | 6,424 | - | No | Academic use only |
| Wider-Face[5] | 393,703 | 32,000 | - | No | No license |

*Table 18: Overview of available face datasets with ground truth blur labels.*

For the purpose of benchmarking face detection methods in challenging scenarios, different researchers collected unconstrained face images, see Figure 13. Jain and Learned-Miller [96] introduced one of the first of such dataset in their Face Detection Dataset and Benchmark (FDDB). Since the dataset has been collected for face detection only, it does not include identity labels like other web-collected datasets. Similarly, Nada et al. [97] introduced the web-collected Unconstrained Face Detection Dataset (UFDD) including different types of blur. The WIDERFACE dataset proposed by Yang et al. [22] is (at the time of this writing) the by far largest dataset for evaluating face detection algorithms in challenging scenarios including presence of blur. In summary, all of the aforementioned face detection benchmark datasets contain images depicting faces in unconstrained environments and usually multiple faces are shown per image.

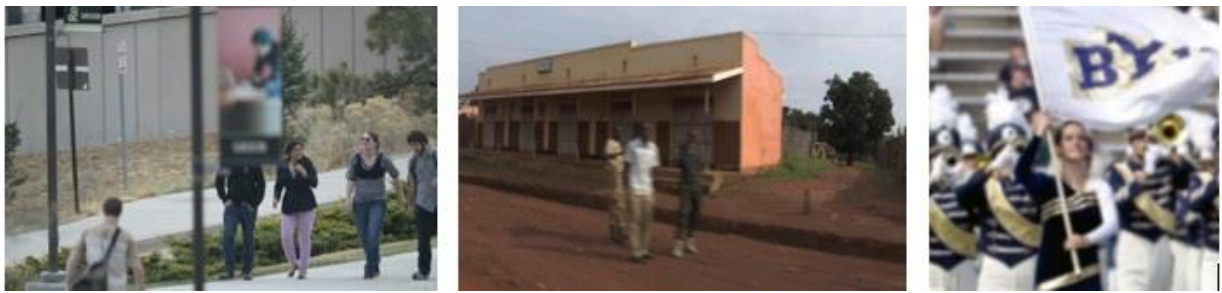

*Figure 13: Example images from UCCS (left), UFDD (middle), and WIDERFACE (right)*

---

138 https://vast.uccs.edu/Opensetface/
139 At the time of this writing the dataset was temporarily unavailable.

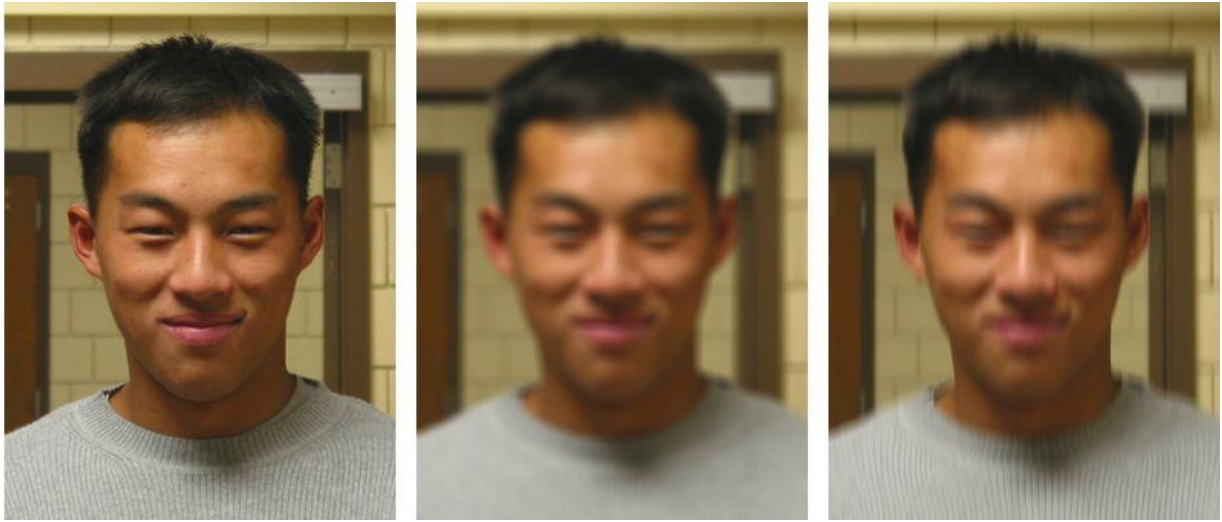

*Figure 14: Example (from FRGCv2) of sharp face image (left) and the application of synthetic averaging blur (middle) and synthetic motion blur (right)*

Günther et al. [98] introduced the Unconstrained Face Detection and Open-Set Face Recognition Challenge (UCCS) which contains unconstrained face images with blur. This dataset was collected for evaluating face detection algorithms as well as identification approaches. Hence, the dataset contains identity labels of a list of known subjects. A dataset containing blurred face images can also be acquired by intentionally configuring the capturing device in a way that the acquired face image is forced to be out-of-focus. This has been done by Han et al. [99], who captured facial images with various out-of-focus states. In contrast to the datasets mentioned before, this dataset has been collected in a constrained environment.

Alternatively, data subjects can be advised to intentionally move at the time of the acquisition in order to obtain de-focused face images or face images with motion blur. In contrast to other quality factors, e.g. pose or illumination (see Section 4 and 13), blurring effects are more difficult to control, in particular if these require certain interaction of the data subjects. This clearly hampers the acquisition of face images with desired blur intensities and has led different researchers to synthetically apply blur to facial images to simulate challenging conditions. This can be achieved by employing a simple averaging blur which assigns an average pixel value to a group of pixels or a blur with a Gaussian kernel. Moreover, a motion blur kernel can be used which averages the pixel values in a particular direction. Examples of synthetically blurred face images are depicted in Figure 14. It can be observed that the synthetic application of blur can lead to realistic results which simulate the effects of de-focus and motion blur. Note that in the shown examples, synthetic methods blur the entire image including the background.

For the evaluation of blur detection methods, further available datasets could be theoretically used. In particular, image datasets which do not necessarily contain images containing human faces. Usually such datasets are intended for evaluation of image visual quality assessment metrics. For example, the Categorical Image Quality (CSIQ) Database[140] or the Tampere Image Dataset (TID) 2013 Dataset[141] are frequently used datasets that contain labelled blurry images with arbitrary contents.

---

[140] https://qualinet.github.io/databases/image/categorical_image_quality_csiq_database/
[141] https://ponomarenko.info/tid2013.htm

## 6.3    Methods and Algorithms

There exists a large number of algorithms to detect blurred images in the scientific literature. These algorithms are not necessarily designed for facial images only. The vast majority of proposed methods are generic and can be applied to images irrespective of their contents. Moreover, said methods are usually benchmarked on general image datasets. However, used datasets and evaluation protocols vary a lot between the different publications which makes a comparison between them very difficult and potentially misleading.

In ISO/IEC WD4 29794-5:2022, de-focus is suggested to be estimated according to the method proposed in [100]. In this method, a mean filter of size 3x3 pixels is applied to the segmented facial region. Subsequently, the difference of the resulting image and the original face image is estimated. If the sum of all pixel differences is small, it is assumed that the image is already blurred. That is, the rationale behind this approach is that applying the aforementioned mean filter to an already blurred image will have less impact than applying it to a sharp image. While this method can be effective for facial images captured in controlled environments, it may be less useful in uncontrolled scenarios, e.g. with variations in illumination.

According to ISO/IEC WD4 29794-5:2022, image sharpness can be estimated by the method proposed in [101]. This sharpness metric is derived by modelling the image sharpness problem as a generalized eigenvalues problem. Firstly, the segmented facial region is normalized by its energy to minimize the effects caused by image contrast. Secondly, the covariance matrix is computed from this normalized image before it is diagonalized using Singular Values Decomposition (SVD) to obtain a series of eigenvalues. Finally, the image sharpness of the normalized facial region is determined by the trace of the first several eigenvalues.

Based on the idea that sharp images contain a larger amount of edges (compared to blurred images), simple edge detection filters can be used to estimate the image sharpness. For instance, the use of the well-known Sobel operator for image sharpness estimation was suggested by Marziliano et al. [102]. Alternatively, other edge detection filters, e.g. Laplacian filter, can be employed for the purpose of sharpness estimation. Ferzli and Karam [103] showcased that some image sharpness metrics fail to predict correctly the sharpness of images in the presence of gaussian noise. A noise-immune wavelet-based sharpness metric is proposed based on the so-called Lipschitz regularity for differentiating between edges and noise singularities.

Bahrami and Kot [104] presented an approach which has a very low computational cost. In this proposal, the maximum local variation (MLV) of each pixel is defined as the maximum intensity variation of the pixel with respect to its eight neighbours. The MLV distribution of the pixels is an indicative of global sharpness. Since high variations in the pixel intensities are a better indicator of the sharpness than low variations, the MLVs of the pixels are subjected to a weighting scheme in a way that larger weights are assigned to greater MLVs. Finally, the standard deviation of the weighted MLV distribution is used as a metric to measure overall image sharpness.

Zhao et al. [105] use the expectation of wavelet transform coefficients for estimating image sharpness. Hence, this method is simply based on the observation that the greater the probability of big detail coefficients, the sharper an image is expected to be. Expectation values are estimated from three wavelet sub-bands obtained by a separable discrete wavelet transform and combined by calculating a weighted sum of them. Hassen et al. [106] proposed a method using Local Phase Coherence (LPC). This approach is motivated by the observation that image blur disrupts the LPC structure, and the strength of LPC can be used to measure image sharpness. Li et al. [107] proposed an image sharpness measure using patch-based sparse coefficients with respect to the pre-learned dictionary, based on which the energies of patches are computed. Image sharpness is then estimated as the variance-normalized energy over a set of high-variance blocks, which is achieved by normalizing the total block energy using the sum of corresponding block variances. Gu et al. [108] presented a sharpness metric in the autoregressive parameter space. Firstly, energy- and contrast-differences in the locally estimated autoregressive coefficients are estimated. Subsequently, image sharpness is quantified with percentile pooling.

Different methods for detecting blurry regions in partially blurred images have been proposed in the scientific literature. Such blur detection and classification approaches, e.g. in [109], [110], [111], [112], [113], could also be used as a sharpness metric. Usually, such methods aim at performing a pixel-wise three-classification (sharp, de-focus, motion blur) based on an analysis of neighbouring pixels. Proposed schemes commonly apply some sort of feature extraction which allows a subsequent pixel-wise classification which is commonly implemented using machine learning techniques. For instance, Su et al. [109] proposed a technique for detecting blurred image regions by examining singular value information for each image pixel. To this end, Singular Value Decomposition (SVD) is applied which is a useful technique of the field of linear algebra that has been applied to different areas of image processing. Based on the extracted blur map constructed by a singular value feature, the presence as well as type of blur (de-focus or motion) can be detected reliably. Similarly, Lee and Kim [110] presented a blur detection and classification algorithm. As features, they estimate the magnitude of gradient for each image pixel and directional coherence for image patches. A Local Binary Pattern (LBP)-based blur detection method was proposed by Yi and Eramian [114]. The proposed sharpness metric exploits the observation that most local image patches in blurry regions have significantly fewer of certain local binary patterns compared with those in sharp regions. However, more recently it has been shown that these types of methods which utilize hand-crafted features are significantly outperformed by deep learning-based blur classification methods [115].

Many scientific works focus on measuring perceptual image sharpness of humans. For instance, Ferzli and Karam [116] introduced the concept of Just Noticeable Blur (JNB). Edge blocks were first selected from the image. Then, local contrast and edge width of the blocks were computed and integrated into a probability summation model, producing the sharpness score. Obviously, perceptual-based sharpness is highly subjective and the collection of reliable ground truth data should involve a sufficient amount of data subjects (at least 15 subjects according to [117]).

In ISO/IEC WD4 29794-5:2022, it is stated that motion blur can be estimated using different methods while the metric proposed in [100] is explicitly mentioned. In this method, a cepstrum transform[142] is applied to a potentially blurred image. The rationale behind this idea is that motion blur has periodic patterns in the frequency domain which result in detectable peaks in the cepstrum domain.

In [118], Ji and Lui presented a spectral analysis of image gradients, which enables identifying the blurring kernel of more general motion types (uniform velocity motion, accelerated motion and vibration). Kumar et al. [119] proposed grey scale-based method to estimated motion blur. This method is based on estimating the sharpness of edges, but instead of computing edge-width, changes in grey scale (luminance) values that are observed at an edge are estimated. This estimate is then used to classify an edge-pixel as sharp or blurred.

More recently, Kumar et al. [120] proposed a novel method to estimate the concurrent defocus and motion blurs in a single image. The authors show that motion blur can be modelled and analysed using Gaussian function. They also derive and experimentally verify the relation between equivalent Gaussian parameters for motion blur with motion blur length, i.e. intensity. These relations are further shown to help in estimating a more accurate defocus map from a single image by correcting the presence of spatially varying motion blur.

---

[142] A mathematical transformation in the field of Fourier analysis

 

To estimate edge density, in ISO/IEC WD4 29794-5:2022, it is suggested to apply the Face Region In Focus Measure (FRIFM) [95]. The FRIFM is simply the average Sobel edge magnitude within the region of the face (similar to [102]).

Although the results of individual papers are not directly comparable due to different protocols, datasets, and metrics, it can be assumed that high degrees of blurriness can be detected with high accuracy (greater than 90%). Challenging scenarios are those in which blur as well as other factors that generally degrade face image quality are present, e.g. nonuniform illumination.

## 6.4    Software

Despite the huge amount of published methods in the scientific literature, the amount of available software for blur estimation is rather limited. One reason for this could be the fact that many methods are rather simple, i.e. application of common filters which are available in common (image processing) libraries. The following table provides an overview of publicly available software for blur detection:

| Name | Method | Model Size | License |
|---|---|---|---|
| Blurred Region Detection using Singular Value Decomposition (SVD)[143] [109] | Blur Detection | n. a. | GPL-3.0 |
| Sharpness Estimation for Document and Scene Images[144] [119] | Sharpness Detection | n. a. | GPL-3.0 |
| LBP-Based Segmentation of Defocus Blur[145] [114] | Defocus Detection | n. a. | No license |
| MotionBlurDetection[146] | Motion Blur Detection | n. a. | No license |
| deep_blur_detection_and_classification[147] | Blur Detection | 128 MB | AGPL-3.0 |
| Blur Detection Haar Wavelet[148] | Blur Detection | n. a. | MIT |
| No reference blur metric based on Just Noticeable Blur (JNB)[149] [116] | Blur Detection | n. a. | No license |
| Efficient-Spatially-Varying-Blur-Detection-python[150] | Blur Detection | n. a. | BSD-2-Clause |
| Blur-detection-with-FFT[151] [111] | Blur Detection | n. a. | MIT |
| Enhancing Diversity of Defocus Blur Detectors via Cross-Ensemble Network[152] | Blur Detection | n. a. | No license |

---

[143] https://github.com/fled/blur_detection
[144] https://github.com/umang-singhal/pydom
[145] https://github.com/xinario/defocus_segmentation
[146] https://github.com/JuliaZur/MotionBlurDetection
[147] https://github.com/HyeongseokSon1/deep_blur_detection_and_classification
[148] https://github.com/pedrofrodenas/blur-Detection-Haar-Wavelet
[149] https://github.com/davidatroberts/No-Reference-Sharpness-Metric
[150] https://github.com/Utkarsh-Deshmukh/Spatially-Varying-Blur-Detection-python
[151] https://github.com/whdcumt/BlurDetection
[152] http://ice.dlut.edu.cn/ZhaoWenda/CENet.html

| Name | Method | Model Size | License |
|------|--------|------------|---------|
| Assessing Face Image Quality for Smartphone based Face Recognition System[153] [86] | Blur Detection, Sharpness Detection | n. a. | No license |
| MotionBlur-detection-by-CNN[154] [121] | Motion Blur Detection | 55 MB | No license |

*Table 19: Overview of available software for blur detection*

---

[153] https://share.nbl.nislab.no/g03-03-sample-quality/face-image-quality
[154] https://github.com/Sibozhu/MotionBlur-detection-by-CNN

 

# 7 Compression

Through the application of heavy image compression detailed textural information of face images may vanish (as illustrated in the previous section). This in turn can have a negative impact on various modules of a face recognition system. Some scientific works have studied the effects of image compression on face recognition, e.g. in [122] [123]. These works reported a strong impact of image compression on face recognition. However, these studies are rather old and therefore have not considered state of the art face recognition systems. Another evaluation of the impact of JPEG and JPEG 2000 image compression on the recognition accuracy of a commercial face recognition system is presented in ISO/IEC 19794-5:2011. Here, recognition accuracy dropped for compressed image of size less than 10KB for JPEG as well as JPEG 2000. A newer evaluation on the impact of JPEG compression on neural network-based face recognition has been presented by Grm et al. [124]. While the authors report significant performance degradations at very low-quality configuration, they did not put these in relation to compression ratios or resulting file sizes.

## 7.1 Impact on Face Recognition

Face recognition systems typically extract feature vectors for recognition purposes from cropped facial regions. Said regions exhibit relatively small resolutions of approximately 100 x 100 pixels. This means that downscaling is applied even if facial images are of higher resolution (with inter-eye distances complying with the requirements of ISO/IEC WD4 29794-5:2022, ISO/IEC 19794-5:2011 and ISO/IEC 39794-5:2019). Further, recently some research efforts have been devoted to the problem of low-resolution face recognition, e.g. in [125]. It has been shown that state of the art face recognition technologies can maintain their recognition performance when processing facial images at even lower resolutions. Therefore, it can be concluded that state of the art face recognition systems are partially robust to image compression and performance degradations are only expected for very strong compression.

In contrast to features extracted for recognition purposes, features needed for attack prevention, e.g. presentation or morphing attack detection, may be impacted more by the application of image compression. Some of the proposed attack detection approaches analyze facial images at pixel level and certain artefacts that are present in attack presentations might be vanished by the application of image compression.

## 7.2 Datasets

No face image datasets containing face images compressed at different compression ratios and corresponding a ground truth (applied compression algorithms and compression rates) have been found. However, such a dataset can be easily created by applying relevant compression algorithms to raw face images or compressed face images that do not contain any visible artefacts (visually lossless compression).

There exist some datasets comprising face images that have been intentionally down-sampled. These datasets are frequently used to benchmark face recognition algorithms designed for low-resolution scenarios, e.g. TinyFaces[155].

---

[155] https://qmul-tinyface.github.io/

## 7.3    Methods and Algorithms

The estimation of the degree of image compression applied to a (face) image can be performed in different ways. On the one hand, the compression ratio can be directly derived from the resolution and file size of an image. On the other hand, image processing can be applied to analyse the degree of applied image compression. The latter can be achieved by tracing the previously mentioned common artefacts induced by image compression methods. Eventually, both approaches can be combined. For instance, it could be analysed if, for an estimated compression, ratio artefacts are expected to be detected and whether this is the case and vice versa.

In ISO/IEC WD4 29794-5:2022, it is recommended to calculate the compression ratio based on the face image resolution and file size. More precisely, the compression ratio is estimated as the fraction of the file size and the corresponding uncompressed file size defined as the product of width, height, number of colour channels and number of bits per channel. Subsequently, the compression ratio is mapped to a quality score in the range [1,100].

Alternatively, there exist image processing methods designed to detect compression artefacts. These methods are frequently designed for image forensic applications where inconsistencies in image compression ratios across image regions indicate digital manipulations. In JPEG forensics, the main methods are either based on the analysis of histograms of DCT coefficients, e.g. in [126], or based on the detection of a higher contrast, i.e. hard transitions, at the block edges, e.g. in [127]. Such algorithms could be utilized to analyse whether the facial region of a face image is affected by strong compression artefacts and if so, low quality values could be assigned depending on the strength of detected artefacts.

Nikoukhah et al. [128] propose a simple method for detecting JPEG-compressed image regions by counting the number of zeros in the DCT of 8 x 8 pixels blocks.  This number depends on the compression factor and can be used to predict it. However, it is important to note that the number of zeros in the estimated DCTs also depends on the image content.

More recently, Uchida et al. [129] proposed a neural network trained to detect JPEG-compressed regions in images.  Training data, i.e. compressed images and ground truth JPEG quality factors, are generated by compressing raw image patches. The resulting network was shown to reliably detected and assign quality factors to image patches in JPEG- compressed images.

Further, many methods for the detection of double-compression artefacts have been proposed for image forensics. If images are edited and re-compressed certain double compression traces can be detected for JPEG, e.g. in [130], as well as JPEG2000, e.g. in [131].[156]

In addition, different methods for removing image compression artefacts from images have been proposed, e.g. in [132]. Recent approaches make use of learning techniques, in particular generative adversarial networks (GAN), to learn image-to-image translations that remove image compression artefacts [133]. With respect to image compression detection, the difference between the resulting restored image and its original counterparts could indicate whether the original image contains artefacts and to which extent.

## 7.4    Software

Table 20 lists open-source software for detection and removal of image compression artefacts. Note, that only software for the detection JPEG artefacts has been found.

---

[156] Note that, as per ISO/IEC 19794-5:2011, multiple compression shall not be applied to derive a full-frontal face image type.

| Name | Method | Model Size | License |
|------|--------|-----------|---------|
| Blocky-Artefacts-Detection[157] | JPEG Compression Detection | n. a. | No license |
| ZERO[158] [128] | JPEG Compression Detection | n. a. | GNU Affero GPL |
| GOD[159] | JPEG Compression Detection | n. a. | GNU Affero GPL |
| Pixelwise JPEG Compression Detection[160] [129] | JPEG Compression Detection | 3 MB | MIT |
| FBCNN[161] [134] | JPEG Compression Artefact Removal | 274 MB | Apache 2.0 |
| JPEG-quantsmooth[162] | JPEG Compression Artefact Removal | n. a. | GNU LGPL |
| Artefact removal GAN[163] [133] | JPEG Compression Artefact Removal | 171 MB or 346 MB[164] | MIT |
| QGCN[165] [132] | JPEG Compression Artefact Removal | 20 MB | MIT |

*Table 20: Examples of software available for the detection and removal of JPEG artefacts.*

---

[157] https://github.com/tho-graf/Blocky-Artefacts-Detection
[158] https://github.com/tinankh/ZERO
[159] https://github.com/tinankh/GOD
[160] https://github.com/kuchida/PixelwiseJPEGCompressionDetection
[161] https://github.com/jiaxi-jiang/FBCNN
[162] https://github.com/ilyakurdyukov/jpeg-quantsmooth
[163] https://github.com/mameli/Artifact_Removal_GAN
[164] https://github.com/mameli/Artifact_Removal_GAN/releases/tag/1.1
[165] https://github.com/VDIGPKU/QGCN

# 8 Unnatural Colour and Colour Balance

In image processing, colour balance is the global adjustment of the intensities of the colours (typically red, green, and blue primary colours corresponding to the respective colour channel). Colour imbalance usually results in colour casts. Alternatively, the intensities of all colours can be lower or higher than the natural appearance, resulting in a too low or too high colour saturation.

A colour cast is a tint of a particular colour, usually unwanted, that evenly affects an image entirely or partially. Similar effects are caused by excessive colour saturation. Certain types of light can cause digital cameras to render a colour cast. Moreover, illuminating a subject with light sources of different colour temperatures will likely causes colour casts. Examples of a colour cast and extreme saturation in a face image is depicted in Figure 15. Furthermore, colour casts and incorrect saturation can result from inappropriate settings (e.g. filters intended for artistic photographic effects, or for the compensation of illumination with unbalanced colour) or defects of the camera. Finally, intentional image editing can result in unnatural colours.

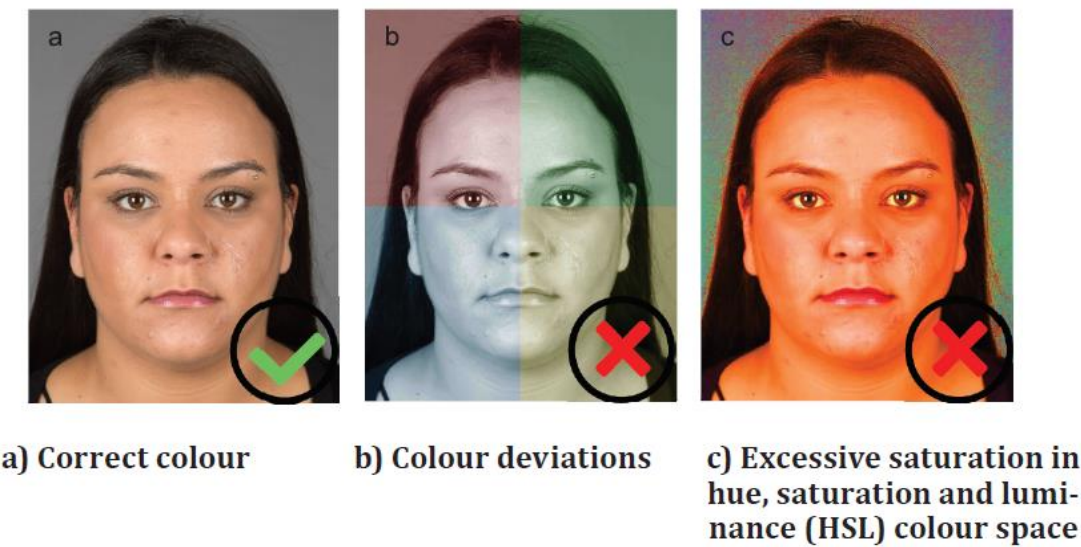

a) Correct colour    b) Colour deviations    c) Excessive saturation in hue, saturation and luminance (HSL) colour space

*Figure 15: Example of face image with a) correct colour b) unnatural colour and c) excessive saturation (taken from ISO/IEC 39794-5:2019)*

Since the colour of the skin, facial hair, moles, etc. are discriminative facial properties, such undesired colour-related effects may impact the performance of pattern recognition methods in general and face recognition algorithms specifically. Therefore, ISO/IEC WD4 29794-5:2022, ISO/IEC 19794:2011 and 39794-5:2019 all require that the image should not show unnatural skin colours. However, in order to prevent discrimination of certain groups of subjects, e.g. persons of African descent, it is important that quality assessment algorithms consider a wide variety of naturally occurring skin tones; furthermore, the skin colours resulting from tattoos, moles or other anomalies of the skin colour (e.g. port-wine stains) should be taken into account.

Obviously, strong variations in illumination can cause very bright or dark and, hence, unnatural skin tone. However, in this section focus is put on unnatural variations in terms of colours, since brightness variations are covered by other quality metrics, in particular illumination (see Section 4). Note that this overlap of quality metrics is also noticeable in existing standards.

## 8.1 Impact on Face Recognition

There are no scientific studies that explicitly investigate the impact of colour imbalance on face recognition performance. However, there exist some works that analyse the general role of colour in face recognition. Focusing on human face recognition, it has been shown that colour information plays an important role, i.e. if colour information is available, human face recognition performance tends to significantly improve as compared to greyscale imagery [135]. More recently, Drozdowski et al. [136] reported that the absence of colour information may hamper reliable face detection and subsequent feature extraction. Nevertheless, the face image quality estimation and comparison modules of state of the art face recognition systems were reported to perform equally well on constrained colour and greyscale images [136]. Figure 16 depicts the score distributions obtained by a well-established face recognition system on a constrained face image dataset. It can be observed that colour information has negligible impact on face recognition performance.

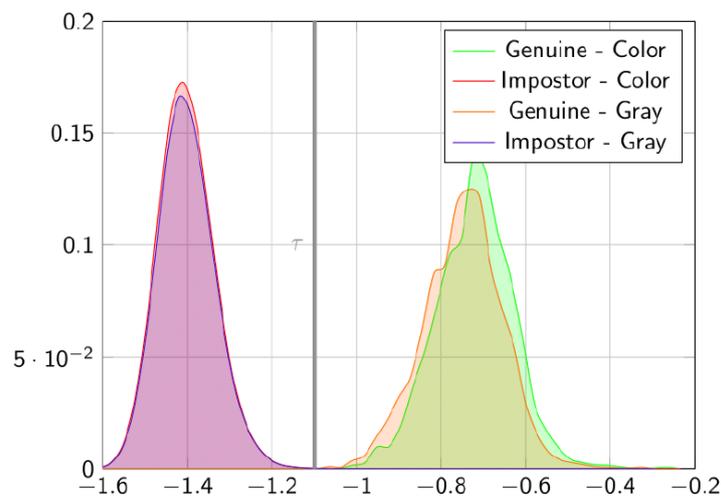

*Figure 16: Mated and non-mated score distributions obtained by the ArcFace face recognition system on colour and greyscale images on the FRGCv2 dataset. T denotes the decision threshold corresponding to an FMR of 0.1%*

The contradictory results of the aforementioned publications may be explained by the investigation in [137], which showed that some CNNs perform better on colour images, while others give higher accuracy for greyscale images, even if trained on colour images.

The aforementioned observations are further backed by technical guidelines that allow the capturing and processing of greyscale face images, e.g. in border control systems where greyscale images are frequently captured in [138]. Based on these observations it may be concluded that colour cast may also have only small impact on most face recognition modules, in particular if a greyscale conversion is firstly performed. However, soft-biometric classifiers such as skin tone estimation methods are expected to be significantly impacted by colour casts.

Apart from face recognition, it was found that visual object classification algorithms which utilize colour information for their predictions are generally more sensitive to colour imbalance [138], see for an example Figure 17.

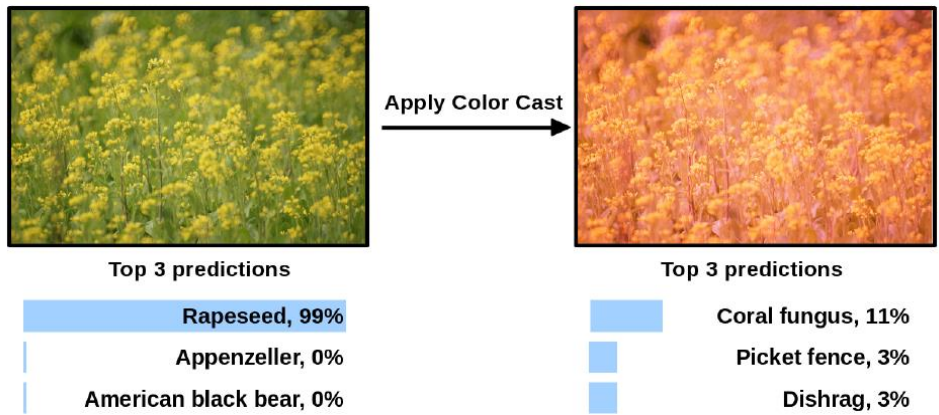

*Figure 17: Example of misclassification in the presence of colour cast [259]*

## 8.2 Datasets

While it is reasonable to assume that many face image datasets contain images with unnatural colours (in particular unconstrained face datasets), no face dataset has been found which contains ground truth labels for the naturalness of colour. However, colour analysis methods can be applied to filter existing datasets, e.g. to obtain a subset of facial images with colour casts. Alternatively, unnatural colour can be obtained synthetically, i.e. by modifying the colour information of existing face images datasets. Further, there exist general image datasets with accurate colour labels, e.g. the "Colour Checker Dataset" [139].

## 8.3 Methods and Algorithms

The ISO/IEC WD4 29794-5:2022 specifies a method to measure unnatural colour by firstly calculating the means of RGB values of the $L$ and $R$ intensity measurement zones. These measurement zones are defined in ISO/IEC 39794-5:2019, see Figure 18. Subsequently, the image unnatural colour aspect is estimated on the two means[166]. Note that it could be argued that those zones could be occluded by facial hair, in particular beards.

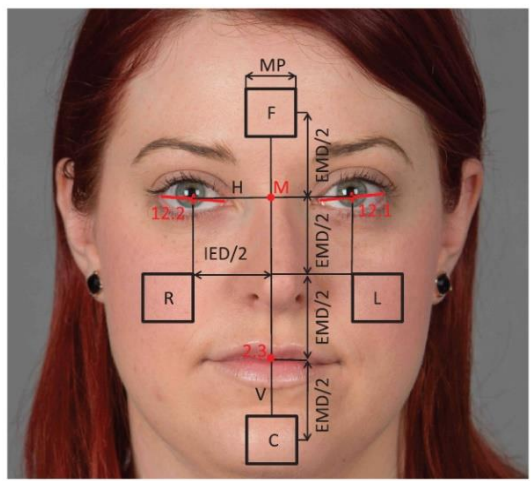

*Figure 18: Location and size of the intensity measurement zones (taken from ISO/IEC 39794-5:2019)*

---

[166] In ISO/IEC WD4 29794-5:2022, there still appeared to be an error such that it remains unclear how the quality metric unnatural colour aspect is calculated.

Alternatively, the CIELAB colour space, also referred to as L*a*b* can be considered. This colour space is defined by the International Commission on Illumination. It expresses colour as three values: L* for perceptual lightness, and a* and b* for the four unique colours of human vision: red, green, blue, and yellow. The lightness value, L*, defines black at 0 and white at 100. The a* axis is relative to the green–red opponent colours, with negative values toward green and positive values toward red. The b* axis represents the blue–yellow opponents, with negative numbers toward blue and positive toward yellow. ISO/IEC 39794-5:2019 only accepts images for which a* and b* are positive. Still, while negative a* and b* infers unrealistic colours, having positive values is not a sufficient condition to guarantee a natural colour. As stated in ISO/IEC WD4 29794-5:2022, the unnatural colour aspect could also be estimated by the number of pixels with negative a* value, and those with negative b* value, respectively, in the zones $L$ and $R$. The maximum of their fractions with respect to the region sizes is returned as the image unnatural colour aspect. Here, high values refer to unnatural colour, i.e. low quality. Detection of colour imbalance in face images can be based on the analysis of histograms of each colour channel. For instance, colour casts can be easily detected by estimating statistics over colour channel histograms, e.g. mean and variance, for facial patches which are expected to contain skin (similar to the intensity measurement zones described above). Subsequently, the obtained colour channel histogram statistics can be compared to that of face images which are known to have natural colour. Here, it is important that possible variations in natural skin tones are considered.

Similar concepts have been proposed in the scientific literature to detect colour imbalance. For instance, Lu et al. [140] proposed a handcrafted approach for arbitrary photographs, which analyses colour histograms to detect colour casts. Obviously, such general approaches, which are designed to works irrespective of the image content may be more complex compared to methods applied to face images only.

Besides approaches to detect unnatural colours in images, several methods have been proposed to correct colour imbalance. These methods are commonly referred to as *colour balancing* or *colour constancy* algorithms [141]. While earlier hand-crafted approaches are based on certain assumptions, e.g. that the average colour in the image is grey, more recent approaches are based on machine learning, e.g. using CNNs [142], [143]. Also, there exists several patents for detecting and correcting unnatural colour in images, e.g. [144].

It is important to note that in ISO/IEC WD4 29794-5:2022, the saturation aspect is not mentioned and, hence, no method for estimating saturation is suggested. It is reasonable to assume that methods detecting colour imbalance may as well be able detect excessive saturation. Further, an image can be converted to a colour space with saturation information. For instance, an RGB image can be converted to the HSL (for hue, saturation, lightness) or HSV (for hue, saturation, value) colour space from which the image saturation can be derived. Such colour space conversions can be implemented with basic functions available in common image libraries. Based on such a simple estimation of image saturation, thresholding can be applied to detected extreme saturation values. Some methods have been proposed to automatically correct saturated pixels in images, e.g. in [145].

Furthermore, there are pre-defined/existing colour palettes, which represent a supplement in the estimation of skin tones. For example, the so-called "Monk Skin Tone Scale"[167] from Google, which provides 10 different shades for people with darker skin tones or the well-known "Fitzpatrick Scale" used as classification system for assessing skin types in dermatology. Certainly, one must also consider redness or pallor due to variations in blood flow or even temporary changes in skin appearance due to illness or too much sun in the form of a sunburn resulting in a strongly reddened/burned skin colour. However, one could make a statement about an unnatural colour exactly when the corresponding colour of the facial image cannot be associated to a colour in the scale assuming that the scale holds a whole range of natural occurring skin tones.

---

[167] https://skintone.google/faqs

## 8.4  Software

It is assumed that methods which correct unnatural colour can also be used to detect whether an image contains unnatural colours or not. Even if algorithms do not do this explicitly, the resulting images can be compared to the original image to detect, if corrections have been applied, which in turn would mean that the image contained unnatural colour. While no algorithms were found which are specifically designed for facial images, Table 21 lists some examples of software available for colour correction in arbitrary images.

| Name | Method | Model Size | License |
|---|---|---|---|
| Edge-based Color Constancy[168] [146] | Colour Constancy | n. a. | No license |
| Deep_White_Balance[169] [143] | Colour balancing | 20 MB | NonCommercial-ShareAlike 4.0 International |
| Exposure_Correction[170] [147] | Exposure correction | 27 MB | NonCommercial-ShareAlike 4.0 International |

*Table 21: Examples of software available for the correction (and detection) of unnatural colour in images.*

---

[168] lear.inrialpes.fr/people/vandeweijer/code/ColourConstancy.zip
[169] https://github.com/mahmoudnafifi/Deep_White_Balance
[170] https://github.com/mahmoudnafifi/Exposure_Correction

# 9     Camera Lens Focal Length

The *camera focal length* is the distance of the camera lens to the camera sensor. The camera focal length in combination with the camera subject distance (see Section 9) affects the distortion of the projection of the face on the image. In the same way as for the camera subject distance, the camera focal length can be accessed from the EXIF meta data of JPEG image files, if the corresponding optional tag is available. These meta data could be accessed for quality assessment.

As explained in Section 9, a separate camera focal length estimation can be obtained from an estimation of the camera subject distance on the inter-eye distance in pixels, assuming an inter-eye distance of the subject of 60-65mm. In this view, a separate estimation of the camera focal length is redundant to other features. Even if accessed from the EXIF metadata tag value, assessing the camera focal length in addition to other quality aspects, does not improve actionable feedback, as it cannot be a changed by a user without replacing hardware.

Therefore, in the authors' view, the camera focal length would be redundant to other features and the assessment of the camera focal length should not be included in facial quality assessment software.

# 10    Camera Subject Distance

As per ISO/IEC 39794-5:2019, the *camera to subject distance* (CSD) is defined as the distance between the eyes plane of a capture subject and the image plane of the camera. The CSD is illustrated in the following figure.

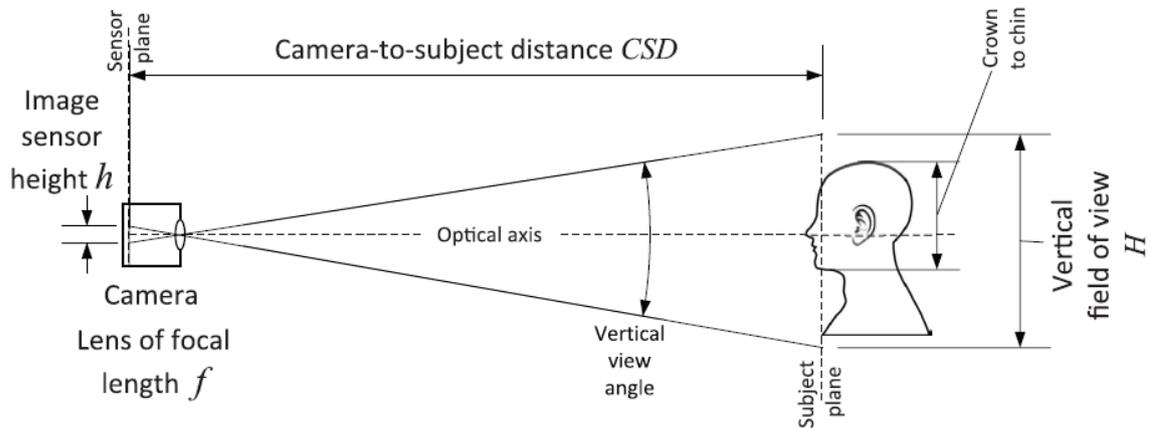

*Figure 19: Visualization of the camera to subject distance (CSD) (taken from ISO/IEC 39794-5:2019)*

## 10.1    Impact on Face Recognition

The CSD affects the distortion of the projection of the face on the image. It is important to note, that distortional effects cannot be avoided when projecting a 3-dimensional object onto a 2-dimensional image. Yet, different CSDs lead to different magnitudes of distortion. In general, facial parts closer to the camera appear larger as compared to facial parts having a larger distance to the camera. The following figure illustrates facial images with a strong distortion. Here one can clearly see that some parts of the face, e.g. the nose or the eyes, appear larger due to distortional effects. Widely known distortions that are created are "selfie effect" or "fisheye effect".

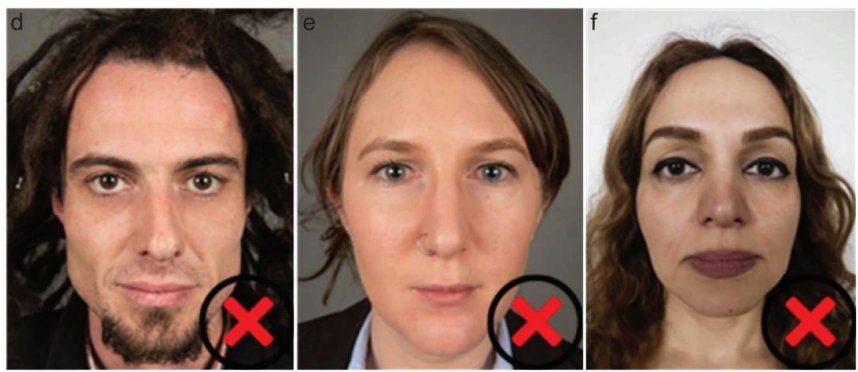

*Figure 20: Facial images with a strong distortion (taken from ISO/IEC 39794-5:2019)*

The relation between these effects is proportional to the inverse of the CSD. Hence, the larger the CSD is, the less distorted is the face projection on the image. Therefore, the larger the CSD is, the less is the negative impact on the capability to recognize the face due to distortion. The following figure illustrates a good facial image appearance through compliance with the CSD requirements given in ISO/IEC 39794-5 Table D.2 where the CSD should map the following requirements:

    

- 0,7 m ≤ CSD ≤ 4 m for 1:1

- −1 m ≤ CSD ≤ 4 m for 1:N

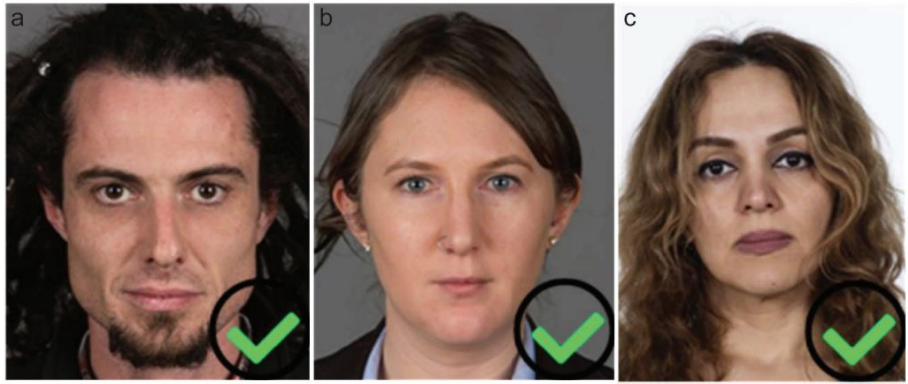

*Figure 21: Facial images with a good appearance from ISO/IEC WD4 29794-5:2022*

Note that, the larger the CSD is the more details of the facial image get lost. From this view, when it comes to mapping the CSD to a quality score, it seems useful to penalize both short and large values of the CSD. On the other hand, large CSD values are implicitly penalized by assessing the inter-eye distance (IED) feature.

## 10.2 Datasets

While it is likely that many face datasets contain images with different CSD's, no face datasets were found in which the distance between the camera and the subject is provided as ground truth data. However, one could use a face dataset consisting of JPEG images and use the obtained EXIF SubjectDistance tag value as ground truth data. Yet, if one intends not to apply a separate CSD estimation method, but to assess the CSD value contained in the EXIF SubjectDistance tag, no dataset for evaluation would be needed.

## 10.3 Methods and Algorithms

According to ISO/IEC WD4 29794-5:2022, if available, the CSD could be obtained from the metadata of a facial image, precisely from the optional EXIF SubjectDistance tag of a JPEG image file. However, if the reference image is a scan from a paper photograph provided by the applicant, this meta data is not available, and it may also not be available in other circumstances, e.g. if the image had been converted to other image file formats.

Alternatively, one may compute the CSD using the following formula:

$$CSD \cong f * \frac{H}{h}$$

where f is the camera focal length (see Section 9), H denotes the image height, and h is the image sensor's height. Note, that the fraction H/h can be approximated using the inter-eye distance (IED) (see Section 12) and the subject's ground truth eye distance (e.g. in millimetres). In this way, each method that estimates the CSD from a facial image, estimate the subject's eye distance (e.g. in millimetres). As a subject's eye distance typically lies between 60mm and 65mm. However, there can be problems with methods that cannot read the CSD from the metadata of an image but try to calculate it approximately. For example, due to disease-related facial malformation or facial disabilities, a distortion can be incorrectly determined although the CSD is optimal.

It may be worth noting, that, if the camera focal length is chosen reasonable, the assessment of the CSD implicitly results from assessment of the location of the head (see Section 13) and, at the same time, from the assessment of the inter-eye distance (see Section 12). From this perspective, a separate estimation of the CSD from a facial image seems to be redundant and not worth the effort. Yet, if accessible from the EXIF SubjectDistance tag, there is no harm in assessing the CSD as this could allow for enhanced actionable feedback to the user.

Under the assumption that the image has not been post-processed (e.g. by scaling or cropping) and the focal length of the camera is known, the distance can be estimated from the size of the face in the image. However, these assumptions cannot be relied on.

## 10.4   Software

The following table lists examples of publicly available software for object-distance measurement from images based on the assumption that the focal length is known.

| Name | Method | Model Size | License |
|------|--------|------------|---------|
| Object Detection and Distance Measurement[171] | Object detection and distance estimation | 237 MB | No license |
| Distance_measurement_using_single_camera[172] | Face detection and distance estimation | n. a. | No license |
| YREyeDistance[173] | IED is used to determine the distance between eyes and phone screen | n. a. | MIT |
| Eye to Screen Distance Measurement with Computer Vision[174] | IED is used to determine the distance between eyes and web cam (screen). | n. a. | No license |

*Table 22: Examples of software available for distance measurement in images.*

---

[171] https://github.com/paul-pias/Object-Detection-and-Distance-Measurement
[172] https://github.com/Asadullah-Dal17/Distance_measurement_using_single_camera
[173] https://github.com/daQiuQiu/YREyeDistance
[174] https://github.com/Ahmed-0357/eye_to_screen_distance_measurement

# 11    No Occlusion of the Face

ISO/IEC WD4 29794-5:2022 defines the criteria "mouth and nose visible" and "eyes visible" without any restriction with respect to why the eyes, mouth and/or nose are not visible. ISO 19794-5:2011, ISO/IEC 39794-5:2019 and ICAO 9303 Part 3 Vol. 1 even contain the recommendation that head coverings should be absent to prevent obscuration of elements of the facial area. BSI TR 03121-3 Part 3 Vol.1 requires that the eyes are not occluded. ISO/IEC 39794-5:2019 (in sections D.1.4.3.3 - D.1.4.3.5 and D.2.3.7 - D.2.3.8) gives very detailed recommendations on which occlusions of the eyes are not acceptable.

In facial images, the eyes, mouth and/or nose can be occluded due to different reasons:

- *Pose:* An extreme pose, e.g. yaw or pitch angle, resulting in one eye, parts of the mouth or nose being occluded by other parts of the face. This case is already covered by the criterion "Pose", and is therefore not further discussed in this section.
- *Illumination:* Shadows on the face, resulting on some parts being hardly visible. This case is already covered by the criteria illumination under-exposure and illumination uniformity, and is therefore not further discussed in this section.
- *External occlusions:* A (partial or total) occlusion of the eyes, mouth, or nose by other parts of the body[175] (in particular the hands), items (e.g. mobile phones), accessories and clothes (sunglasses, scarf, veil, hat, eye patch) or medical masks. The investigations in this section exclusively deal with this kind of occlusion.

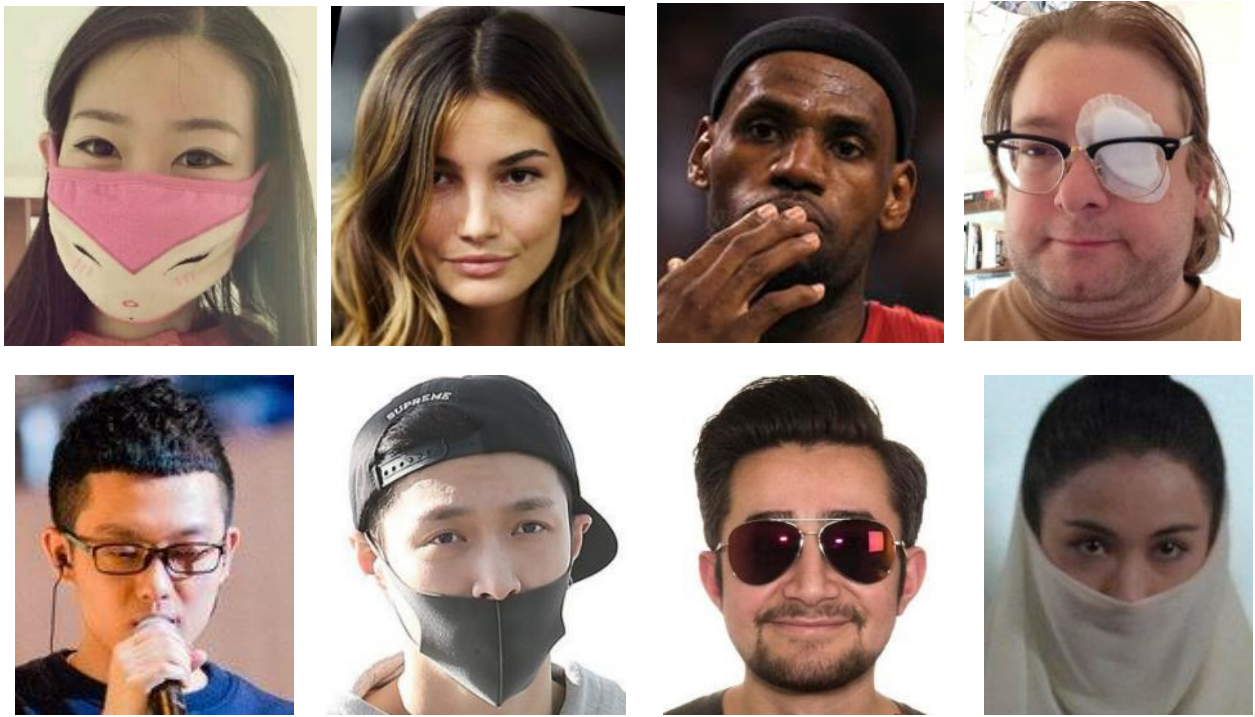

*Figure 22: Examples (from the MAFA dataset) for external occlusions of faces*

ISO/IEC 39794-5:2019 (in Section D.2.3.7) specifies that the "eye pupils and irises shall be visible", that "eye patches shall not be worn", except if "the subject asserts a need to retain the patch (e.g. a medical reason)".

---

[175] The occlusion of the eyes by the eye lids (closed eyes) is covered by the criterion "Eyes Open"

A special case of external occlusion is occlusion by eyeglasses. Glasses that are worn for medical reasons (due to ametropia) and that have transparent lenses are permitted for reference images of most identity documents and also for the EES, even though their rims, in particular in case of heavy frames, may occlude small parts of the face. (Even glasses with moderately tinted glasses may also be permitted if they are medically indicated). However, for reference images, it is important that the rims and frames of the glasses do not obscure eye details. Annex D (normative) of ISO/IEC 39794-5:2019 specifies the *Eye Visibility Zone (EVZ)* "as the covering rectangle having a distance V of at least 5 % of the IED to any part of the visible eye ball", and requires that the "EVZ shall be completely visible and unobscured". Furthermore, it stipulates that "frames shall not be thicker than 5 % of the inter-eye distance".

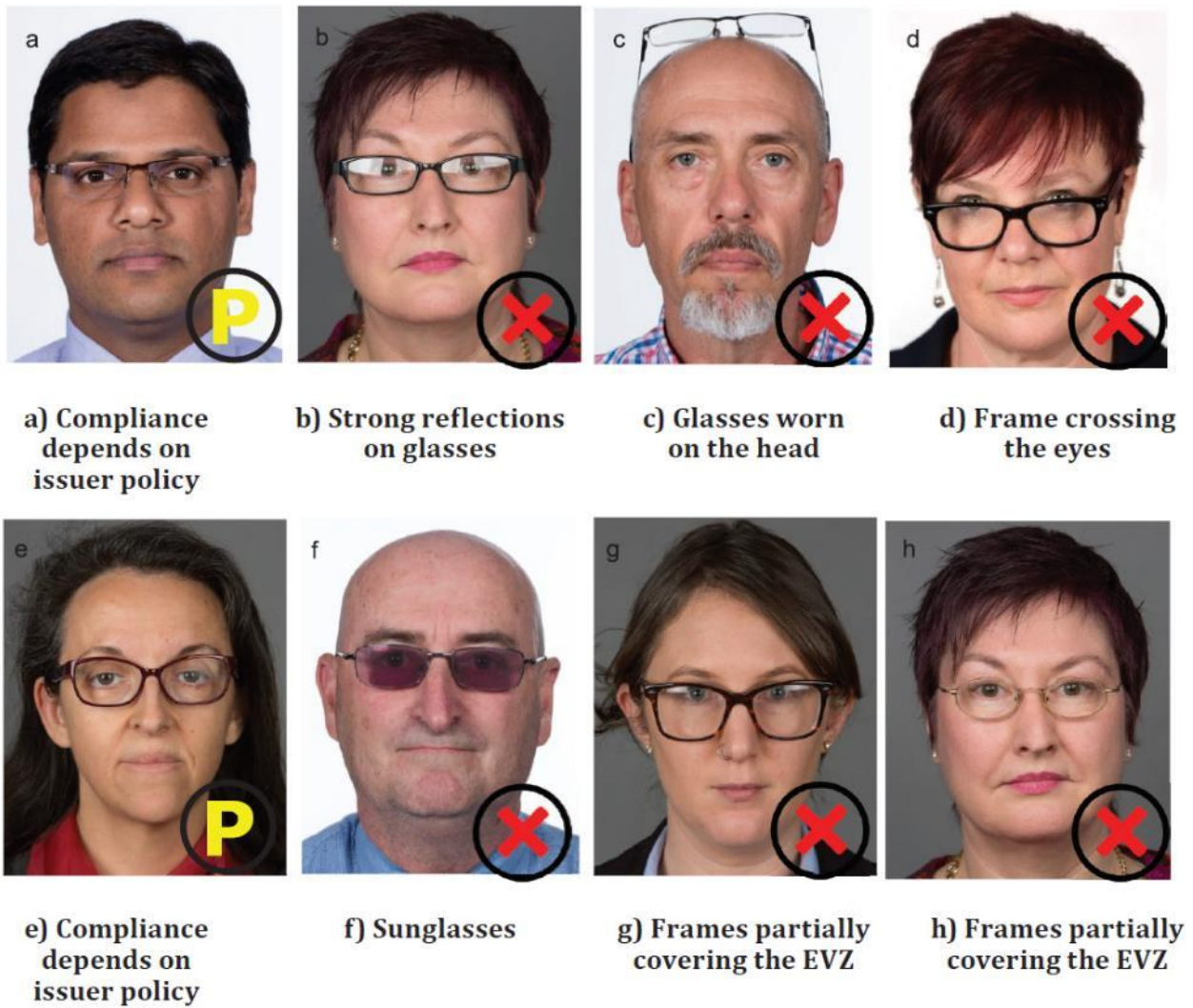

a) Compliance depends on issuer policy

b) Strong reflections on glasses

c) Glasses worn on the head

d) Frame crossing the eyes

e) Compliance depends on issuer policy

f) Sunglasses

g) Frames partially covering the EVZ

h) Frames partially covering the EVZ

*Figure 23: Examples of glasses on facial images that may be permitted, depending on the policy of the document issuer (P), and that are typically not acceptable (X) (taken from ISO/IEC 39794-5:2019)*

Another type of occlusion can result from reflections on glasses. Annex D.2.3.6 of ISO/IEC 39794-5:2019 stipulates the following: "There shall be no lighting artefacts or flash reflections on glasses. Lighting artefacts covering any region of the eyes shall not be present." However, such reflections cannot only result from lighting but also from natural illumination.

Figure 23 shows examples of glasses in facial images that are typically not acceptable as reference images of identity documents because they partly obscure the EVZ, and examples that may be compliant, depending on the policy of the document issuer. Note, however, that the case of example c) - glasses worn on head – is not covered in this document.

## 11.1 Impact on Face Recognition

The negative impact of occlusions of mouth and nose on face recognition is well-known [148], in particular for face masks [149], but has also been confirmed (based on synthetized data) for sunglasses and glasses with thick frames [150], [151]. In [84], Grm et al. investigated the impact of the occlusion of different face parts on the performance of CNNs trained for face recognition and found the highest deterioration for the complete periocular region, followed by the eyes and, with clear distance, the nose.

From the results of public challenges for masked face recognition, it is known that, even for facial recognition algorithms specifically trained on masked faces, the recognition performance drops by orders of magnitude if the subject is wearing a mask in the reference and/or probe image [152] [153]. The degradation of recognition performance implies a reduced utility which is also reflected in lower quality scores output by face image quality assessment (FIQA) algorithms [154].

Furthermore, the face detection and landmark localization, which are necessary pre-processing step for most facial recognition algorithms and also for the assessment of many QA criteria, becomes much more challenging in case of occlusions [148] [155], even for detection algorithms specifically trained on masked face images [156].

However, for probe images, e.g. live images in an automatic border control gate, moderate occlusions, e.g. by sunglasses or face masks, do not always result in a rejection, because current face recognition algorithms are so effective that they can successfully perform verification at the required security level even with such occlusion in the probe image. Nevertheless, even slight occlusions in the probe image can cause problems for PAD and MAD systems.

## 11.2 Datasets

The following table summarizes several available datasets used for arbitrary occlusions. During the investigations for writing this report, the public project "Sunglass_Overlay_Effect"[176] has stood out, as it implements methods to artificially add templates in front of the face, such as sunglasses on eyes. This method can be useful for creating datasets in the future.

| Name | Subjects | Images | Remark | Constrained? | License |
|------|----------|--------|--------|--------------|---------|
| CMU Face Images Dataset[177] | 20 | 640 | Varying pose, expression, eyes (sunglasses or not) and size | No | Royalty-free, educational purpose |
| AR Face[51] | 126 | 4,000+ | Features frontal view faces with different facial expressions, illumination conditions, and occlusions (sun glasses and scarf) | Yes | Royalty-free, educational purpose |
| Selfies-with-sunglasses[178] | unknown | 5,536 | Selfies with sunglasses, contains, 2768 unannotated images | No | No license |
| Kaggle (amol07)[179] | unknown | 3,855 | Train: 1,776 (no glass), 1,475 (glasses) Valid: 362 (no glass), 242 (glasses) | No | CC BY-NC-SA 4.0 |
| SoF[180] | 112 | 42,592 | All images with eyeglasses under harsh illumination conditions, synthetic occlusions for nose and mouth | No | Royalty-free, educational purpose |
| ROF[181] | 180 | 5,559 | 3,195 neutral images, 1,686 sunglasses images, 678 masked images | No | MIT |

---

[176] https://github.com/GH0STH4CKER/Sunglass_Overlay_Effect
[177] http://archive.ics.uci.edu/ml/datasets/cmu+face+images
[178] https://github.com/shreyas0906/Selfies-with-sunglasses
[179] https://www.kaggle.com/amol07/sunglasses-no-sunglasses
[180] https://sites.google.com/view/sof-dataset
[181] https://github.com/ekremerakin/RealWorldOccludedFaces

| Name | Subjects | Images | Remark | Constrained? | License |
|------|----------|--------|--------|--------------|---------|
| SejongFaceDatabase[182] | 100 | 24.800 | 8 facial addons (incl. cap, scarf, mask) plus 7 combinations of these addons per subject. Each facial image is captured in visible, visible plus infrared, infrared, and thermal spectra. | Yes | GPL 3.0 |
| MERL-RAV[65] | unknown | 19.000 | Annotations consisting of 68 landmarks and their visibility (visible, self-occluded, externally occluded) for the AFLW data set | No | No license |
| Webface-OCC[183] | 10,575 | 804,704 | Synthetic occlusions applied to the CASIA-Webface dataset, no labels on type of occlusion | No | Royalty-free, academic use |
| COFW | unknown | 1,007 | Annotations of 29 facial landmarks and their visibility | No | CC-BY |

*Table 23: Summary of available datasets regarding sunglasses detection*

The following table summarizes several available datasets used in face mask detection. During the investigations for writing this report, no constrained datasets were found in regards to face mask detection. Therefore, all datasets listed below should be considered to contain "in the wild" images.

| Name | Subjects | Images | Remark | License |
|------|----------|--------|--------|---------|
| AIZOO[184] | unknown | 7,959 | Relabelled combination of Wider-Face and MAFA. | MIT |
| Dataset by Chandrika Deb[185] | unknown | 4,095 | Mask (2,165), no mask (1,930). Real images of faces wearing masks. Collected from the internet. | MIT |
| Dataset by Sunil Singh[186] | unknown | 7,500 | Labelled dataset composed of MAFA, WIDER FACE and manually prepared images by surfing various sources on the web. | No license |
| Dataset by VictorLin000[187] | unknown | 678 | Classes: No mask, improperly and mask | No license |
| FDDB[7] | unknown | 2,845 | Contains annotations for 5,171 faces taken from the Faces in the Wild Dataset. | Academic use only |
| FMLD[188] | unknown | 41,934 | Combined, relabelled dataset of MAFA and Wider Face. With mask (correctly worn) (29,532), with mask (incorrectly worn) (1,528), without mask (32,012) | MIT |

---

[182] https://github.com/usmancheema89/SejongFaceDatabase
[183] https://github.com/Baojin-Huang/Webface-OCC
[184] https://github.com/AIZOOTech/FaceMaskDetection
[185] https://github.com/chandrikadeb7/Face-Mask-Detection
[186] https://drive.google.com/drive/folders/1pAxEBmfYLoVtZQlBT3doxmesAO7n3ES1?usp=sharing
[187] https://github.com/VictorLin000/YOLOv3_mask_detect
[188] https://github.com/borutb-fri/FMLD

| Name | Subjects | Images | Remark | License |
|------|----------|--------|--------|---------|
| Kaggle (Alexandra Lorenzo)[189] | unknown | 1,226 | Dataset to detect subjects with/without mask. | No license |
| Kaggle (Ashish Jangra)[190] | unknown | ~12,000 | Images with face mask (~6,000) scraped from the internet | CC0 1.0 |
| Kaggle (Dhruv Makwana)[191] | unknown | 440 | With mask: 220 Without mask: 220 | No license |
| Kaggle (ivan)[192] | unknown | 1,148 | 3 Classes: mask, none, poor | No license |
| FMD[193] | unknown | 853 | 3 Classes: with_mask, without_mask, mask_weared_incorrect | CC0 1.0 |
| Kaggle (Mohamed Loey)[194] | unknown | 1,415 | Based on MMD and FMD. Removed bad quality and redundancy. | DbCL 1.0 |
| Kaggle (Spandan Patnaik)[195] | unknown | 2,079 | 3 Classes: Subjects wearing mask, subjects not wearing mask, subjects wearing mask incorrectly | CC0 1.0 |
| Kaggle (Sumansid)[196] | unknown | 339 | Mask: 208 images No Mask: 131 images | No license |
| Kaggle (Wobot Intelligence)[197] | unknown | 6,024 | Variety of classes regarding occlusion: face_with_mask, mask_colourful, face_no_mask, face_with_mask_incorrectly, mask_surgical, face_other_covering | CC0 1.0 |
| Kaggle (Vinay Kudari)[198] | unknown | 772 | Classes: mask_weared_incorrect (229), with_mask (277), without_mask (266) | No license |
| MAFA[199] | unknown | 30,811 | 35,806 masked faces with various orientations and occlusion degrees, at least one part of each face is occluded by mask | No license |
| MaskedFace-Net[200] | unknown | 137,016 | Generated masks on face images, "correctly masked" (67,193), "incorrectly masked" (69823) | CC BY-NC-SA 4.0 |

---

[189] https://www.kaggle.com/alexandralorenzo/maskdetection
[190] https://www.kaggle.com/ashishjangra27/face-mask-12k-images-dataset
[191] https://www.kaggle.com/dhruvmak/face-mask-detection
[192] https://www.kaggle.com/ivandanilovich/medical-masks-dataset-images-tfrecords
[193] https://www.kaggle.com/andrewmvd/face-mask-detection
[194] https://www.kaggle.com/mloey1/medical-face-mask-detection-dataset
[195] https://www.kaggle.com/spandanpatnaik09/face-mask-detectormask-not-mask-incorrect-mask
[196] https://www.kaggle.com/sumansid/facemask-dataset
[197] https://www.kaggle.com/wobotintelligence/face-mask-detection-dataset
[198] https://www.kaggle.com/vinaykudari/facemask
[199] https://imsg.ac.cn/research/maskedface.html
[200] https://github.com/cabani/MaskedFace-Net

| Name | Subjects | Images | Remark | License |
|---|---|---|---|---|
| MMD[201] | unknown | 6,000 | Acquired from the public domain, extreme attention to diversity (people of all ethnicities, ages, and regions). 20 classes of different accessories, classification of faces with mask, without mask, or with incorrectly worn mask. | CC0 1.0 |
| Moxa3K[202] | unknown | 3,000 | Combined images from Kaggle datasets and Internet images. Contains a few faces without masks. | No license |
| PWMFD[203] | unknown | 9,205 | Training set: with mask (6,702), without mask (9,680), incorrect mask (320) Validation set: with mask (993), without mask (791), incorrect mask (47) | No license |
| RMFD[204] | Part 1: 525, Part 2: 32,203, Part 3: 426 | Part 1: 95,000 Part 2: 510,000 Part 3: 4,015 | Part 1: Crawled sampled from the web, sorted, cleaned, labelled, 5,000 masked, 90,000 not masked Part 2: Simulated mask face, 10,000 masked, 50,0000 not masked Part 3: Real mask face verification, 3,589 pairs with same identity and 3,589 pairs of different identities (mask face/normal face) | No license |
| SMFD[205] | unknown | 1,570 | Generated masks on face images (785 mask, 785 no mask) | No license |
| Wider-Face[5] | 32,203 | 393,703 | 393,703 faces with high degree of variability in scale, pose and occlusion, expression, make up and illumination | No license |
| WMD[206] | unknown | 7,804 | Training set: 5,410 images, with 17,654 wearing masks; validation set: 800 images, with 1,936 wearing mask; testing set: 1,594 images, with 6,813 wearing masks. | Academic use only |

*Table 24: Summary of available datasets regarding face mask detection*

---

[201] https://humansintheloop.org/resources/datasets/medical-mask-dataset/
[202] https://shitty-bots-inc.github.io/MOXA/index.html
[203] https://github.com/ethancvaa/Properly-Wearing-Masked-Detect-Dataset
[204] https://github.com/X-zhangyang/Real-World-Masked-Face-Dataset
[205] https://github.com/prajnasb/observations
[206] https://github.com/BingshuCV/WMD

There are also some datasets with annotations on whether the subject is wearing eye glasses or not.

| Name | Subjects | Images | Mated Samples? | Remark | Constrained? | License |
|------|----------|--------|----------------|--------|--------------|---------|
| MeGlass[207] | 1,710 | 47,919 | yes | 14,832 face images with eyeglasses | No | MIT |
| VGGFace2[97] | 9,131 | 3.31 million | yes | Annotations for eyeglasses released as MAAD-Face Currently offline | No | CC 4.0, non-commercial use [157] |
| Colour FERET[208] | 994 | 11,338 | yes | 1,105 face images with eyeglasses, 350 of which are frontal | yes | Royalty-free |
| FRGCv2[55] | 569 | 45,000 | yes | 2,857 face images with eyeglasses | yes | Royalty-free |
| CelebAMask-HQ[209] | unknown | 30,000 | no | Segmentation masks specifying region of eyeglasses (if present) | no | Royalty-free, non-commercial use |

*Table 25: Datasets containing labels for eyeglasses*

---

[207] https://github.com/cleardusk/MeGlass
[208] https://www.nist.gov/itl/products-and-services/colour-feret-dataset
[209] https://github.com/switchablenorms/CelebAMask-HQ

## 11.3  Methods and Algorithms

There is abundant scientific literature on detection, recognition, and attribute estimation (e.g. expression) of partially occluded faces (e.g. masked faces). However, for an assessment of the face image quality, it is necessary to determine, whether the face is partially occluded and, preferably, which face parts (in particular, eyes, nose, mouth) are not visible. Henceforth, we denote this classification task as *facial occlusion detection*. In order to perform facial occlusion detection, the detection and localization of the face is a required pre-processing step and, thus, detection of faces under occlusion is required for facial occlusion detection. On the other hand, algorithms performing detection of faces under occlusion may also be able to perform facial occlusion detection as a side effect. In particular, if an algorithm detects **only** faces with certain occlusions (e.g. wearing a mask), it can be combined with standard face detector to decide if a face in an image is occluded or not.

The scientific literature on facial occlusion detection can be classified into literature on arbitrary facial occlusion detection and literature on face mask detection in the context of the Covid-19 pandemic.

### 11.3.1 Detection of Arbitrary Occlusions

Most publications on the detection of arbitrary facial occlusions aim for the detection of potentially criminal activities in surveillance footage, e.g. of ATM machines; a notable exception is [158], aiming to allow face recognition algorithms to accommodate occlusions. Some of these publications are restricted to the visibility of the mouth and nose (by masks or scarves), but others also consider occlusions of eyes and other face areas, e.g. by hats, helmets and sunglasses [159]. Because of the differences in the definitions of the classification problem (i.e. the type and extent of occlusion considered), but also due to different performance metrics applied in the evaluation (accuracy, precision, true positive rate (TPR), mean average precision, etc.) and different datasets used (most of which are not publicly released), the detection performance of the individual approaches can hardly be compared [159].

Many approaches try to detect occlusions by detecting certain facial components, in particular eyes and mouth. For instance, in [160], Suhr, et al. deploy a Viola-Jones object detector to detect candidates for both eyes, nose and mouth and use these to infer candidates for the face location. These candidates are then verified by extracting PCA features and applying a SVM classifier. The trained algorithm was evaluated using a dataset of 3,168 image sequences, collected using a real ATM machines, covering 21 different types of occlusions; the accuracies achieved ranges from 78.1% to 100% depending on the type of occlusion and whether the user was instructed to look into the camera.

In [158], Min, Hadid and Dugelay use features extracted with a Gabor filter to train two SVMs to classify, whether the face is occluded by a scarf or sunglasses (both SVMs were not trained on the two classes "occluded" and "not occluded"). The approach was evaluated on a set of 240 non-occluded face images, 240 images with sunglasses, and 240 images with scarfs, taken from the AR Face dataset; the detection rate was over 99% for the non-occluded faces and 100% for the occluded faces. However, the method does not include face detection step and was trained and tested on images that had already been tightly cropped to the face area.

Other approaches exploit the skin colour to detect occlusions. For instance, in [161], Kim, et al. propose a method that, after detecting the head by fitting its shape to an ellipse, estimates the *Skin Colour Area Ratio* (SCAR), the ratio of pixels in the face area with typical skin colour to determine whether the face is occluded or not; in order to distinguish the type of occlusion (e.g. sunglasses or mask), the SCAR is also computed for areas where the location of the mouth and the eyes is expected. This approach is applied on videos by computing the maximum SCAR value for all frames in the video. An evaluation on a dataset of 120 video sequences of 8 subjects (showing more or less frontal pose) gave a detection rate of 86.7%. However, the dataset seems not contain facial images of subjects with dark skin colour.

In [162], Zhang, et al. utilize a similar elliptical shape matching of the head contour to detect the face area in the first video frame, and a head tracking algorithm based on a Bayesian model to update the head's position throughout the video. As features, the skin colour area ratio was computed using a Gaussian mixture model to adapt the skin colour model to different illuminations, and a matching (using the structural similarity measure) of the face area to predefined face templates (patterns resembling facial features). On each of these two features, a naïve Bayesian classifier was trained to decide whether the face is occluded or not, and the outputs of these two weak classifiers were fused using an AdaBoost classifier. The classification was performed for each video frame independently without any aggregation. The method was evaluated on a dataset of 120 video sequences of 8 subjects including 12,120 frames[210] resembling ATM surveillance footage and comprising strong variations in pose; on this dataset, an accuracy of 98.56% has been achieved.

In recent years, some approaches have been published to deploy CNNs for detecting facial occlusions. In [163], Mao, Sheng and Zhang propose an algorithm to classify, whether the face is occluded or not. First, the face is detected by minimization of an energy function measuring the match of the shape of head and shoulders with a Gaussian distribution function. Then, they apply a sparse representation classification based on deep features extracted from the detected faces to decide whether the face is occluded or not. The evaluation applied on the individual frames of 120 CCTV video sequences of 8 subjects gave a detection rate of occluded faces of 97.35% and a false alarm rate of 1.86%. However, some aspects of this publication are dubious: It is not explained, which CNN model was deployed for feature extraction and how it was trained; furthermore, the "state of the art methods" with which the performance of the new method is compared to entirely are algorithms for face recognition and not for detection of face occlusions.

In [164], Xia, Zhang and Coenen train two CNNs comprising 10 layers for the detection of the head and the classification of the occlusion type. For training and evaluation, three datasets were used:

- A newly created *Face Occlusion dataset* of 1,320 CCTV video sequences of 220 subjects, comprising 30 seconds and 750 frames each. This dataset has not (yet) been published.
- The AR Face Dataset comprising more than 4,000 images of 126 subjects.
- 1,000 images of the Labelled Faces in the Wild (LFW) Dataset, were black patches (rectangles) were applied as simulated occlusions over one or both eyes, the nose and/or the mouth.

---

[210] Even though the number of video sequences, subjects and frames are identical to that of the test data used in [157], these two Datasets are completely different.

The head detection CNN was pre-trained using 10% of the images from the ImageNet dataset, and was then fine-tuned on the original (uncropped) images of the three datasets listed above. The occlusion detection is pre-trained (with respect to face recognition) on a subset of the cropped images of the AR Face dataset; after pre-training the classification layer is exchanged and the CNN is fine-tuned using the Face Occlusion Dataset on predicting the facial parts (left eye, right eye, nose, mouth) that are occluded. Their evaluation of the head detector gives an accuracy of 100%, 97.58% and 85.61% for the modified LFW, the AR Face and the Face Occlusion Dataset, respectively. For the classification, they obtain an accuracy of 95.41%, 98.58% and 94.55% for the modified LFW, the AR Face and the Face Occlusion Dataset, respectively. In [165], the authors (Xia, Zhang and Coenen) apply the same method for occlusion detection and report even higher accuracies (97.24% for the Face Occlusion Dataset). However, in [165] no details on the head/face detection method applied are given.

In principle, face segmentation methods that segment visible parts of the face, e.g. the CNNs of Saito et al. [24], Nirkin et al. [25], and Yin et al. [27] (see Section 2.2.2) can also be used in combination with facial landmarks to detect arbitrary occlusions and to identify the occluded face parts. However, it is not clear, how reliable an occlusion detection based on this segmentation and facial landmarks would be, because landmarks may not be accurately determined in occluded areas of the face. Furthermore, methods that exclude beards from the segmentation mask, e.g. that of Masi et al. [26], are not eligible for this purpose. The CNN of Yin et al. has even been trained to detect occlusion by frames of eyeglasses; however, no example images have been found that show the effectiveness of the detection of eyeglasses.

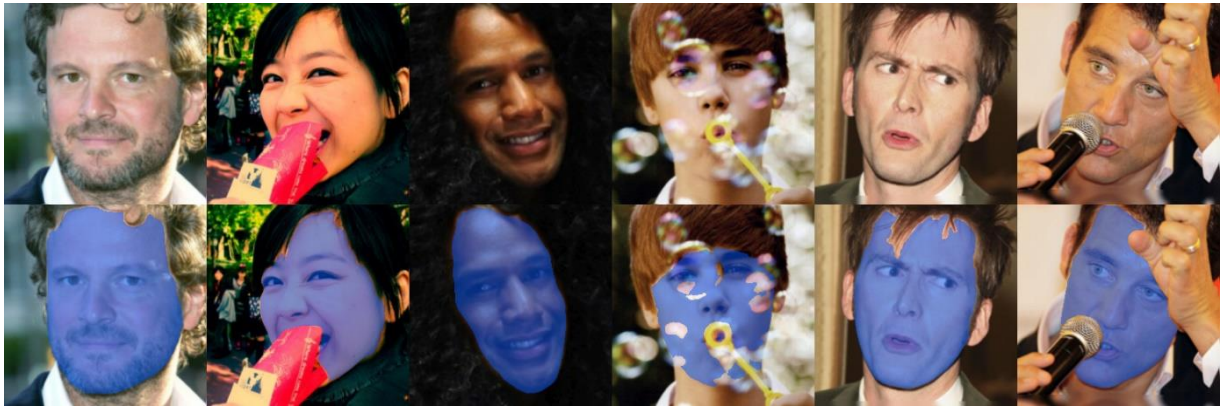

*Figure 24: Example segmentations generated by the CNN of Yin et al. [27]*

Face segmentation in (externally) occluded and un-occluded areas is also performed as an intermediate step by many methods for *face de-occlusion*, e.g. by Xu et al. [166], Zhao et. al [167], Song et al. [168], Yuan et al. [169], Yin et al. [170], Cai et al. [171], or Zhang et al. [172]. Other approaches for face de-occlusion do not explicitly identify the occludes face regions, e.g. [173] or [174], but since visible artifacts are typically restricted to the generated (previously occluded) image regions, they could also be used to detect occlusions by comparing the original (input) image with the output image.

The following table summarizes the described approaches by specifying what cases the algorithms can distinguish and which head/face detection approach is applied.

| Publication | Output | Head/face detection method |
|---|---|---|
| [160] | Occlusion (Yes, No)<br>Occlusion = at least one eye and mouth are visible | Viola-Jones |
| [158] | Occlusion (Yes, No)<br>Occlusions = scarf of sun glasses worn | None |
| [161] | Eye Occlusion (Yes, No), Mouth Occlusion (Yes, No)<br>Segmentation of occluded area | Shape fitting of head |
| [162] | Occlusion (Yes, No)<br>Occlusions by hats, sunglasses and masks | Shape fitting of head |
| [163] | Occlusion (Yes, No)<br>Mouth occlusion, head occlusion and other face occlusion) | Shape fitting of head & shoulders |
| [164], [165] | Occlusion (Yes/No) per facial part: left eye, right eye, nose and mouth | Dedicated CNN |
| [27] | Areas of the un-occluded parts of the face | Implicitly done by the CNN |
| [24], [25], [27] | Segmentation mask identifying occluded face regions | Implicitly done by the CNN |
| [166], [167], [168], [169], [170], [171], [172]. | Segmentation mask identifying occluded face regions, Image of un-occluded face | Implicitly done by the CNN |
| [173], [174] | Image of un-occluded face | Implicitly done by the CNN |

*Table 26: Summary of the described approaches for face occlusion detection*

## 11.3.2 Detection of Occlusions by Face Masks

With the advent of the COVID-19 pandemic, a large number of methods have been published to classify if faces in images are wearing a medical face mask. Some methods do not only check, if a subject is wearing a mask but also, if the mask is worn correctly, i.e. they perform a three-class classification (no mask, correct mask, incorrect mask). The corresponding research is mainly motivated by the goal to monitor and enforce the wearing of masks as a countermeasure to spreading of the coronavirus. Face mask detection methods can be used to detect occlusions of mouth and nose but not of the eyes.

Almost all publications deploy CNNs. Some approaches apply two separate algorithms for face detection and face mask classification, often deploying face detectors from the literature, while others perform both tasks in a single step. In order to cope with the limited amount of training data, pre-trained CNN models are typically adapted by transfer learning. Because of the different performance metrics applied in the evaluation (accuracy, precision, TPR, mean average precision, etc.) and different datasets used, the detection performance of the individual approaches can hardly be compared [175].

In [176], Lin at al. apply a modified version of LeNet-5, a CNN with 7 layers introduced in [177] for character recognition, to decide if an image shows a face wearing a mask. Since the CNN processes image parts of fixed size (28x28), the input window is sampled from multiple scales (pyramid matching) and positions (sliding window) to cover all potential locations and sizes of faces. First, the CNN was first pre-trained on the character recognition using the MNIST dataset; then transfer learning for masked face detection was applied. As training and test data, images from surveillance videos were used comprising 240 images of masked faces and 900 of unmasked faces. In order to accommodate the small training set, the training data was augmented by horizontal flipping. The evaluation gave a precision of 68% and a recall of 85%.

In [178], Loey et al. deploy a CNN to extract features that are then classified using three different classifiers (alternatively): SVM, decision tree and an ensemble of a K nearest neighbor and Linear Regression and Logistic Regression. As CNN, a ResNet-50 model is obtained by transfer learning using a residual learning approach. For training and testing three datasets were used: The Real-World Masked Face Dataset (RMFD), the Simulated Masked Face Dataset (SMFD), and Labelled Face in the Wild (LFW) with simulated masks applied to the images. For the SVM and the ensemble classifier, accuracies over 99% have been achieved depending on the datasets used for training and testing. However, many aspects are not documented and remain unclear. For example, it remains unclear how the faces are detected in the image, on which task and how the CNN was trained in the first place, and on which task and how the transfer learning of the CNN was done.

In [179], the same authors (Loey et al.) train a combination of 2 CNNs, ResNet-50 (for feature extraction) and YOLOv2 (for object detection and classification) to detect masked faces. YOLO stands for "You only look once" and is a one-stage approach to detect and classify one or several objects within an image, making a separate face detector obsolete. YOLOv2 does not deploy any fully connected layers, making it very fast. For training and testing, the authors combine the Medical Masks Dataset (MMD), comprising 682 pictures with over 3,000 faces wearing masks, and the Face Mask Dataset (FMD), comprising 853 images of faces with or without masks. Their evaluation resulted in an average precision of 81%.

In [180], Qin and Li train a MobileNet-v2 CNN to distinguish between no-facemask-wearing, correct-facemask-wearing and incorrect-facemask-wearing. The CNN model was trained in three steps: First, on general object detection using the ImageNet Dataset; secondly, on face recognition on the CASIA Web Face Dataset; and finally, it was fine-tuned using the Medical Masks Dataset (MMD) containing 3,835 images on the mask-wearing classification. In order to better deal with low resolution images, an image super-resolution network trained on CelebA is applied prior to the classification CNN. For face detection, alignment and cropping the authors apply the MTCNN face detector, which uses a cascaded CNN for detection of faces on various scales. Depending on the mask type (FFP2 or surgery mask), the method achieved an accuracy of 98.02% and 98.99% on the test set of MMD; however, these statistics only refer to the classification CNN, and do not cover the face detection method, for which no accuracy measures are indicated.

In [181], Jiang, Fan and Yan utilize a so-called *Feature Pyramid Network (FPN)* based on the first ResNet50 layers to detect the face in the image and to classify, if the face is wearing no mask, correctly wearing a mask or incorrectly wearing a mask. FPN feed features from several convolutional layers to the final layers to allow the detection of the relevant patterns (e.g. a masked face) at different scales; in this case, each convolutional layer is supposed to detect faces of different sizes. The CNN model is first trained on general face detection using the Wider Face Dataset, and then adapted to the detection and classification of (potentially) masked faces by transfer learning. For transfer learning and testing, two datasets were used: MAFA-FMD, a version of the MAFA Dataset with additional annotations (which have been published by the authors) and the AIZOO Face Mask Dataset. The trained model was coined RetinaFaceMask, and a computationally less demanding version RetinaFaceMask-Light was also trained based on MobileNetV1. A mean average precision of 94.8% for RetinaFaceMask and of 92.0% for RetinaFaceMask-Light.

In [182], Nagrath et al. propose to use SSD (Single Shot Multibox Detector) CNN based on a ResNet-10 backbone to detect the faces in the image and a MobileNet-v2 to predict if the face is wearing a mask or not. The authors coin their model SSDMNV2. For training and testing, the authors combined datasets from Kaggle (real masks) and PyImageSearch (simulated masks) and other (unspecified) sources, resulting in a class-balanced dataset of approximately 11,000 images. In their evaluation, they compare SSDMNV2 with other CNNs (LeNet-5, AlexNet, VGG-16, ResNet-50) trained using the same data. SSDMNV2 achieves the best performance with an accuracy of 92.64%, a precision of 94% and a recall of 93%.

In [183], Yu and Zhang deploy the YOLOv4 CNN to detect faces and classify the mask wearing (face, face mask, wearing masks irregularly) in a single step. The authors apply some modifications to the CNN structure of YOLOv4 to improve its computational and detection performance. As training and test data, they compiled a custom dataset of 10,855 images from the RMFD and MaskedFace-Net Datasets. Since RMFD does not contain labels on whether the mask is worn correctly, the authors determined these labels manually, classifying faces where parts of the chin are uncovered as "wearing masks irregularly". For evaluation purposes, the authors also trained versions with the original YOLOv4, YOLOv3, SSD (Single Shot Detector, another single-stage object detection CNN), and Faster R-CNN. Their improved YOLOv4 CNN achieves the highest mean average precision of 98.5% (precision: 95.1%, recall: 98.2%) while being faster than the other constructions. Without providing any data, the authors also claim that their approach is more efficient than those of [178], [179], and [182].

In [184], Talahua et al. proposed a two-stage approach to detect face masks: First, the face is detected using a deep learning-based face detector in OpenCV (presumably, the SSD-based face detector described in Section 2.1). Second, a MobileNetV2 is applied to classify, whether the subject is wearing a mask or not. For training and testing, approximately 13,000 self-collected images were used, 20% of which were used for testing. In their evaluation, their approach achieves an accuracy of 99.65%, precision of 99.09%, sensitivity of 99.77%, and specificity of 99.6%; according to their definition of false negatives, these metrics also cover failures of the face detector.

In [156], Batagelj et al. also follow a two-stage approach to detect if face masks are worn correctly. For the detection of the face, seven open-source face detectors, some of which specialized on masked faces, were evaluated with respect to the detection of masked faces: MTCNN, the SSD face detector implemented in OpenCV, the Dual Shot Face Detector (an improved variant of the SSD face detector), RetinaFace (see Section 2.1), and the face detectors used in the open source face mask detectors from Baidu's PaddlePaddle framework, the GitHub repository InsightFace (coined RetinaFace AntiCov, not available anymore), and AIZOOTech. For the second stage, several CNN architectures were trained on classification whether face-masks are worn correctly or not: AlexNet, VGG-19, ResNet (with various numbers of layers), SqueezeNet (V1.0 and V1.1), DenseNet (with various numbers of layers), GoogLeNet, MobileNetv2. For training and evaluation, they compiled and publicly released a new dataset MFLD comprising more than 40,000 images from the MAFA and WiderFace Datasets, adding annotations on whether masks are worn correctly, gender, pose and ethnicity. In their evaluation of the face detectors, RetinaFace performed best, while the evaluation of the classifiers showed a clear trade-off between computation time and accuracy. Finally, the authors combined the RetinaFace dface detector with the classifier of best accuracy (ResNet-152) and compared the resulting method with the face mask detectors of PaddlePaddle, InsightFace, and AIZOOTech. Their evaluation shows that their solution achieves a mean Average Precision of over 95% and by far outperforms the face mask detectors of PaddlePaddle, InsightFace, and AIZOOTech; however, it must be considered, that the latter three had been trained to detect whether faces are wearing masks or not, while the evaluation distinguished (and the author's solution was trained on), whether the face was correctly covered by a mask or not, where the second class ("not compliant") covered faces with incorrectly worn masks and faces without masks. The classification objective of the proposed solution differs considerably from the goal to detect significant occlusions of the mouth and nose. This publication is mainly of interest because of its comparison of face detectors and face mask detectors.

The following table summarizes the described approaches by specifying what cases the algorithms can distinguish and which face detection approach is applied.

| Publication | Output | Face detection method |
|---|---|---|
| [176] | Mask or No Mask | Sampling windows at various scales and positions |
| [178] | Mask or No Mask | unspecified |
| [179] | Mask or No Mask | Integrated in CNN |
| [180] | Correct Mask, Incorrect Mask, No Mask | MTCNN |
| [181] | Correct Mask, Incorrect Mask, No Mask | Integrated in CNN |
| [182] | Mask or No Mask | Dedicated CNN |
| [183] | Correct Mask, Incorrect Mask, No Mask | Integrated in CNN |
| [184] | Mask or No Mask | SSD face detector in OpenCV |
| [156] | Correct Mask, Not A Correct Mask (either Incorrect Mask or No Mask) | RetinaFace |

*Table 27: Summary of described approaches for face mask detection*

## 11.3.3 Detection of Occlusion by Frames of Glasses

No publications were found on the detection of occlusions of the eyes or the EVZ by frames or rims (in case of frameless eyeglasses) of eyeglasses in facial images. However, since state of the art algorithms for facial landmark localization can detect the eye's landmarks quite accurately even in the presence of slight occlusions (see Section 2.3.2), it suffices to localize the rims of the eyeglasses.

Very few publications that focus on the localization of eyeglasses have been found. In [185], Borza et al. apply edge detection to extract the shape of potential glasses, which is then compared to a dataset containing different shapes of eyeglasses. On their own dataset of frontal face images, their method successfully extracts the contour of the eyeglasses in 92.3% of cases, where success is defined by an overlap of at least 95% between the extracted contour and the ground truth. Examples results of the method are shown in Figure 25.

Many publications propose methods for the removal of eyeglasses in facial images. These algorithms

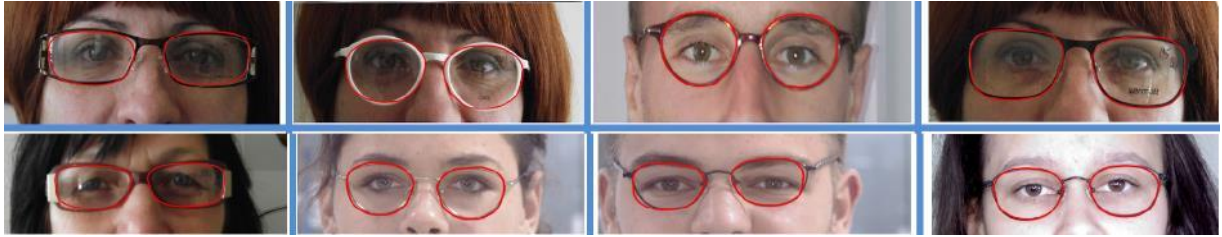

*Figure 25: Example results of the eyeglass contour detection method from [185]*

can also be used to localize frames and rims:

- Many algorithms perform localization and segmentation of frames, explicitly as a processing step.
- By comparing the original images with the image where the glasses have been removed, the occlusions by the frames and rims of the glasses can be localized. However, this pixel-wise difference would also represent shadows casted by the eyeglasses as well as reflections.

Early approaches for eyeglasses removal are based on classical methods of statistical learning while more recent publications use deep learning. However, there exist no benchmarks for the task of eyeglasses removal, thus, a fair comparison of the effectiveness of the individual methods is impossible. Therefore, we present some example images from the respective publications to give an impression of the results.

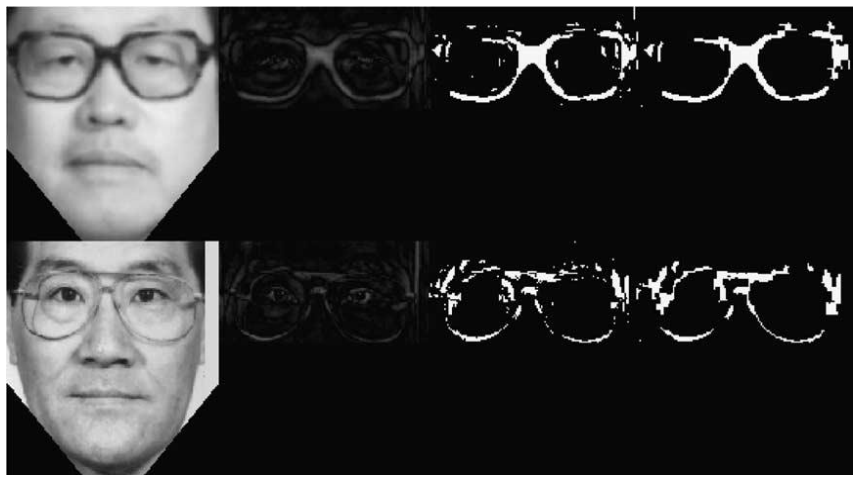

*Figure 26: Results of the segmentation of eyeglasses presented in [186]*

In [186], Du and Wu apply the *Eigenfaces approach*, i.e. the construction of an eigenspace representation of facial images from a training set using *Principal Component Analysis* (PCA), to remove the glasses from the target image. By applying adaptive binarization, a segmentation map of the frames of the glasses (and shadows casted by them) is obtained. However, the segmentations shown in the publication are not very accurate.

In [187], Wu et a. used a similar approach: Based on the eigenspace distributions of the images with eyeglasses and the images without eyeglasses in the training set, a statistical mapping is determined. In order to apply this mapping, the glasses are located by an active shape model that determines key points (landmarks). The results presented by the authors look quite appealing, however, they also show cases, where their method fails due to "significant highlights in these [...] images [which] were not present in the training data".

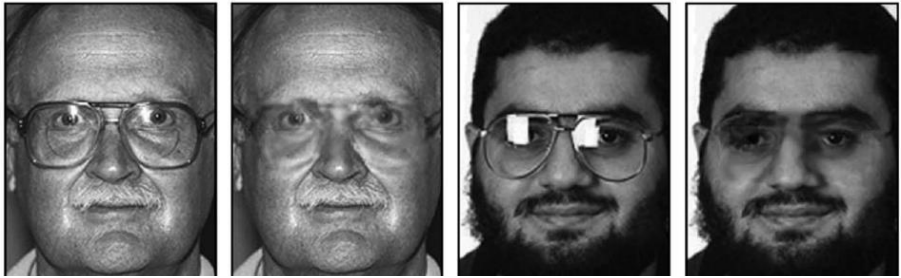

*Figure 27: Failure cases of the method from [187]*

Recent publications on eyeglasses removal apply deep learning, in particular *auto-encoders* or *generate adversarial networks* (GAN). Many of these publications applied so called *image-to-image translation* architectures, which can be trained on virtually any kind of modification of images, to the task of eyeglass removal, e.g. CycleGAN [188], UNIT [189], MUNIT [190], Council-GAN [191], ACL-GAN [192], U-GAT-IT [193] or UVCGAN [194]; in many cases, the results are not completely convincing.

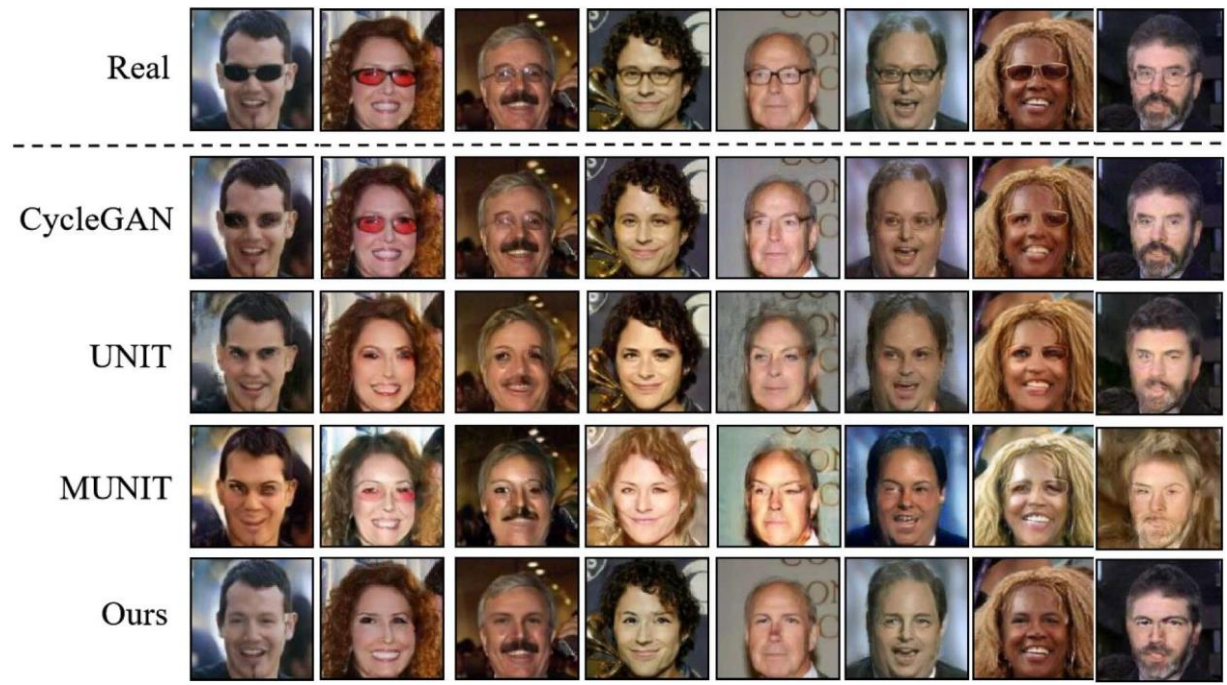

*Figure 28: Comparison of results of ERGAN (bottom row) compared with those of some general image-to-image-translation networks trained on eyeglass removal (from [195])*

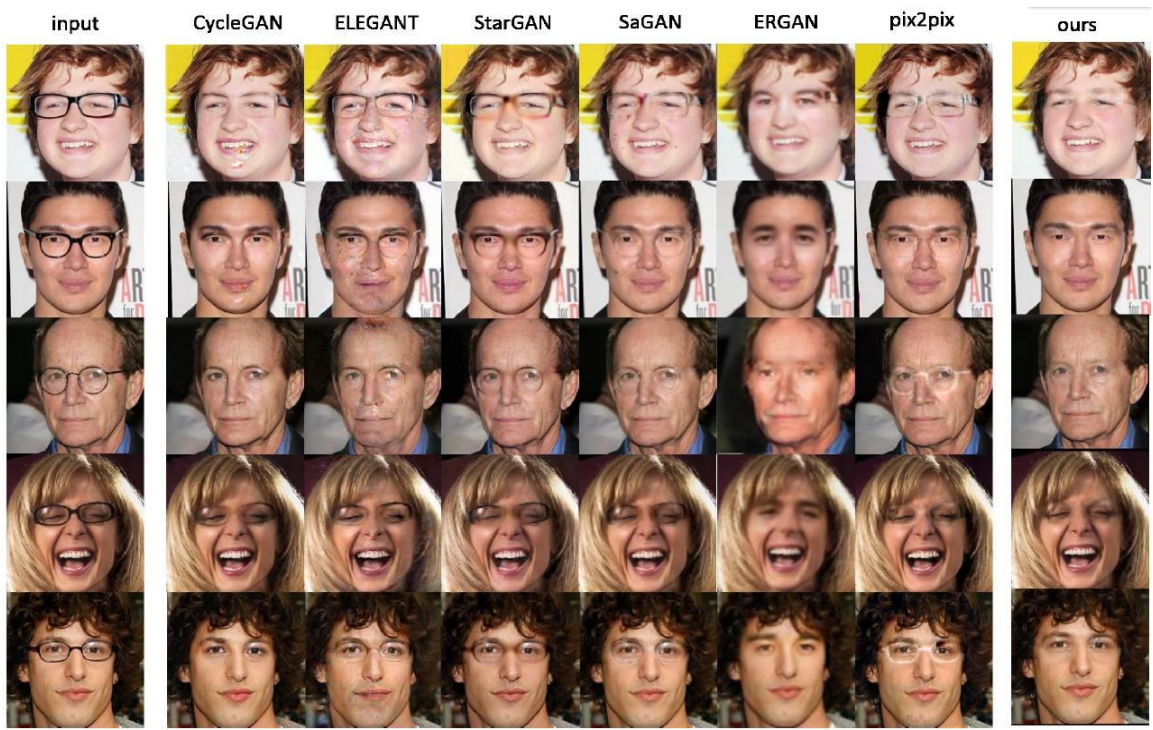

*Figure 30: Results of ByeGlassesGAN (rightmost column) compared with other solutions for eyeglasses removal (taken from [78])*

Other publications proposed solutions specialized on eyeglass removal, e.g. ByeGlassesGAN [78], ERGAN [195] or MADL [196]. The results of ERGAN appear a bit blurry (especially when other researchers reproduce the training), but ByeGlassesGAN and MADL seem to achieve good results, see Figure 30 and Figure 29.

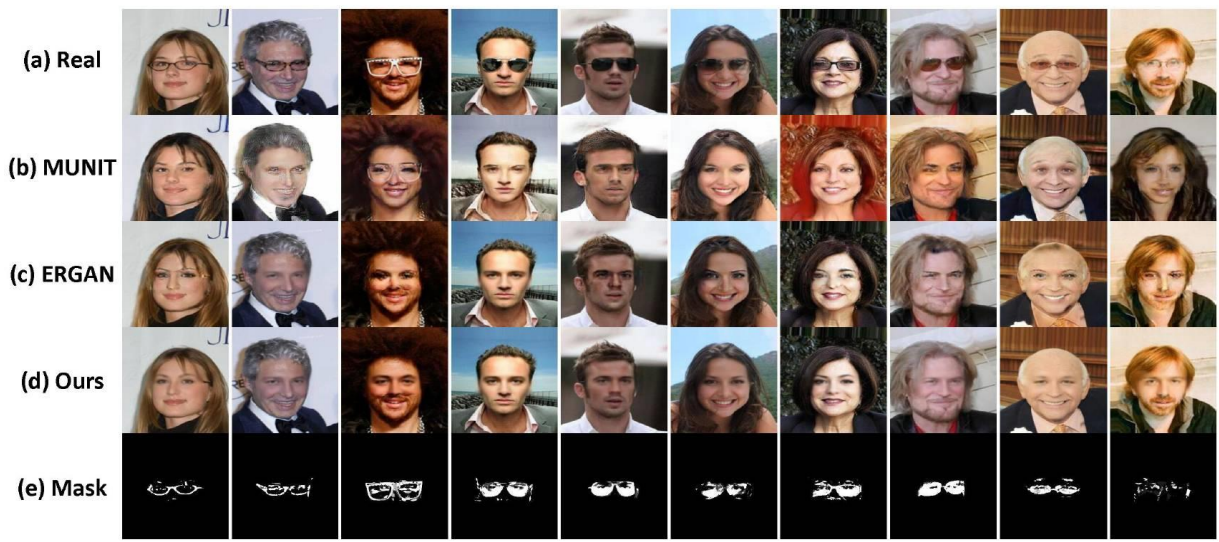

*Figure 29: Results of MADL ("Ours") compared with other solutions for eyeglasses removal, taken from [196]. The mask in the bottom row is simply the magnitude of the difference between original and output image.*

## 11.4   Software

### 11.4.1 Detection of Arbitrary Occlusions

Besides the implementations for face segmentation that identify the visible parts of the face (see Footnotes 40, 45, 46), the following software implementations for arbitrary occlusion detection have been found, some of which perform segmentation of occluded face regions as intermediate step in the course of de-occlusion.

| Name | Model size | Remarks | License |
|------|-----------|---------|---------|
| Face-Occlusion-Detect[211] | 59 MB | Outputs occluded area<br>Github issues report poor performance of published models | No license |
| Face-Deocclusion[212] [172] | 368 MB[213] | Segmentation CNN outputs occluded area | No license |
| Face-Occlusion-Detection[214] | n.a. | Uses Haar cascade classifier to detect eyes, mouth and nose | No license |

*Table 28: Summary of relevant GitHub repositories regarding occlusion detection*

### 11.4.2 Detection of Sunglasses

The following table provides information (including evaluation metrics) about repositories found on GitHub implementing sunglasses detection.

| Name | Model size | Evaluation results given | License |
|------|-----------|--------------------------|---------|
| ATM-Sunglasses-Detection[215] | 1 MB | Accuracy: 0.999 | No license |

*Table 29: Summary of relevant GitHub repositories regarding sunglasses detection*

### 11.4.3 Face Mask Detection

The following table provides information about repositories found on GitHub implementing face mask detection. Compared to sunglasses detection, there seems to be a significantly higher number of projects about face mask detection. This could be based on the current pandemic COVID-19 outbreak. However, it is very difficult to compare the detection performance of these implementations merely from the reported numbers since different datasets and various metrics are used.

---

[211] https://github.com/Oreobird/Face-Occlusion-Detect
[212] https://github.com/KaiyuanWan/Face-Deocclusion
[213] The size if the segmentation (face parsing) CNN is specified, as the CNN for impainting (hallucination of occluded parts) is not needed for occlusion detection.
[214] https://github.com/ankit-dhokariya/Face-Occlusion-Detection
[215] https://github.com/jw2533/ATM-Sunglasses-Detection

| Name | Model size | Evaluation results given | License |
|---|---|---|---|
| AIZOOTech[184] | 4 MB | AP Mask: 0.919,<br>AP No mask: 0.896 | MIT |
| Ayushparikh-code[216] | 10 MB | Accuracy: 0.98 | MIT |
| Backl1ght[217] | 245 MB | AP Mask: 0.9604,<br>AP No mask: 0.9484,<br>mAP@0.50: 0.9544,<br>L-AMR Mask: 0.05,<br>L-AMR No mask: 0.1 | GNU General Public License v3.0 |
| Chandrikadeb7[185] | 10 MB | **Mask:**<br>Precision: 0.99,<br>Recall: 0.86,<br>F1-Score: 0.92<br>**No Mask:**<br>Precision: 0.88,<br>Recall:0.99,<br>F1-Score: 0.93 | MIT |
| Karan-Malik[218] | 11 MB | Accuracy Train: 0.982,<br>Accuracy Test: 0.973 | MIT |
| MINED30[219] | 235 MB | AP Mask: 0.9696,<br>AP Improperly: 0.928,<br>AP No mask: 0.9237,<br>mAP@0.50: 0.9404,<br>conf_thresh: 0.25,<br>precision: 0.88,<br>recall: 0.95,<br>F1-score: 0.92,<br>average IoU: 0.6955 | No license |
| MINED30[219] | 235 MB | AP Mask: 0.9768,<br>AP Improperly: 0.9154,<br>AP No mask: 0.8657,<br>mAP@0.50: 0.9193,<br>conf_thresh: 0.25,<br>precision: 0.90,<br>recall: 0.94,<br>F1-score: 0.92,<br>average IoU: 0.7184 | No license |
| achen353[220] | 11 MB | Accuracy: 0.99 | MIT |
| Hott-J[221] | 10 MB | **Mask:**<br>precision: 0.99,<br>recall: 0.86,<br>f1-score: 0.92 | MIT |

---

[216] https://github.com/Ayushparikh-code/Face--Mask--Detection
[217] https://github.com/Backl1ght/yolov4_face_mask_detection
[218] https://github.com/Karan-Malik/FaceMaskDetector
[219] https://github.com/MINED30/Face_Mask_Detection_YOLO
[220] https://github.com/achen353/Face-Mask-Detector
[221] https://github.com/Hott-J/Face-Mask-Detection

| Name | Model size | Evaluation results given | License |
|------|-----------|-------------------------|---------|
|  |  | **No mask:**<br>precision: 0.88,<br>recall: 0.99,<br>f1-score: 0.93<br>**Accuracy:** 0.93 |  |
| PaddlePaddle[222] | 1 MB | **Mask:**<br>AP@0.50: 0.6263±0.0216,<br>AP@0.40:<br>0.6912±0.0198<br>**No Mask:**<br>AP@0.50:<br>0.7834±0.0201,<br>AP@0.40:<br>0.8492±0.0142<br>**mAP@0.50:**<br>0.7049±0.0136<br>**mAP@0.40:**<br>0.7702±0.0132[223] | Apache 2.0 |
| xinqi-fan[224] | 4 MB | **AIZOO:**<br>AP Mask: 0.94,<br>AP No mask: 0.936,<br>mAP@0.50: 0.938<br>**Moxa3K:**<br>AP Mask: 0.7852,<br>AP No mask: 0.5556,<br>mAP@0.50: 0.6704 | MIT |
| ksdkamesh99[225] | 58 MB | Accuracy: 0.98,<br>F1-Score: 0.97 | MIT |
| FarhanSadaf[226] | 179 MB | Accuracy: 0.9681 | MIT |
| Saisree-123[227] | Own model:<br>33 MB<br>ImageNet:<br>11 MB | **Own model:**<br>Accuracy (without augmentation): 0.948,<br>Accuracy (with augmentation): 0.975<br>**ImageNet:**<br>Accuracy: 0.9917 | No license |

*Table 30: Summary of relevant Github repositories regarding face mask detection*

---

[222]
    https://github.com/PaddlePaddle/PaddleHub/tree/release/v2.2/modules/image/face_detection/pyramidbox_lite_mobile_mask

[223] The average precisions are referenced from [149] and the metrics were evaluated on the FMLD Dataset, as no metrics from the provider of the model is released as of right now

[224] https://github.com/xinqi-fan/SL-FMDet

[225] https://github.com/ksdkamesh99/Face-Mask-Detection

[226] https://github.com/FarhanSadaf/face-mask-detection

[227] https://github.com/Saisree-123/Face_mask_detection

### 11.4.4 Eyeglasses Removal (Eyeglasses Detection)

No implementations were found for the localization or segmentation of frames and rims of eyeglasses. However, as described in Section 11.3.3, occlusions by frames of eyeglasses could be detection by applying software for eyeglasses removal and comparing its output to the original facial image. Therefore, we list the relevant implementations for eyeglass removal.

The following table lists the most relevant implementations found for eyeglasses removal. No information about the accuracy of these methods is provided.

| Name | Model size | License |
|------|-----------|---------|
| glasses-removal-gan[228] | 50 MB | MIT |
| remove-glass[229] | 44 MB | No license |
| take-off-eyeglasses[230] | 136 MB | No license |
| eyeglassesRemoveforFaceRecognition[231] [197] | 11 MB | MIT |

*Table 31: Summary of relevant GitHub repositories for eyeglasses removal*

---

[228] https://github.com/lecomte/glasses-removal-gan
[229] https://github.com/ash11sh/remove-glass
[230] https://github.com/StoryMY/take-off-eyeglasses
[231] https://github.com/wangxiangxue/eyeglassesRemoveforFaceRecognition

# 12 Inter-Eye Distance

The *inter-eye distance* (IED) is defined as the distance between the centres of the eyes typically measured in pixels (cf. ISO/IEC 39794-5:2019). The IED is affected by the camera subject distance (see Section 9) and the resolution of the acquired facial image.

## 12.1 Impact on Face Recognition

The IED correlates with facial information contained on the image: The higher the IED is, the more information can be extracted from the facial image that can be used for facial recognition and other pre-processing scenarios, such as *presentation attack detection* or *morphing attack detection*. Specifically, the ISO/IEC 19794-5:2011 suggests to prefer an IED of 120 pixels (or more) as a best practice for an image of *frontal image type* and the ISO/IEC 39794-5:2019 specifies that the image shall be at least 90 pixels. The following figure, taken from table D.4 from the ISO/IEC 39794-5:2019 depicts the IED capturing requirements and recommendations.

| Criterion: | Requirement | IED ≥ 90 pixels |
|---|---|---|
| live capture IED | Best practice | IED ≥ 240 pixels |
| Criterion: | Requirement | IED ≥ 90 pixels |
| scanned image IED | Best practice | IED ≥ 240 pixels |
| Criterion: | Requirement | IED ≥ 90 pixels |
| electronic submission IED | Best practice | IED ≥ 240 pixels |
| Criterion: | Requirement | IED ≥ 90 pixels |
| issuer repository IED | Best practice | IED ≥ 240 pixels |
| Criterion: | Requirement | IED ≥ 90 pixels |
| MRTD chip storage IED | Best practice | IED ≥ 120 pixels |

*Figure 31: IED capturing requirements and recommendations (taken from ISO/IEC 39794-5:2019)*

It is important to note, that a reasonable IED alone does not imply that the subject's face is well presented on the image and that a reasonable amount of facial information is provided. For instance, if the image has cropped to achieve a reasonable IED, the facial information would not increase. However, it seems plausible that in conjunction with assessing blurriness or sharpness metrics (see Section 4), the IED is a useful feature of which assessment can predict the capability of recognizing the face.

## 12.2 Datasets

Since a successful landmark detection of the eyes is required for the evaluation of the inter-eye-distance, in general, the same datasets which are referenced in Section 2.2 can also be used for evaluating the performance of an implementation of an IED estimation.

## 12.3 Methods and Algorithms

Given the eye centres, e.g. resulting from a previous landmark detection, the computation of the IED is straightforward and may be performed as follows.

1. Computation of the eye centres via four landmark points.

2. Computation of the Euclidean distance D between the eye centres.

It is important to note, that the computed eye centre in step 1 is not necessarily the centre of the pupil.

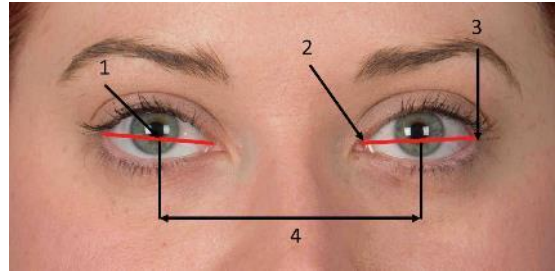

*Figure 32: Visualization of a IED measurement (taken from ISO/IEC 39794-5:2019)*

The following figure visualizes the computation of the IED, where (1) is the eye centre, (2) the inner canthus, (3) the outer canthus and (4) the IED.

# 13      Location and Coverage of the Head

ISO/IEC 39794-5:2019 requires that "the head shall be centred in the final face portrait" and defines measures for the optimal positioning of the head and face in the image, visualized in Figure 33. If the position of the crown, chin or ears cannot be determined exactly due to poor positionings, ISO/IEC 39794-5:2019 suggests that "a good guess shall be made".

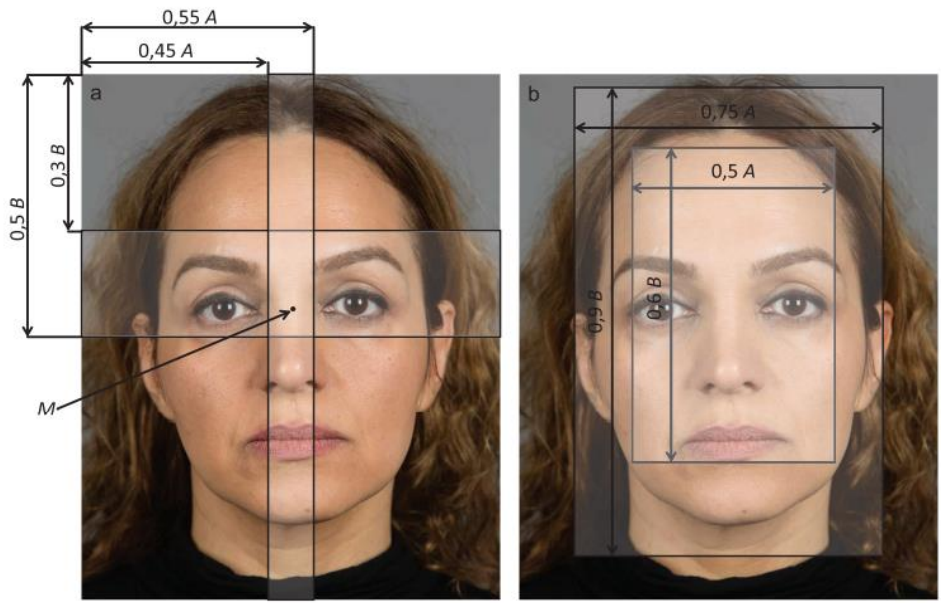

*Figure 33: Minimal and maximal optimal head dimensions from ISO/IEC 39794-5:2019 D.1.4.4*

These measures aim to exclude images, where the head or face region is either not entirely included (i.e. partially outside) in the image or extends so close to the image border.  Figure 34 visualises a good and poor face positioning in an image.

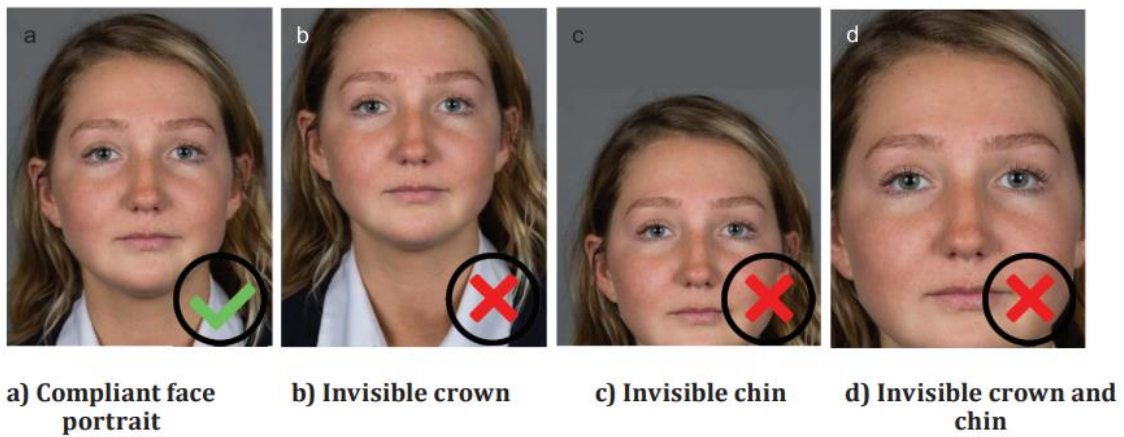

*Figure 34: Acceptable and unacceptable head location examples from ISO/IEC 39794-5:2019 D.1.4.4*

ISO/IEC WD4 29794-5:2022 includes two quality metrics related to the location and coverage of the face:

- **Horizontal position of the face**: Horizontal regulation of face size and positioning, to avoid faces being cut off or too close to the edge of the image.
- **Vertical position of the face**: Vertical regulation of face size and positioning, to avoid faces being cut off or too close to the edge of the image.

The descriptive text of these metrics refers to the respective section of ISO/IEC 39794-5:2019, but only specifies an algorithm to compute the position of the midpoint of the eyes, and no method for estimating the size of the face relative to the image size or its coverage by the image.

In contrast, ISO/IEC 19794-5:2011 and BSI TR-03121-3 Vol. 1 not only specify where the face (precisely, the midpoint of the eyes) should be located but also limits for the size of the head relative to the image.

In some publications, e.g. [148], partial coverage of the face by the image is considered as a case of occlusion. However, in this present report, we distinguish between partial occlusions by objects and cases, where the face or head region is entirely covered by image.

## 13.1   Impact on Face Recognition

It is obvious that if parts of the face are missing in a facial image, the corresponding information (features) cannot be extracted and therefore not been used for face recognition. This will negatively affect recognition performance. The extend of deterioration will depend on the importance of the face parts that are not visible; for instance, a missing forehead may be less critical as a mouth outside the image area.

Moreover, if the head region extends too close to the image border, face detection and facial landmark localization may fail; however, state of the art algorithms for face detection and facial landmark localization are also able to detect faces that are only partly visible (in this case, the coordinates defining the bounding box of the face can lie outside the image area).

## 13.2   Datasets

Current research has shown that no existing datasets with corresponding ground truth labels are available for this specific scenario. However, such a dataset can be easily created, by cropping facial images vertically or horizontally in dependence of the determined location of the face/head.

## 13.3   Methods and Algorithms

By constraining the positions of the left and right eye to a certain position in the image (far enough from the image borders), it can be assured that the head (including the hair) is completely covered by the image. The positions of the eyes can be easily determined using facial landmarks extractors (see Section 2.2). Since current landmark extractors can also extrapolate landmarks outside the image area, this approach will even work in cases, where the face is not fully covered by the image.

However, if as in ISO/IEC WD4 29794-5:2022, ISO/IEC 19794-5:2011 and ISO/IEC 39794-5:2019, only the position of the midpoint between the eyes is constrained, it is necessary to also constrain the size (width and height) of the head relative to the image dimension. Since facial landmark extractors do not compute landmarks for the ears and the crown of the head (which are used in ISO/IEC 39794-5:2019 to define the width and height of the face, respectively), it is necessary, to estimate these points (or, directly, the size of the head) from the position of the eyes and the face contour. Since the ratio between the inter-eye-distance and the size of the head varies between individuals, this estimation should be done carefully.

Alternatively, a face detection algorithm (see Section 2.1) could be used to compute a bounding box of the face, and based on the position and size of that bounding box it is possible to roughly estimate if the head is fully covered by the image. However, it needs to be considered that the typical size of the bounding box relative to the size of the face can vary between different face detectors; thus, the estimation method needs to be calibrated for each face detector individually. Current face detection algorithms (see Section 2.1) can also determine the location of faces that are partly outside the image area (in this case, the coordinates of the bounding box output may be outside of the image area), state of the art methods can also work for images, where the head region is not fully covered by the image.

Alternatively, a software for head detection or segmentation could be used to determine the exact location and coverage of the head in the image. For instance, in [198], a GAN has been developed that performs segmentation of all parts of the head. However, these methods will not interpolate the position of head regions outside the image area and, thus, this approach will only work if the head is covered by the image.

# 14 Pose

## 14.1 Impact on Face Recognition

Variations in head pose are commonly described by three angles: the pitch, yaw, and roll angles which are the rotations about the vertical (y), the horizontal side-to-side (x), and the horizontal back to front (z) axes, respectively. Pitch, yaw, and roll angles of the head pose are defined in ISO/IEC 39794-5:2019. A frontal face image has 0° for all three angles. In case these angles strongly deviate from 0°, the visible facial parts change substantially, see Figure 35.

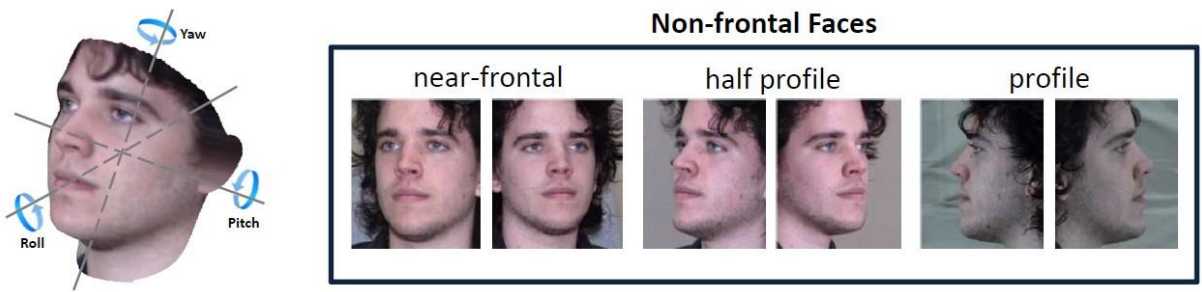

*Figure 35: Angles of the head pose and examples of non-frontal poses (yaw) (adapted from [201])*

It has been shown that non-frontal pose variations have large negative impact on the biometric performance of traditional face recognition systems utilizing hand-crafted feature extraction methods [199]. In contrast, state of the art face recognition systems are based on deep-learning techniques which achieve more robustness against pose variations [200]. Of the different possible pose variations, changes in the roll angle are expected to have the least impact on face recognition accuracy, since this rotation can be corrected during the pre-processing, i.e. face detection and alignment. However, it is difficult to quantize the actual impact of different pose relative to the frontal face recognition, since this will highly depend on the used face recognition algorithm and its robustness towards pose variations.

Numerous efforts have been devoted to developing pose-invariant face recognition methods [201]. The main goal of such techniques is to extract and compare facial features in a pose-invariant way. In deep learning-based approaches, this is usually achieved by including non-frontal face images in the training process. Moreover, pose-invariant face recognition methods may apply an additional processing step in which a transformation of a non-frontal face image to a frontal face image is performed. Methods for approximating frontal face images from non-frontal face images are usually referred to as *face frontalization* [202]. Since these approaches hallucinate facial parts which are not visible (based on the distribution of facial features learned during training), the approximated frontal face may not resemble the original face of a subject which hampers reliable recognition. Nevertheless, while current face recognition technologies are robust to slight variations in pose, stronger pose variations remain a challenge [201]. Here, a clear distinction between "slight" and "strong" depends on the robustness of the face recognition algorithm; it is expected that state of the art methods allow for reliable recognition if poses deviate slightly from a frontal pose. Nevertheless, in a frontal pose, the features of a subject's face are usually optimally visible (provided that there are no external occlusions). Therefore, exclusive use of facial images with frontal pose is expected to generally achieve the highest recognition performance of face recognition systems.

## 14.2 Datasets

Several datasets exist for benchmarking head pose estimation algorithms. Usually, these datasets contain ground truth annotations for the different rotation angles. The majority of available datasets contains images with pose variations in yaw and pitch angles, while some additionally include variations in the roll angle. It is important to note that some datasets which have been used in earlier works are not available anymore and therefore not mentioned here. Moreover, some datasets are claimed to be made available while no information can be found about them. A list of frequently used face datasets for head pose estimation is given in the following table. The listed databases contain pose ground truth values that are either measured directly through a specific camera setup of have been annotated by humans.

| Name | Subjects | Images | Remark | Constrained? | License |
|------|----------|--------|--------|--------------|---------|
| CMU Multi-PIE[232] | 337 | ~750,000 | 13 different yaw angles | Yes | Royalty-bearing |
| SynHead[233] | 10 | ~510,000 | Synthetic images | Yes | Royalty-free, non-commercial use |
| SASE[234] | 50 | 30,000 | Labels for angle values of head pose | Yes | No license |
| Pointing'04[235] | 15 | 2,940 | Labels partly correspond to viewing direction not pose | Yes | CC 4.0 |
| AFLW-2000[236] | 2,000 | 2,000 | Images from the AFLW data set with 3D models | No | Royalty-free |
| AFLW[58] | 25,993 | ~21,000 | Wide range of natural face poses, pose angles estimated from landmarks | No | Royalty-free, non-commercial use |
| BIWI[237] | 20 | ~15,000 | Labels for rotation angles of head pose | No | No license |
| 300W-LP[236] | unknown | ~61,000 | - | No | No license |
| CAS-PEAL-R1[128] | 1,040 | ~30,000 | 21 different poses labelled | yes | Non-commercial use |

---

[232] https://www.cs.cmu.edu/afs/cs/project/PIE/MultiPie/
[233] https://docs.google.com/forms/d/e/1FAIpQLSe6EP2FznQmlIf-QP8VJ638p9Da4wiZz2u-iTGtDgWV0qzXOA/viewform
[234] http://icv.tuit.ut.ee/datasets/
[235] https://figshare.com/articles/dataset/Pointing04_DB/5142466
[236] http://www.cbsr.ia.ac.cn/users/xiangyuzhu/projects/3DDFA/main.htm
[237] https://icu.ee.ethz.ch/research/datsets.html

| Name | Subjects | Images | Remark | Constrained? | License |
|------|----------|--------|--------|--------------|---------|
| FEI[238] | 200 | 2,800 | Yaw only | Yes | Non-commercial use |
| M2FPA[239] | 229 | ~400,000 | 62 poses (including 13 yaw angles, 5 pitch angles and 44 yaw-pitch angles) | Yes | Non-commercial use |
| Colour FERET[208] | 994 | 11,338 | Yaw only | Yes | Royalty-free |
| UMDFaces[11] | 8,277 | 367,888 | Currently unavailable | No | Released for use by the academic community |

*Table 32: Publicly available datasets with head pose annotations.*

The majority of the surveyed pose estimation methods have been evaluated on the AFLW-2000 (also referred to as AFLW2000-3D) and BIWI datasets, while many works use the 300W-LP dataset for training purposes. Example images are depicted in the following figure.

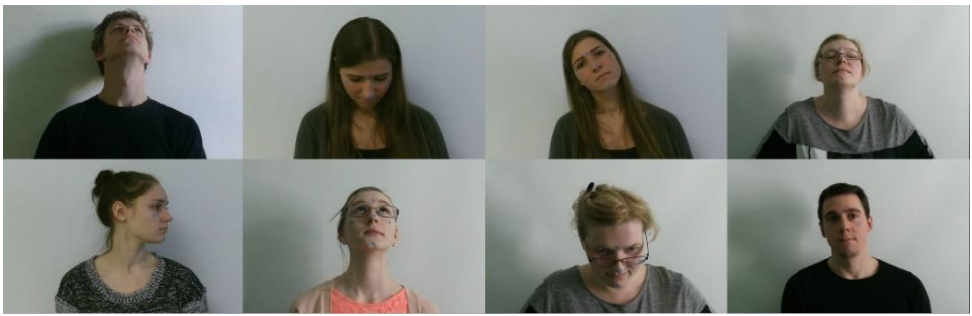

*Figure 36: Example images from the SASE dataset*

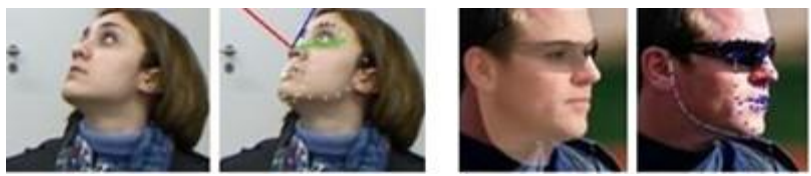

*Figure 37: Example images of the datasets BIWI (left) and AFLW-2000 (right)*

---

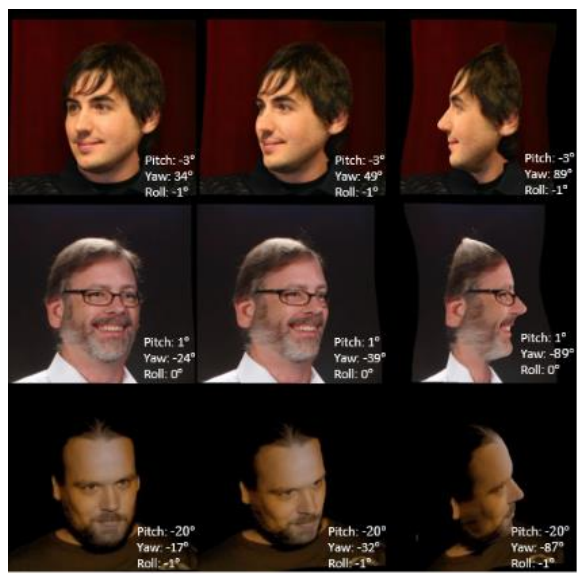

*Figure 39: Example images from the 300W dataset-LP*

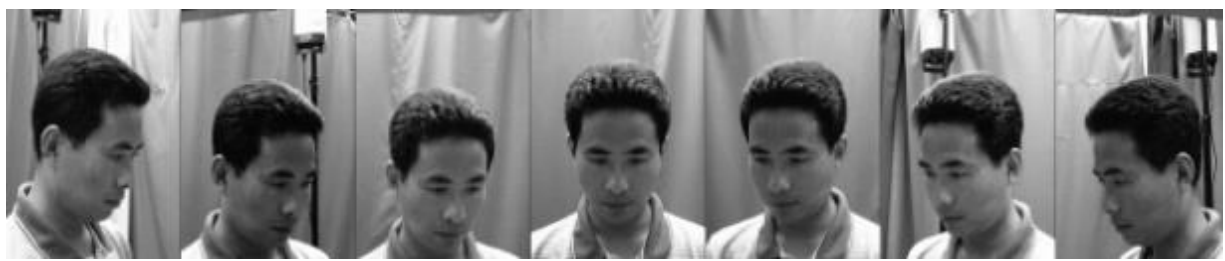

*Figure 40: Example for CAS-PEAL-R1 dataset*

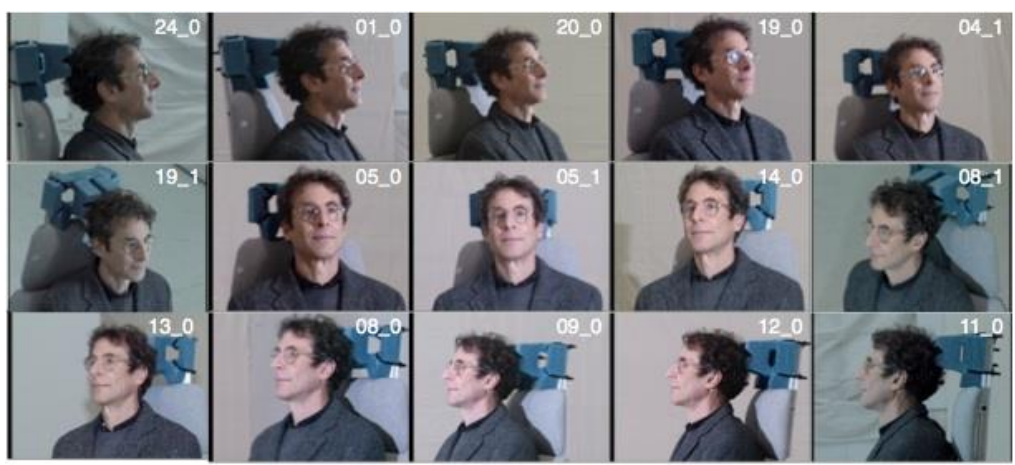

*Figure 38: Example images from the dataset CMU Multi-PIE*

## 14.3 Methods and Algorithms

Head pose estimation has numerous applications ranging from human-computer interaction to gaze prediction. A head pose estimation algorithm can operate at different levels of granularity. Coarse level head pose estimation maps a face image to a finite set of orientations, e.g. frontal versus non-frontal. In contrast, the majority of head pose estimation methods tries to calculate a granular pose in terms of pitch, yaw, and roll angles.

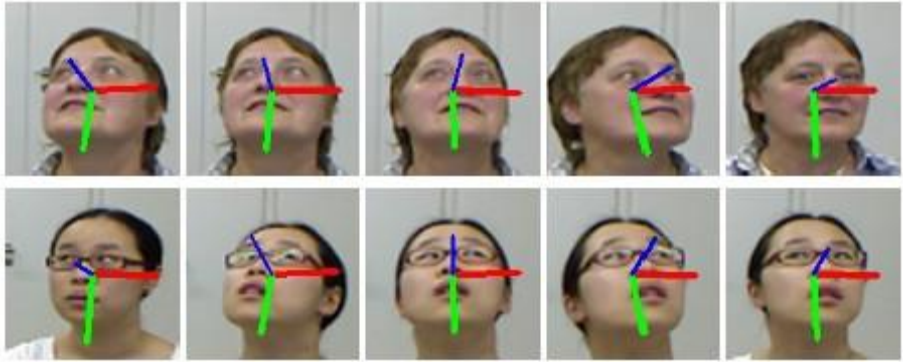

*Figure 41: Visualization of head pose angles by coloured lines (from [213])*

With respect to face image quality assessment, it is required to map an estimated facial pose, i.e. the angles yaw, pith and roll, to a single quality scalar in a predefined range, e.g. [0,100]. In ISO/IEC WD4 29794-5:2022, this mapping is not yet defined. Nevertheless, such mapping functions or look-up tables are necessary and could be defined for each rotation angle separately, since the different rotations are expected to have different impact on face sample quality.

Similar to other computer vision tasks, head pose estimation has been researched for a long period of time. A considerable amount of approaches has been proposed for video data. Such methods usually require a sequence of image frames and are, thus, out of scope in this project.

The most commonly used metric for evaluating the accuracy of face pose estimating methods is the *Mean Absolute Error (MAE)* with respect to ground truth pose data. The MAE is usually estimated per pose angle and the average MAE is often reported as a single performance measure which facilitates the comparison of different methods. The smaller the MAE values, the better the pose estimation algorithm and vice versa. For more quantized ground truth values, the *Pose Estimation Accuracy (PEA)* is sometimes reported. However, this metric is considered as less informative since it highly depends on the number of possible poses. In many scientific publications, the results of a performed head pose estimation on single images are commonly visualized through green, blue, and red lines which indicate the estimated yaw, pitch, and roll angles as shown in the figure below.

Like for other facial analysis tasks, early approaches were based on handcrafted texture features, i.e. texture descriptors, combined with machine learning-based classifiers. For instance, Jain and Crowley [203] suggested to apply Gaussian filters to a cropped facial image. Subsequently, an SVM was trained to do a multi-class classification. This means, for each rotation angle, fixed pose classes are defined, e.g. {-90°, -75°, ..., 90°}, to which a face image is classified. On the Pointing04 dataset, a MAE of 7-8% was achieved.

Another class of early approach derives pose from facial key points, i.e. landmarks. The eyes and the tip of the nose could be employed as such landmarks as, for instance, in [204]. These methods are simple and efficient since facial landmarks are also expected to be extracted for other quality metrics. Another type of approach is to solve the 2D to 3D correspondence problem based on estimated landmarks with a mean human head model, e.g. in [201]. Certainly, such methods strongly depend on the accuracy of the used landmark detectors, which remains challenging in some cases, e.g. if eyes are occluded by sunglasses. Therefore, these methods are considered as fragile and are commonly not considered as the current state of the art in head pose estimation. In contrast, Xin et al. [205], proposed a *Graph Convolutional Network* (GCN) to extract head pose from facial landmarks. To this end, a so-called landmark-connection graph is firstly determined from the facial landmarks. The method was shown to achieve state of the art-performance in terms of MAE. Specifically, MAEs of 4.64 and 3.92 were obtained on the AFLW-2000 and BIWI datasets, respectively. For landmark extraction, the authors suggest to employ the FAN method [206].

Like other computer vision tasks, pose estimation algorithms recently benefited from deep learning techniques. Moreover, accurate pose estimation can be a useful pre-processing step for other tasks, e.g. facial landmark detection or gaze estimation. In [207], Chang et al. propose to employ a simple CNN to estimate face pose as a 6-dimensional vector (3 rotation values and 3 translation values). This means that compared to the previous pose definition consisting of only 3 rotations, said vector also comprises the position of the face in the image. This method was introduced for the purpose of aligning face images as a pre-processing step for facial landmark detection. Therefore, the authors did not explicitly evaluate the accuracy of the pose estimation process, but of the landmark detection accuracy. It was shown that a preceding alignment of facial images based on estimated poses can improve the accuracy of a subsequently applied landmark detection algorithm. Similarly, Albiero et al. [208] suggested a similar concept for head pose estimation. They argue that extracting head pose vector like the one in [207] can be used as an alternative for conventional face bounding box detection. Hence, it is proposed to estimate head pose without face or facial landmark detection. This is achieved in a two-stage process: First, potential faces are located in the image using a *region proposal network* (RPN); subsequently, a CNN is trained to extract features from the corresponding *regions of interest* (ROIs) indicating whether these contain a face as well as the face pose vector. The application of this alignment method was shown to improve face recognition accuracy while experiments in terms of MAEs with respect to ground truth pose data were omitted.

Similar to [207], Fischer et al. [209] presented an eye gaze estimation scheme that uses coarse head pose estimation as pre-processing. It was shown that the robustness of eye gaze estimation can be greatly improved if a deep learning-based head pose estimation is incorporated into the processing pipeline.

Instead of classifying a facial image directly into a pose class, i.e. regressing all pose angles at once, Ruiz et al. [210] proposed a multi-loss network using three losses for yaw, pitch, and roll. By using three losses for the training (instead of a single loss), learning is shown to improve. This is realized by augmenting a shared backbone network with three fully-connected layers which predict the angles. On the AFLW-2000 and BIWI datasets, this approach achieves MAEs of 6.15 and 4.89. Similarly, Zhou and Gregson [211] introduced *WHENet* which uses backbone with classification and regression losses on pitch, yaw, and roll angles. Moreover, the authors introduced a so-called wrapped loss that significantly improves yaw estimation accuracy for anterior views. An MAE of 5.42 was reported for the AFLW-2000 dataset.

Ranjan et al. [212] introduced *HyperFace*, a multi-task network which is trained to simultaneously learn face and landmark detection, pose estimation, and gender prediction. The idea behind this approach is that information contained in the intermediate layers of a CNN is hierarchically distributed: Lower layers respond to edges and corners, and hence contain better localization properties. They are more suitable for learning landmarks localization and pose estimation tasks; on the other hand, deeper layers are class-specific and suitable for learning complex tasks such as face detection and gender classification. The intermediate layers of a deep CNN are fused using a separate CNN. Subsequently, a multi-task learning algorithm is applied to the fused features. For the task of pose estimation, an MAE of 4.62 is reported on the AFLW-2000 dataset. A similar method to learn meaningful intermediate features for pose estimation was also applied in *FSA-Net* [213]. The authors suggest to spatially group pixel-level features together into a set of features encoded with spatial information. Through such an aggregation, meaningful region-level features shall be learned. On the AFLW-2000 and the BIWI datasets, this method achieves MAEs of 5.07 and 4.00, respectively. This method is designed to have a compact model being achieved by building upon the so-called soft stage-wise regression scheme resulting in a model size of about 3 MBs.

Another multi-task network approach was presented by Wu et al. [45]. Their *SynergyNet* aims at simultaneously predicting various features of facial geometry including head pose. They suggest to use a backbone network to extract features useful for reconstructing 3D face models from 2D images. While the overall goal of this method is to refine facial landmark detection by analysing said facial geometry features, the estimated pose angles reveal a competitive MAE of 3.35 on the AFLW-2000 dataset. Liu et al. [214] also propose a multi-task network *MOS* that performs face and landmark detection, as well as head pose estimation. For the task of head pose estimation, an MAE of 4.43 is achieved on the AFLW-2000 dataset. Their lightweight architecture results in a model size below 15 MBs.

In [42], Wang et al. combine a backbone network with a fully connected layer consisting of up to 198 nodes representing different yaw/pitch/roll angle combinations of which the total loss is estimated during training. This rather simple approach is reported to achieve an MAE of 5.09 on the AFLW-2000 dataset and 3.02 on the BIWI dataset.

A simple method to penalize non-frontal poses in the computation of a unified quality score has been proposed by Babnik et al. [70]: they use a face recognition CNN and compute the difference between the embeddings of the image and the horizontally flipped image. The idea behind this approach is that faces are typically horizontally symmetric. This approach could also be used to obtain a general measure for the frontality of the pose.

## 14.4 Software

Numerous open-source implementations of facial pose estimation algorithms are available.

| Name | Model Size | Accuracy (MAE) | | License |
|------|------------|----------|------|---------|
| | | **AFLW2000** | **BIWI** | |
| FacePoseNet[240] [207] | 512 MB | n. a.[241] | | No license |

---

| Name | Model Size | Accuracy (MAE) | | License |
|------|------------|-----------|------|---------|
| | | **AFLW2000** | **BIWI** | |
| RT-GENE[242] [209] | 350 MB | n. a.[243] | | CC BY-NC-SA 4.0 |
| Hopenet[244] [210] | 92 MB | 6.15 | 4.89 | Apache 2.0 |
| HyperFace[245] [212] | 120 MB | 4.26 | n. a. | MIT |
| FSA-Net[246] [213] | 3 MB | 5.07 | 4.00 | Apache 2.0 |
| accurate-head-pose[247] [42] | 97 MB | 5.09 | 3.02 | No license |
| 3DDFA_V2[75] [46] | 13 MB | n. a.[241] | | MIT |
| | 4 MB | | | |
| Dense-Head-Pose-Estimation[76] [46] | 13 MB | n. a. | n. a. | MIT |
| WHENet[248] [211] | 17 MB | 5.42 | 3.81 | BSD-3-Clause |
| SynergyNet[249] [45] | 72 MB | 3.35 | n. a. | MIT |
| MOS[250] [214] | 13 MB | 4.43 | n. a. | MIT |
| img2pose[251] [215] | 150 MB | 3.91 | 3.79 | Attribution- non-commercial use 4.0 International |
| FacePose_pytorch[29] | _[252] | n.a. | n.a. | MIT |

*Table 33: Publicly available open source implementations of state of the art head pose estimation algorithms.*

---

[242] https://github.com/Tobias-Fischer/rt_gene
[243] Targeted application: eye gaze estimation
[244] https://github.com/natanielruiz/deep-head-pose
[245] https://github.com/takiyu/hyperface
[246] https://github.com/XiaoJiNu/FSA-Net
[247] https://github.com/haofanwang/accurate-head-pose
[248] https://github.com/Ascend-Research/HeadPoseEstimation-WHENet
[249] https://github.com/choyingw/SynergyNet
[250] https://github.com/lyp-deeplearning/MOS-Multi-Task-Face-Detect
[251] https://github.com/vitoralbiero/img2pose
[252] The implementation uses a linear regression from 3 landmarks implemented in the source code

# 15 Expression Neutrality

## 15.1 Impact on Face Recognition

The effects of facial expression on the biometric recognition performance of a face recognition system are well documented in the scientific literature. While state of the art face recognition systems are robust against slight variations in facial expression, more extreme variations between comparison images of a subject can lead to false rejection. This has been showcased in different works, e.g. in [216] [217]. In [216], face images with neutral expression were compared against face images with different types of expressions using different open-source face recognition systems and different public datasets. Obtained score distributions revealed that face images with certain types of facial expressions, e.g. sadness or disgust, may significantly impact the resulting comparison scores. Other types of expression, e.g. Joy or Surprise, were reported to have less negative impact of comparison scores. In [217], it was demonstrated that extreme facial expressions (grimaces) can be performed to prevent from being successfully recognized by state of the art face recognition systems.

With a neutral facial expression, features of a subject's face are usually clearly visible, while they can be partially occluded (e.g. by wrinkles or squinted eyes) or distorted for other expressions. Furthermore, a neutral expression ensures a lower variance of the features as compared to other expressions that allow more variety. Therefore, exclusive use of facial images with neutral facial expressions is expected to improve the recognition performance of face recognition systems. This has, for instance, been empirically confirmed in [216].

## 15.2 Datasets

Different datasets have been collected and made available for benchmarking the accuracy of facial expression recognition algorithms. These datasets include face images with expression labels. Commonly, the aforementioned six basic expressions are used together with the label "neutral". Datasets may also include subject labels which enables to perform mated comparison trial. As mentioned above, this becomes relevant when the effect of different facial expressions on biometric performance needs to be estimated. Lately, various unconstrained (in-the-wild) datasets with facial expression labels have been released. However, these are out of scope for this project and therefore not discussed. The following table lists most relevant datasets which are frequently used to benchmark expression recognition algorithms. The majority of available datasets are free of costs with a few exceptions, e.g. CMU Multi-PIE.

| Name | Subjects | Images | Remark | Constrained? | License |
|------|----------|--------|--------|--------------|---------|
| CMU Multi-PIE[232] | 337 | 750,000 | 6 different expressions labelled | Yes | Royalty-bearing |
| CK+[253] | 123 | 593 (sequences) | 8 different expressions labelled | Yes | No license |
| CFD[254] | 827 | 1441 | 6 different expressions labelled Non-neutral expressions available only for 158 subjects (610 images) | Yes | Non-commercial use |
| MMI[255] | 25 | 740 | Several facial expressions | Yes | Academic use only |
| FEAFA+[256] | 154 | ~230,000 | Manually (between 0 and 1) labelled expressions | partly | Academic use only[4] |
| Oulo-CASIA[257] | 80 | 2,880 | 6 facial expressions labelled | Yes | No license |

---

[253] https://github.com/spenceryee/CS229/tree/master/CK%2B
[254] https://www.chicagofaces.org/
[255] https://mmifacedb.eu/
[256] https://www.iiplab.net/feafa+/
[257] https://www.oulu.fi/en/university/faculties-and-units/faculty-information-technology-and-electrical-engineering/centre-machine-vision-and-signal-analysis

| Name | Subjects | Images | Remark | Constrained? | License |
|------|----------|--------|--------|--------------|---------|
| MUG[258] | 86 | ~1,400 | 6 facial expressions labelled | Yes | Non-commercial use |
| BU-3DFE[57] | 100 | 2,500 | Includes 3D data, six facial expressions labelled | Yes | Non-profit use |
| FER-2013[259] | unknown | ~35,000 | six facial expressions labelled | No | No license |
| RAF-DB[260] | unknown | ~30,000 | seven facial expressions labelled | No | Non-commercial use |
| AffectNet[261] | unknown | ~440,000 | eight facial expressions manually labelled | No | Non-commercial use |
| Aff-Wild[262] | unknown | unknown | uses the EmotiW dataset - partly labels for basic expressions | No | Academic and commercial use Royalty-bearing |
| Aff-Wild2[263] | 458 | ~2,800,000 | extended AFF-Wild partly labels for basic expressions | No | Academic and commercial use Royalty-bearing |
| EmotiW[264] | unknown | unknown | Seven emotions labelled | No | Academic use only |

*Table 34: Datasets used for facial expression recognition algorithm*

---

[258] https://mug.ee.auth.gr/fed/
[259] https://www.kaggle.com/c/challenges-in-representation-learning-facial-expression-recognition-challenge
[260] http://www.whdeng.cn/raf/model1.html
[261] http://mohammadmahoor.com/affectnet/
[262] https://ibug.doc.ic.ac.uk/resources/first-affect-wild-challenge/
[263] https://ibug.doc.ic.ac.uk/resources/aff-wild2/
[264] https://sites.google.com/view/emotiw2020/challenge-details

                                                                            Federal Office for Information Security

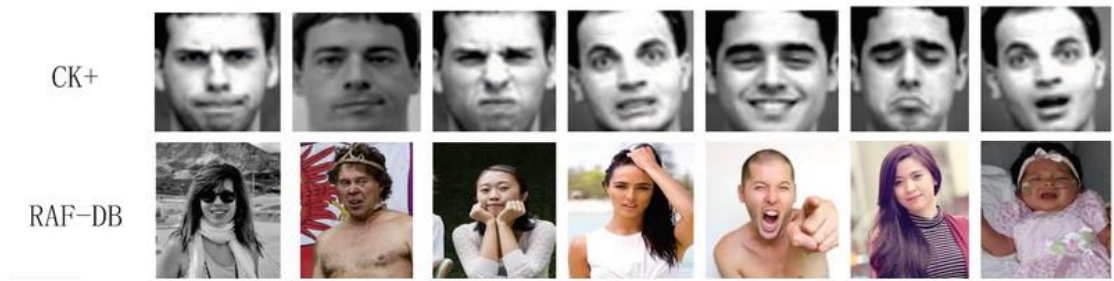

*Figure 43: Comparison of constrained and unconstrained images from the datasets CK+ and RAF-DB*

Figure 43 shows a comparison between a popular constrained and unconstrained dataset for facial expression recognition. As indicated above, w. r. t. the quality metric face neutrality, expression estimation can also be seen as a two-class problem. That is, any face dataset for which the ground truth labels "neutral" and "non-neutral" are available could be employed to benchmark expression estimation algorithms.

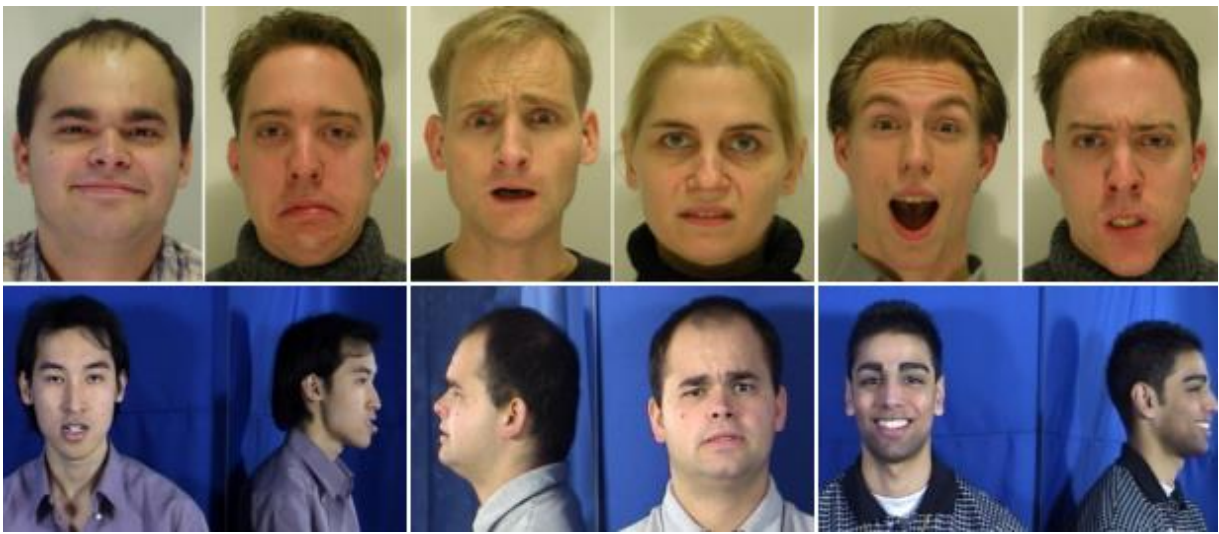

*Figure 42: Example images from dataset MMI*

## 15.3   Methods and Algorithms

Since any deviation from the neutral expression can negatively impact the recognition performance [216], it is of interest to estimate how neutral a facial expression is. However, in the literature, expression recognition is usually seen as a multi-class problem [218]. Specific classes of facial expressions have been established. The most important classes are

- Joy,
- Sadness,
- Anger,
- Fear,
- Surprise and
- Disgust.

The above expressions are commonly referred to as the six basic expression [218]. Additionally, "neutral" is usually taken into account. It should be noted that across different works, different classes may be considered. This also depends on the application scenario. For example, for some applications the detection of fear might be more relevant than surprise. Moreover, additional expressions such as "fatigue" may be considered. It is important note, that different classes of facial expressions deviate from a neutral expression to a different extent. For instance, a moderate degree of sadness usually only slightly differs from a neutral expression while joy usually changes the corners of the mouth more drastically. In ISO/IEC 39794-5:2019, various aspects of expressions are defined as Boolean in the data block "Expression block", e.g. "eyes looking away from the camera" or "raised eyebrows".

Basically, facial expression classification approaches work as follows: (1) a face is detected and normalized (pre-processing), then (2) features are extracted based on which the facial expression is classified. Finally, (3) confidence values are calculated for the respective classes (vector of values) with subsequent classification. The classification can be implemented directly by determining the maximum confidence score across expression classes, see Figure 44.

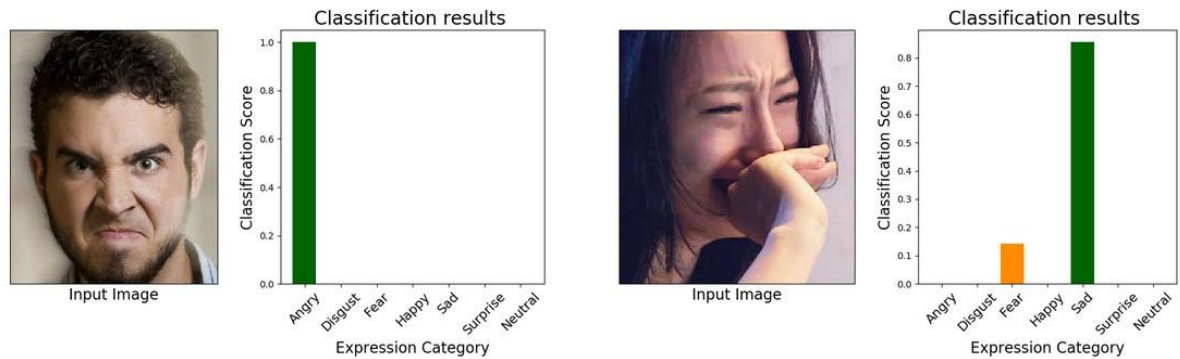

*Figure 44: Example of expression classification results (adapted from https://github.com/WuJie1010/Facial-Expression-Recognition.Pytorch)*

Early works used hand-crafted methods, e.g. texture descriptors, for the purpose of feature extraction [218]. Further, expressions may be derived from facial landmarks if a frontal pose can be assumed [219]. However, hand-crafted approaches are not state of the art anymore. Analogous to face recognition, the accuracy of methods for classifying facial expressions have been drastically improved by using deep convolutional neural networks (DCNN). Nowadays, the current state of the art regarding facial expression classification is based exclusively on DCNN [220]. For the final classification step, two approaches are prevalent in state of the art algorithms:

1. The outputs of the DCNN are fed to a classical machine learning algorithm, e.g. support vector machines. In this case, the DCNN only serves as a feature extractor.
2. The DCNN is trained to perform the classification resulting in an end-to-end learning. In such an approach, the final layer of the DCNN contains a node for each expression class. Then, the prediction probability of each sample can be directly output by the network based on the activation of the according nodes using a suitable loss function, e.g. soft-max.

Identifying the most relevant existing works with respect to this project is non-trivial due to several reasons.

- Facial expression estimation is a very active and dynamic research field. There exists a huge amount of approaches published in the scientific literature. For instance, a recent survey focusing on deep learning-based expression estimation [220] discusses a very large number of relevant approaches in a time-span of approximately ten years. Nevertheless, the survey is already partially outdated as many new interesting approaches have been published since. Likewise, this present document can only be considered as a snapshot of the state of the art.
- Besides, numerous additional methods have been published in less prominent scientific conferences and journals by research groups that are not well-known in the field of pattern recognition. Such works sometimes state auspicious performance rates outperforming state of the art approaches without providing a proper justification or any implementation, e.g. [221]. Usually, said kind of works are rarely cited by other scientific works, limiting its creditability.
- Another issue is that publications report classification accuracy with respect to different datasets that are not directly comparable. Earlier released datasets are usually small in size and captured in a laboratory-controlled scenario, e.g. the frequently used CK+ dataset [222] (released in 2010) contains face images with various expressions of 123 subjects. In contrast, newer large-scale datasets usually contain more challenging unconstrained (in-the-wild) face images. In the context of this project, face images with extreme variations in terms of facial expression (as well as other factors) are of less relevance. Such types of images are mostly contained in in-the-wild datasets. Obviously, a comparison between methods that have been evaluated on different datasets is difficult and often misleading. It is important to note that newer methods are usually exclusively evaluated on challenging in-the-wild datasets, e.g. RAF-DB [223] (released in 2017). Nevertheless, based on similar observations made for face recognition, it is reasonable to assume that methods which achieve competitive results on large in-the-wild datasets will also achieve good accuracy on more constrained datasets.

- In addition, as previously mentioned, the vast majority of works aims at assigning face images to one of the aforementioned expression classes. In order to directly compare different methods on a single dataset, the average classification accuracy across all classes is usually reported as a single performance rate. However, facial image quality focuses on expression neutrality, and thus, the proposed algorithms for multi-class expression estimation must be transferred to a two-class (neutral, not neutral) classification algorithm. This means, certain kind of classification errors which impact the average classification accuracy, e.g. a misclassification of the expression "sad" into class "angry", might be less relevant for facial image quality.

In order to use existing methods for facial expressions classification to determine the degree of neutrality of the expression, the confidence score for the class "neutral" can be used. Such confidence scores for the individual classes are available in most methods [220]. From this, a scalar value in a predefined range, e.g. [0,100], can be derived directly by quantization and normalization, e.g. by employing methods summarized in [220]. On the other hand, some expressions, e.g. moderate sadness, deviate less from the neutral one than others. Thus, it can be advantageous to define a mapping of the vector of confidence values for all expression classes to a scalar value in a predefined range in which the individual facial expressions classes are weighted according to their effect on facial recognition performance. In order define such a mapping (either hand-crafted or by machine learning), expression ground truth data with mated samples need to be available. On the other hand, the individual confidence values of a multi-class facial expression estimation algorithm may provide actionable feedback to the user, e.g. "please smile less" or "do not frown".

Expression estimation methods can further be categorized based on the required input data. So-called static approaches perform an expression estimation based on a single input image while dynamic approaches consider the temporal relation among image frames in an input video. The latter type of methods usually achieves the most competitive performance results. However, in the context of this project, dynamic approaches are only relevant if these can also be applied to a single image (and still achieve state of the art accuracy). Note that dynamic approaches are commonly tested on laboratory-controlled datasets containing sequences of facial expressions which is usually not the case for in-the-wild-datasets.

One of the first works focusing on expression estimation in-the-wild was done by Li et al. [223]. They were demonstrating that handcrafted approaches, e.g. LBP and Gabor filters combined with SVM or LDA for classification, can achieve good results on constrained datasets, e.g. above 90% accuracy on the CK+ dataset. However, on unconstrained datasets which contain more variations, e.g. in terms of pose, the performance of such methods significantly drops, i.e. to approximately 50% accuracy on the unconstrained RAF-DB. Therefore, Li et al. [223] trained a neural network (from scratch) to learn facial representations that are suitable for expression estimation. As expected, this deep learning-based approach significantly outperforms handcrafted methods, achieving an accuracy of 74% on the RAF-DB and approximately 95% accuracy on the CK+ dataset. In addition, a comparison against networks pre-trained on other tasks, e.g. image classification or face recognition, was conducted. Features extracted by these networks have been classified using basic machine learning techniques, e.g. LDA or SVM. It was observed that models pre-trained on face recognition, e.g. VGG-Face, are also suitable for expression estimation in contrast to general models, e.g. ImageNet, which tend to obtain inferior performance. In detail, on the RAF-DB the dedicated DCNN obtained an average accuracy of 74% while the use of the pre-trained VGG-Face and the AlexNet networks obtained 58% and 55%, respectively. Zhang et al. [224] presented an interesting approach based on a multitask network that is capable of learning from rich auxiliary attributes such as gender, age, and head pose, beyond just facial expression data (unlike existing models that typically learn by utilizing facial expression labels alone). This turned out to be very beneficial resulting in improved classification performance, i.e. 98.9% on the CK+ dataset. Liu et al. [225] proposed an identity-disentangled facial expression recognition method. The idea of this work is to untangle the identity from the face image prior to the expression estimation. An accuracy of 97.76% is reported for the CK+ dataset.

Motivated by the fact that unconstrained datasets usually contain a large amount of partially occluded faces, e.g. due to sun glasses or medical masks, some research efforts have been made towards expression estimation from partially occluded faces. Li et al. [226] suggest a patch-based classification based on a neural network with attention mechanism which enhances some parts of the input data while diminishing other parts. Their methods achieved good results on different unconstrained datasets as well as an accuracy of 97% on the constrained CK+ dataset. Similarly, Wang et al. [227] proposed a Region Attention Network (RAN), to adaptively capture the importance of facial regions for occlusion and pose invariant facial expression estimation. The RAN aggregates and embeds a variable number of region-based features extracted by a pre-trained neural network, e.g. ResNet-18 or VGG16, into a compact fixed-length representation. Additionally, a region-biased loss is proposed to encourage high attention weights for the most important regions. On the RAF-DB, this approach achieves an average accuracy of 86.9%. Similarly, Farzaneh and Qi [228] presented a network incorporating an attention mechanism to estimate attention weights correlated with feature importance for expression recognition. An accuracy of 87.78% on RAF-DB was reported by the authors.

While different neural network architectures may obtain further slight improvements in terms of classification accuracy, different research groups recently started to put focus on the so-called *ambiguity problem*: deep learning-based techniques require a huge amount of labelled training data, i.e. faces with expression class labels. However, these labels are affected by uncertainties caused by ambiguous facial expressions, low-quality facial images, and the subjectiveness of annotators. An example of this problem is shown in Figure 45.

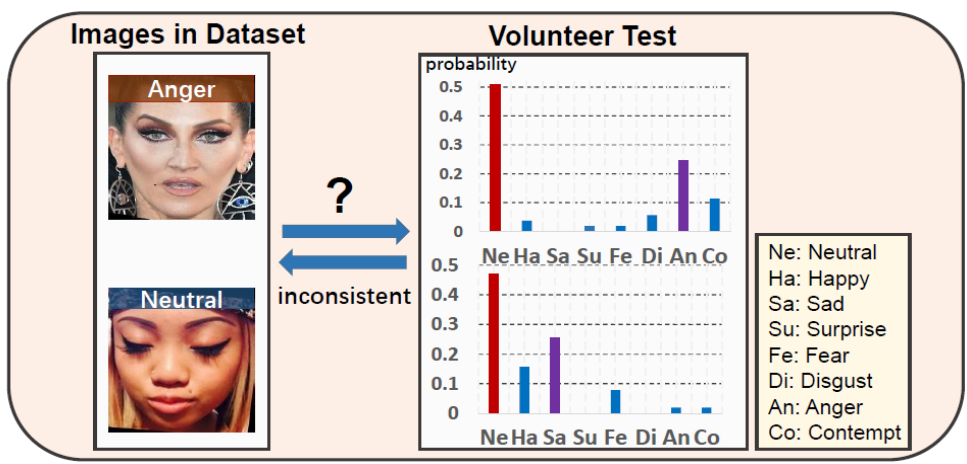

*Figure 45: Example of an ambiguity problem towards facial expression (taken from [233])*

The labels of the two depicted face images are "Anger" and "Neutral". A number of 50 volunteers labelled both images. The majority of these volunteers labelled both mages as "neutral". Such uncertainties may result in over-fitting on the uncertain samples which may be mislabelled. Further, it is difficult for a model to learn useful facial expressions, if a high ratio of labels is incorrect and it may even result in the model's disconvergence in training. Due to this issue, several works have been proposed which leverage label distributions, i.e. a distribution of probabilities (see above figure), as labels during algorithm training instead of one-hot ground truth labels.

Wang et al. [229] propose a so-called Self-Cure Network (SCN), to suppress the uncertainties for large-scale facial expression recognition. In this method, noisy labels are automatically re-labelled during training (updated labels of the used datasets are, however, not made available). This simple yet efficient method was shown to achieve an accuracy of 88.14% on RAF-DB. This demonstrated that the ambiguity problem may have significant impact on facial expression estimation algorithms. A similar approach was presented in [230], reporting an accuracy of 86.6% on RAF-DB and 99.69% on CK+, respectively. Alternatively, to suppressing uncertain labels, Label Distribution Learning (LDL) can be employed, which allows multiple labels with different intensity to be linked to one expression. Chen et al. [231] proposed to apply LDL to expression estimation. However, LDL cannot be directly applied to most existing datasets because these only contain one-hot labels rather than label distributions. To solve the problem, the authors suggest to leverage the topological information of the labels from related tasks such as facial landmark detection, e.g. to use the facial landmark positions as additional features for their facial expression recognition method. On the RAF-DB, their method achieved an accuracy of 85% while it only achieved 91% on the less constrained CK+ dataset.

Zhao et al. [232] create ground truth label distributions by employing a pre-trained network which extracts confidence values for each expression class. Afterwards, another network which performs a patch-wise feature extraction is trained with LDL. The authors report a performance of 88.36% on the RAF-DB. More recently, She et al. [233] proposed a similar approach by which the performance on the RAF-DB was improved to almost 90% accuracy. Vemulapalli and Agarwala [234] argue that the ambiguity problem remains an obstacle that hinders reliable face expression estimation. In particular, the assignment of a facial image to an expression category may be inconsistent among different people. In contrast to the aforementioned approaches, they propose a neural network to extract a 16-dimensional expression embeddings from face images which allows a distance measure between emotions. Unfortunately, they did not evaluate their method with commonly used metrics, which prevents a comparison with previous approaches. Zhang et al. [235] propose a so-called relative uncertainty learning process. In this scheme, image pairs are automatically processed with the goal of assigning small uncertainty values to easy images and large uncertainty values to uncertain images. During the learning process a so-called feature mix-up step is assigning larger weights to uncertain images where the loss is estimated pair-wise. The overall aim of this method is to encourage the model to learn reliable uncertainty values. An accuracy of 88.98% on the RAF-DB is reported.

Shi et al. [236] raised the issue that the usual zero-padding of convolutional layers causes disturbances of the calculated features in that the contribution of the padded zero-values (resulting in reduced ("albino") feature values) spread layer by layer from the outside to the inside. To counter this, they employ two techniques: a) Feature arrangement, in which the output arrays of the individual filters (feature channels) of a layer are combined in a single large array. Since the fraction of edge values decreases linearly with the size of the array, by this re-arrangement, the influence of the padding is reduced. b) A so-called De-Albino approach, in which convolutional layers are applied without padding. Since such unpadded filters have a perception bias towards the inner values, the spread of the Albino effect, the spread of this effect is decreased. Besides, the authors apply a regulation technique similar to batch normalization: When training, they calculate the features not only per image but also averaged over the batch. These average values are added as generic features to the features of the individual image. This allows the network to better learn features that are typical for certain expressions.

Massoli et al. [237] highlight the problem that the accuracy of facial expression recognition algorithm is impacted by image resolution, i.e. performance is expected to decrease with image resolution. To tackle this issue, the authors propose a multi-resolution training procedure in which a random down-sampling of face images is performed. An existing face recognition model (VGGFace2) is then re-trained for facial expression recognition using said training data resulting in an accuracy of 88.07% on the RAF-DB dataset.

## 15.4 Software

On GitHub, over 280 entries could be found for the topic of facial expression estimation.[265] The most promising implementations with the relevant implementations are listed in the table below. For these implementations, competitive classification performance on the public benchmark datasets REF-DB and AffectNet are reported.

| Name | Model size | License | Accuracy | |
|---|---|---|---|---|
| | | | RAF-DB | AffectNet |
| Self-Cure Network[266] [229] | 257 MB | No license | 88.14% | 60.23% |
| BEP[267] | 588 MB | No license | 87.8% | 64.06% |
| FacePose_pytorch[29] | 128 MB | No license | 95% [268] | 80% [268] |
| EfficientFace[269] [232] | 5 MB | MIT | 88.36% | 59.89% |
| DMUE[270] [233] | 43 MB[3] | Apache 2.0 | 89.42% | 63.11% |
| DAN[271] [238] | 75 MB[3] | MIT | 89.7% | 62.09% |
| MAFER[272] [237] | 271 MB | No license | 88.07% | n. a. |

*Table 35: Relevant implementations of facial expression estimation algorithms*

By far, the best accuracy is reported for the implementation of FacePose_pytorch[29], which is based on ResNet18 trained with an undocumented method (no training code is published). However, note that it is not clear, if this implementation is completely available: In this repository, the author states that they plan to release their "ultra-high precision model" in the future or on request by email; such a request had already been sent 6 months ago by secunet but remained unanswered since then, and the author has been inactive on GitHub for the last 12 months. Thus, it is not clear, if the model provided in the link in the repository is indeed that one for which the claimed accuracy is achieved.

---

[265] https://github.com/topics/facial-expression-recognition
[266] https://github.com/kaiwang960112/Self-Cure-Network
[267] https://github.com/bmanczak/BEP
[268] Unclear, if these accuracy numbers refer to the published model, see text below table.
[269] https://github.com/zengqunzhao/EfficientFace
[270] https://github.com/JDAI-CV/FaceX-Zoo/tree/main/addition_module/DMUE
[271] https://github.com/yaoing/DAN
[272] https://github.com/fvmassoli/mafer-multires-facial-expression-recognition

# 16 Mouth Closed

This section summarizes the quality aspect mouth closed. According to ISO/IEC 39794-5:2019 (Section D.1.4.3.2), the captured facial subject is required to have a neutral facial expression, which requires the subject not to smile, having a closed mouth, and shows no teeth. As per ISO/IEC WD4 29794-5:2022, expression neutrality is a mandatory image quality aspect, when it comes to capturing reference samples for ID documents.

Figure 46 visualizes compliant and non-compliant (by means of ISO/IEC 39794-5:2019, Section D.1.4.3.2) facial images in view of the open mouth and visible teeth property. Here it should be noted that even a smile in combination with a closed mouth is not compliant. More information about expression neutrality can be found in Section 15.

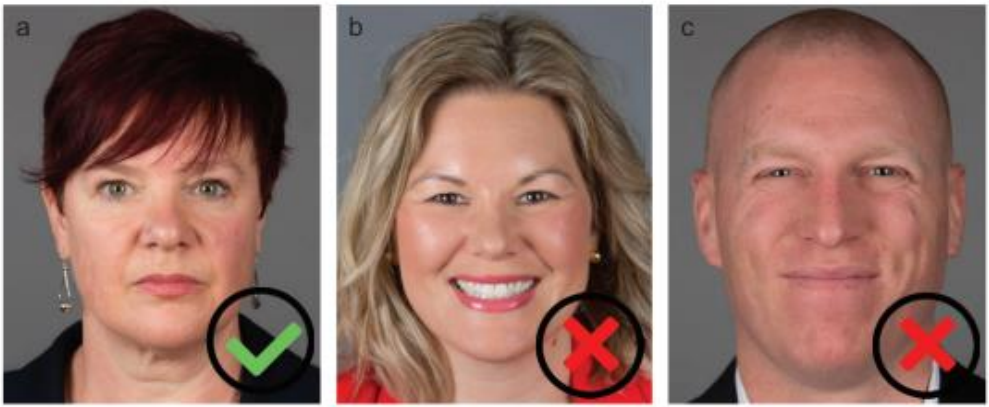

*Figure 46: Example of compliant and non-compliant facial expressions considering the mouth region (taken from ISO/IEC 39794-5:2019)*

## 16.1 Impact on Face Recognition

According to [239] there are differences in face recognition performance based on the facial mouth shape/expression. For example, it has been shown, that an opened mouth has the potential to significantly affect the recognition accuracy, whereas minor variations of the corners of the mouth or a slight smile do not have a substantial effect. However, in this analysis, only simple face recognition algorithms based on PCA and LDA were used, which are not state of the art. Given the variance of mouth openness in the in-the-wild datasets used to train state of the art face recognition CNNs, it seems plausible to assume that these CNNS are much more robust with respect to the mouth open/close aspect. Nevertheless, other biometric algorithms, e.g. PAD or MAD, and manual identity verification may benefit from a clothed mouth.

## 16.2 Datasets

Searches for datasets focusing on closed and open mouths result in datasets in the context of yawn detection. Yawn detection is often used in the context of sleepiness detection for car drivers, but can also be used as a dataset for open and closed mouths. The following table lists datasets found during investigation for writing the present report:

| Name | Subjects | Images | Remark | License |
|------|----------|--------|--------|---------|
| Yawning Detection[273] | unknown | 8186 | | No license |
| Drowsiness_dataset [274] | unknown | 2900 | Eyes open/closed, Yawn/No Yawn | No license |
| EURECOM Kinect Face Dataset[275] | 52 | 936 | open mouth/occluded mouth | Royalty-free license |
| The Chicago Face Database[276] | 597 | n. a. | Neutral expression, happy (open mouth), happy (closed mouth), angry, fearful | Non-commercial |
| CelebA[277] | 10.000 | 200.000 | Attribute "Mouth Open" computed with CNN, see [240] | Non-commercial |
| CelebA-HQ[35] | n.a. | 30.000 | Subset of CelebA with high-resolution images | Non-commercial |

*Table 36: Summary of potential datasets for assessing the mouth closed aspect.*

## 16.3   Methods and Algorithms

It seems impossible to make a statement about the mouth (open or closed) when it is covered, for example by wearing a mask in the current COVID-19 pandemic situation. The following methods assume that the mouth is clearly visible on the captured image.

According to ISO/IEC WD4 29794-5:2022 a method for measuring if a mouth in a facial image is open/closed can be summarized as follow: First the maximum distance between the borders of the upper and lower lip is computed. Then the inter-eye distance (see Section 12) is computed, which is then needed for the mouth openness aspect; the mouth openness aspect is computed as the ratio between the lip-distance and the eye-distance.

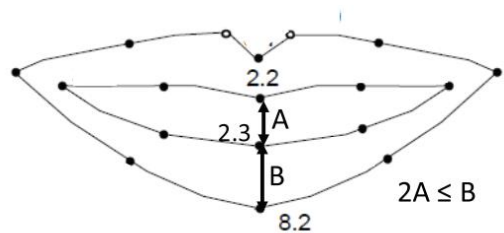

*Figure 47: Definition of a closed mouth (taken from ISO_IEC_39794-5:2019)*

According to ISO/IEC 39794-5:2019, the mouth in a given facial image is considered to be closed if the distance A between the inner borders of the lips is less than 50 % of the thickness of the lower lip B (see Figure 47). For this calculation the facial landmarks of the mouth need to be computed.

---

Some approaches for drowsiness/fatigue detection (see Section 17.3) also aim to detect yawning. While yawning (typically) implies that the mouth is open, such methods will likely not detect cases, where the mouth is only slightly open. Thus, we do not consider such approaches here.

Other implementations for the assessment/detection of an open/closed mouth will be listed in the next section.

## 16.4   Software

Generally, an open mouth can be detected using landmark detection (see Section 2.3), provided that an appropriate model detects landmarks in the mouth area. However, there are other approaches. For example, in facial expression recognition, it is often the case that the class "surprised" provides an open mouth (see Section 15). The following table lists the most relevant free software implementations for open/closed mouth detection, all of which are based on landmark detection or neural networks:

| Name | Model Size | Approach | License |
|------|-----------|----------|---------|
| Mouth-open[278] | n. a. | Landmarks | MIT |
| Mouth-open-js[279] | 9 MB | CNN | Apache 2.0 |
| HippoYD[280] | 30 MB | CNN | Apache 2.0 |

*Table 37: Examples of software available for open/closed mouth detection.*

---

[278] https://github.com/mauckc/mouth-open
[279] https://github.com/iglaweb/mouth-open-js
[280] https://github.com/iglaweb/HippoYD

# 17 Eyes Open

The presence of two open eyes is considered as an essential condition for a good face quality. ISO/IEC 39794-5:2019 (Section D.1.4.3.3) states that both eyes should be opened naturally, but not forced wide-opened. Figure 48 depicts examples of acceptable and unacceptable face images regarding open eyes.

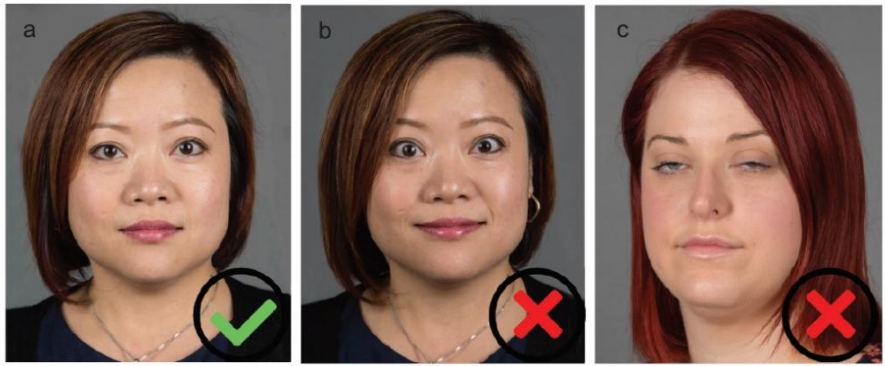

**a) Compliant portrait**  **b) Unnaturally wide**  **c) Eyes not fully opened**
**opened eyes**

*Figure 48: Examples of acceptable open eyes (a), forced wide-opened (b) and not (fully) opened (c) (taken from ISO/IEC 39794-5:2019)*

The pupils and irises should be completely visible. However, it is important to consider that due to the ethnicity, ageing or other reasons, the size and the shape of eyes may vary. For example, people of Caucasian descent often have larger eyes as compared to people of East Asian descent. Thus, there might be exceptions regarding the complete visibility of the eyes.

## 17.1 Impact on Face Recognition

State of the art face detectors have virtually no problems with detecting faces, even under difficult conditions (see Section 2.1). Furthermore, modern landmark extractors (see Section 2.3) can successfully detect the eyes and localize their boundary even if they are closed. Thereby, considering current SOTA face detectors and landmark extractors, it is rather irrelevant, if eyes are closed or open.

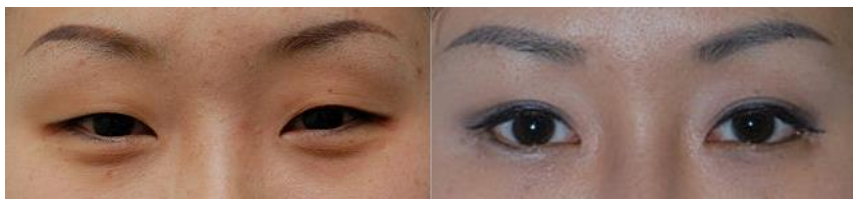

*Figure 49: A woman, before (left) and after (right) blepharoplasty (taken from Wikipedia)*

However, the shape of the eyes and the textures of eyelids and eyelashes are discriminative features. For instance, the so-called "double eyelid surgery" (a.k.a. blepharoplasty) can significantly change a person's appearance (see Figure 49). Thus, it is plausible to assume that these features can considerably influence face recognition. However, when eyes are fully closed, parts of these features can be occluded or modified. On the other hand, state of the art CNN-based face recognition implementations take images scaled to a comparably small dimension (typically 112x112 pixels) as input [65]. In small dimensioned facial images, the textures of the irises can hardly be seen and can hardly be distinguished from lighting artefacts (reflections). On the other hand, no research that investigate the impact of closed eyes on face recognition was found.

## 17.2 Datasets

The following table lists datasets that have a focus on eyes state and that can be used for blink detection or sleepiness/drowsiness detection.

| Name | Subjects | Images | Remark | License |
|---|---|---|---|---|
| Eyeblink8[281] | 4 | 8 videos | 408 eye blinks on 70 992 annotated frames | GPL3 License |
| CEW[282] | 2423 | 2423 | 1192 subjects with both eyes closed, and 1231 subjects with eyes open | No commercial use |
| AR Face[125] | 126 | 4,000+ | Eyes openness labels provided along with CEW dataset | Academic use only |
| CAS-PEAL-R1[128] | 1040 | 99,594 | Eyes openness labels provided along with CEW dataset | Non-commercial use |
| BioID[50] | unknown | 1,521 | Eyes openness labels provided along with CEW dataset | No license |

*Table 38: Summary of datasets focusing on eyes openness state*

## 17.3 Methods and Algorithms

The working draft ISO/IEC WD4 29794-5:2022 depicts the detection of the state of the eyes by a simple algorithm involving landmarks of detected eyes. With those landmarks, the largest distance between the upper and lower eyelids of both eyes DL and DR are measured and the inter-eye distance (IED) is determined (see Section 12). Afterwards the ratio between the smaller eyelid distance and the inter-eye distance is calculated, thereby computing the eye openness aspect $\omega = \frac{\min(D_L, D_R)}{D_{IED}}$.

Early publications on the detection of the closeness of eyes combine hand-crafted feature descriptors for texture and shape with classical machine learning approaches (i.e. no CNNs). For example, in [241], Zhou et al. trained an Adaboost classifier on features extracted by local binary patterns (LBP). On their dataset composed from various other datasets of constrained images (including CAS-PEAL[128], AR Face[51] and BioID[50]), they achieve an accuracy of 99.84%. In [242], Song et al. train a Support Vector Machine (SVM) on feature descriptors obtained from Histograms of Oriented Gradients (HOG), Local Ternary Patterns (LTP), and Gabor wavelets. On the constrained datasets CAS-PEAL, AR and BioID, they achieve an accuracy between 97.14% and 99.75%, and emphasize its superior detection performance for unconstrained images, like those in their proposed new dataset Closed Eyes in the Wild (CEW)[282], where they achieve an accuracy of 94.72%.

More recent methods, typically, do not just focus on eye state detection but consider the more general task of fatique/drowsiness detection (e.g. for car drivers), which can also involve the detection of yawning.

In [243], Redhaei et al. presents a real-time visual-based driver drowsiness detection by detecting the face and then the eye region through facial landmarks. Afterwards, the Eye Aspect Ratio (EAR) is calculated by the following equation

$$EAR = \frac{\|p2 - p6\| + \|(p3 - p5)\|}{2\|(p1 - p4)\|},$$

---

[281] https://www.blinkingmatters.com/research

[282]

http://parnec.nuaa.edu.cn/_upload/tpl/02/db/731/template731/pages/xtan/ClosedEyeDatabases.html

where p1 through p6 indicates the facial landmarks of the driver's eye. A closed state of the eye resembles an EAR of almost 0; an eye is considered to be opened if the EAR is larger than or equals to the threshold 0.2. Figure 50 illustrates an example of the eye with their extracted landmarks in open and closed state (right) and the EAR change over time (left).

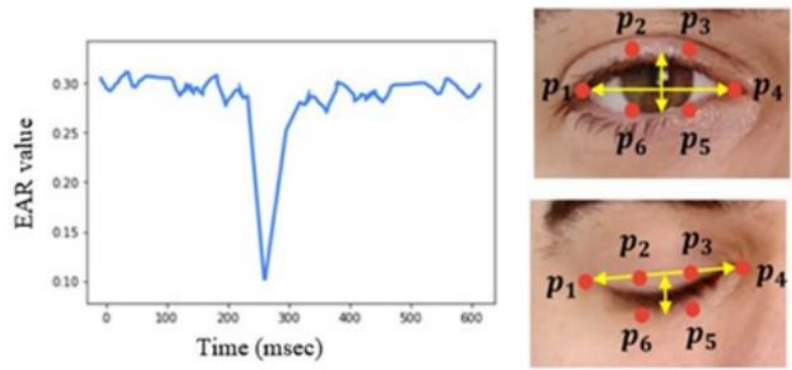

*Figure 50: EAR change over time and examples of an open and closed eye [243]*

The EAR estimation is applied to each frame of a video clip. Different classifiers are used to detect sleepy behaviour. Especially, different classifiers have been investigated to separate eye blinks from eyes closed for a longer period of time. However, sleepiness detection on video clips can generally not be applied for detecting closed/open eyes from a single image. Yet, the approach of EAR estimation can be applied.

Some other methods for sleepiness/drowsiness detection that are applied to video streams (and cannot be applied to single images) were found as well [244], [245]. Furthermore, two other methods have been found were intermediate steps can potentially be used for detecting if a subject's eyes are open or closed.

Dwivedi et al. [246] use a CNN-based approach for driver drowsiness detection by using the MobileNet CNN model [182] and the MRL eyes dataset for transfer learning. The MRL dataset consists of labelled human eye images of 37 different people with and without glasses. The model receives an accuracy of 98.45% on images in detecting whether eyes are open or closed.

Ji et al. [247] use a multi-task cascade CNN to detect the openness state of the eyes and the mouth. On the CEW dataset, they achieve an accuracy of 97.87%.

Another CNN-based approach was proposed by Gandhi et al. in [248] named CONV-LSTM. CONV-LSTM uses a CNN to extract information from images and feed a sequence to the LSTM (Long Short-Term Memory) for prediction. LSTMs are a special kind of *recurrent neural networks* (RNN) capable of learning long-term dependencies while avoiding the vanishing and exploding gradients problem. Gandhi et al. chose the pretrained model Inception-v3283 and then used their own dataset collection for transfer learning and also to train the LSTM. Their dataset consists of 19 videos in total with eight subjects imitating signs of alertness and drowsiness (slow eyelid closure, droopy eyes followed by a quick recovery of head posture to imitate micro-sleep). The model achieves a confidence of 93.65% for detecting a drowsy person (closed eyes) and 99.63% of detecting an alert person (open eyes). Even though, the report refers to the evaluation on video clips, the CNN can potentially be used for being applied to single images.

## 17.4 Software

Two kinds of implementations were found, one based on algorithmic approaches and the other based on deep convolutional neural networks (DNN). The two kinds of implementations are listed in separate tables below.

The following table summarizes implementations found on GitHub that consider eye state recognition through algorithmic (non-DNN) approaches. It is important to note that many algorithmic implementations are based on the same approach of eye aspect ratio calculation and a fixed threshold. These implementations generally vary only in the use of different thresholds and therefore only one representant is listed in the following table.

---

283 https://en.wikipedia.org/wiki/Inceptionv3

| Name | Method | Remarks | License |
|------|--------|---------|---------|
| Fatigue-Detection[284] | **Method 1:** Facial landmark detection (dlib), aspect ratio of main axes eyes **Method 2:** Skin colour thresholding and count of black pixels in eyes area | - | No license |
| Driver-Fatigue-Detection[285] | Aspect ratio of main axes | Representant for multiple eye aspect ratio-based implementation | No license |
| Driver-Sleep-Detection-System[286] | Pupil detection through Sobel operator, considered open if detectable | Implemented in MATLAB | No license |
| sleepinessDetection[287] | OpenCV's eye cascade classifier to detect eyes. If not, eyes are assumed to be closed. | | No license |

*Table 39: Summary of available implementations regarding eye state detection (algorithmic approach)*

The following table summarizes implementations found on GitHub for CNN-based eye state recognition.

| Name | Dataset | Model size | Metrics | License |
|------|---------|-----------|---------|---------|
| Eye-State-Detection[288] | Own dataset | 5 MB | Accuracy: 0.97 | No license |
| Eye Closure State Detection using CNN[289] | Own dataset | 0.06 MB | Accuracy: 0.934 | No license |

*Table 40: Summary of available implementations regarding eye state detection (CNN approach)*

---

# 18    No Reflections on Eyeglasses

Facial images of persons wearing eyeglasses can show reflections on glasses. Annex D.2.3.6 of ISO/IEC 39794-5:2019 requires that there are no lighting artefacts or flash reflections on glasses that cover regions of the eyes. However, such reflections not only result from lighting but also from natural illumination. Figure 23 shows an example of reflections on glasses.

## 18.1    Impact on Face Recognition

No publications could be found, reporting on evaluations of the impact of reflections of eyeglasses on the recognition accuracy of face recognition or other biometric algorithms. However, since reflections partly or completely obscure the texture of the face at this position, and the eye region is of particular importance for face recognition, it is quite plausible to assume that strong reflections have a negative impact on face recognition accuracy.

## 18.2    Datasets

No datasets have been published that contain labels for reflections on eyeglasses.

## 18.3    Methods and Algorithms

No publications could be found that focus on the detection of reflections of eyeglasses in facial images. However, by using a method for the removal of reflections, either on glasses in particular, or in images in general, it is possible to detect reflections by comparing the original input image with the output image where reflections have been removed. In order to perform a quantitative assessment of the extent to which the image is impaired by the reflections, the pixel intensities in the difference image can be aggregated into a single numeric value, e.g. by a weighted sum of their magnitude, where higher weights are used in the location of the eyes.

There are a few approaches that are specialized on the removal of reflections on eyeglasses. In [249], Sandhan and Choi propose anti-glare, an approach that exploits the difference of the image gradient distributions and the symmetry of faces, to separate a reflection layer from the background. The results look quite promising (see Figure 51) but the processing time of their implementation in Matlab is very high (> 10s per image).

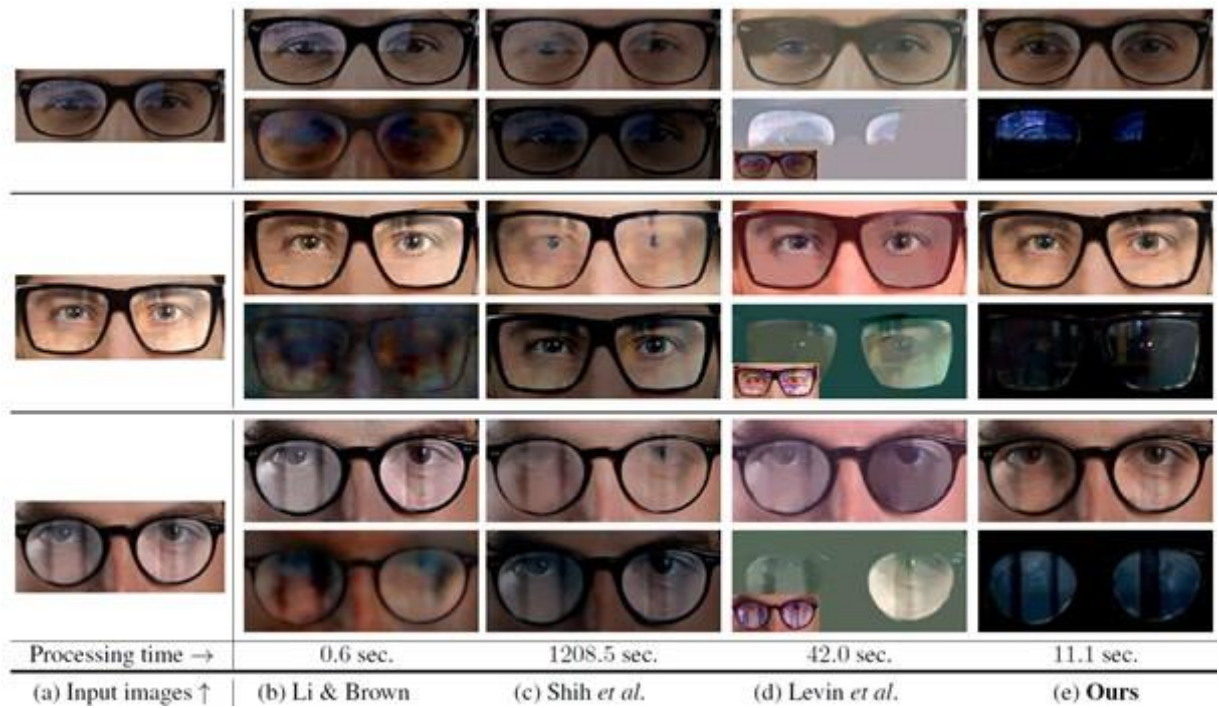

| Processing time → | 0.6 sec. | 1208.5 sec. | 42.0 sec. | 11.1 sec. |
| (a) Input images ↑ | (b) Li & Brown | (c) Shih *et al.* | (d) Levin *et al.* | (e) **Ours** |

*Figure 51: Results of the eyeglasses reflection removal of [249] (rightmost column) with previous methods. In each row, the lower image shows the extracted reflection layer*

In [250], Watenabe and Hasegawa propose two methods to remove reflections on eyeglasses, one based on an auto-encoder and the other on U-Net (a CNN architecture commonly used for image-to-image-translations). The example results presented by the authors show better results for U-Net, see Figure 52.

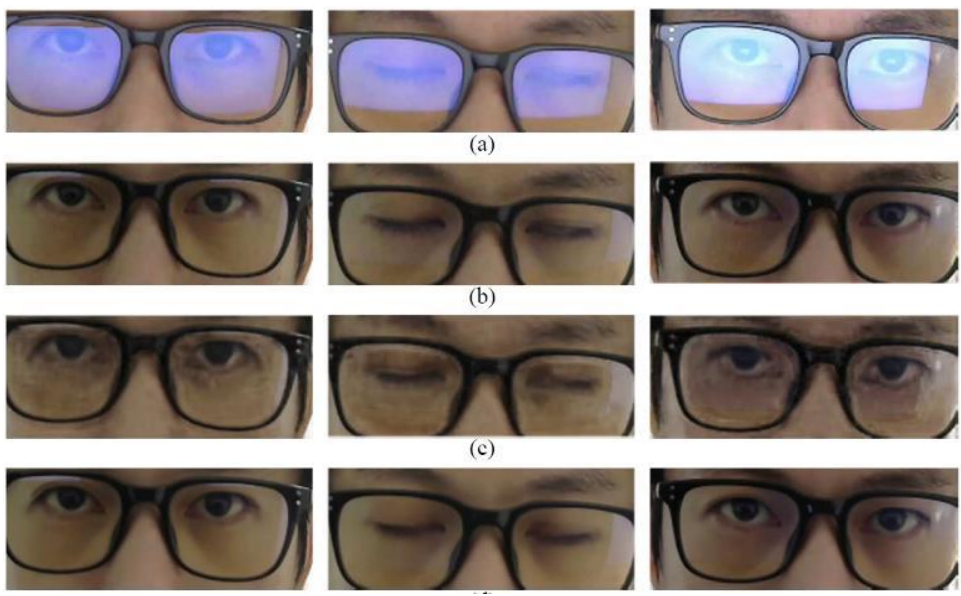

*Figure 52: Example results from [250], showing (a) the input images, (b) real images captured without reflection, (c) the output of the auto-encoder and (d) the outputs of the U-Net*

| Input | Zhang et al. [253] | Wei et al. [252] | Li et al. [254] | Ground Truth |
|---|---|---|---|---|

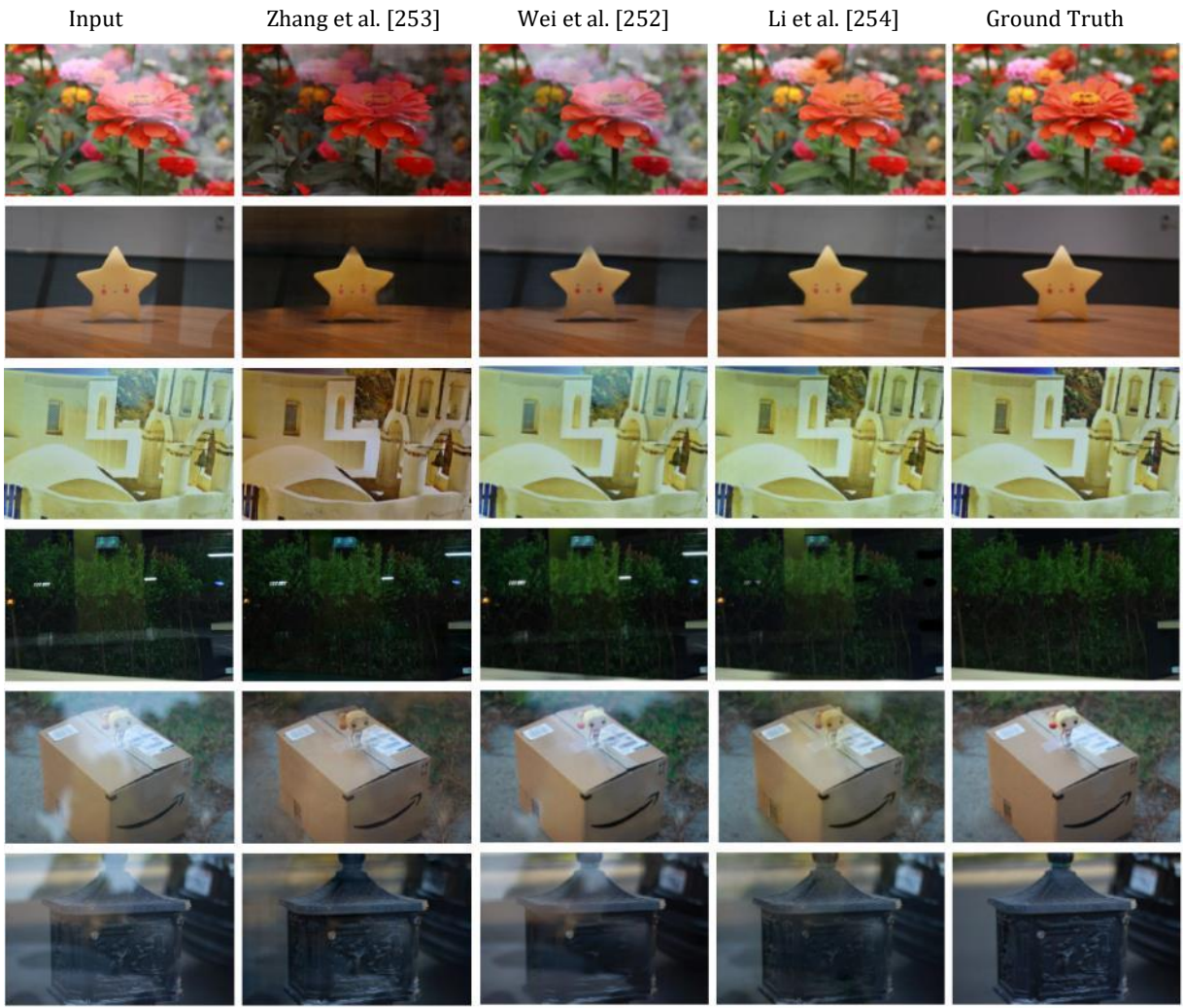

*Figure 53: Qualitative comparison (from [254]) of the methods of [253], [252] and [254]*

There are many more papers on the removal of general reflections (on glass surfaces) in images. Provided that these approaches do not require a second image taken with a different illumination as input (as it is the case in [251], for instance), they could also be applied to detect reflections on glasses in facial images. These algorithms aim to remove not just simple reflections from point-shaped or planar illumination sources but also complex reflections resulting in "ghost images" of a reflected scenery. Most of these approaches are based on deep learning, e.g. Zhang et al. [252], Wei et al. (ERRNet) [253] or Li et al. (IBCLN) [254].

In [254], the authors perform a quantitative comparison of their method with previous approaches, including [253] and [252], using two different measures for the similarity between the output image and the ground truth image (image variant taken without reflections):

- The *peak signal-to-noise ratio* (PSNR) which measures the similarity on a pixel level.
- The *structural similarity* (SSIM) which measures the differences in the visible structures.

In both measures, the method of [254] gave the best results. The results are listed in the following table.

| Measure | Method from [253] | Method from [252] | Method from [254] |
|---|---|---|---|
| PSNR | 23.45 | 20.85 | 24.08 |
| SSIM | 0.870 | 0.829 | 0.875 |

*Table 41: Quantitative comparison of the methods of [151], [149] and [152], taken from [152]. The reported measures are in higher-is-better semantics.*

However, it is important to note that the reported measures reported are unrelated to biometric performance and, hence, of limited relevance for the selection and evaluation of facial image quality assessment algorithms.

## 18.4   Software

The following table provides information about repositories found on GitHub which implement general reflection removal for images. No implementations have been found that specialize on reflections on glasses.

| Name | Model size | License |
|------|------------|---------|
| ERRNet[290] [253] | 330 MB | No license |
| IBCLN[291] [254] | 77 MB | BSD 2-Clause "Simplified" |
| perceptual-reflection-removal[292] [252] | 350 MB | Apache 2.0 |

*Table 42: Implementations on reflection removal in images (not specialized for faces)*

---

[290] https://github.com/Vandermode/ERRNet
[291] https://github.com/JHL-HUST/IBCLN
[292] https://github.com/ceciliavision/perceptual-reflection-removal

# 19    No Other Faces in the Background

The following section summarizes the requirements related to assisting persons and faces of other persons in face images. Other persons standing in the background are captured in a face image could often be the case in System border enrolment scenarios. The ISO/IEC 19794-5:2011 contains in its subclause 7.2.4 "Assistance in positioning" a requirement that in no cases any other face shall be captured in the image, the Annex B of the same document gives best practice recommendations and provides in the subclause B.2.2 "Assistance in positioning the face" a recommendation that hands, arms etc. of an assisting person used to support the positioning should not be visible.

The subclause D.1.4.2.5 "Background" of the ISO/IEC 39794-5:2019 requires that not any objects like supporting persons etc. shall be visible in the background and additionally in the subclause D.1.4.5.2 "Children below one year" is required that "hands, arms and other body parts of an assisting person used to support the positioning of the subject, e.g. parents supporting their child, shall not be visible in the image".

Beside the case where a person assists a child who is being photographed, there could also be the case where a person who is being photographed is holding a child. In both cases the above stated requirements shall be met, i.e. there shall be no other faces in the background.

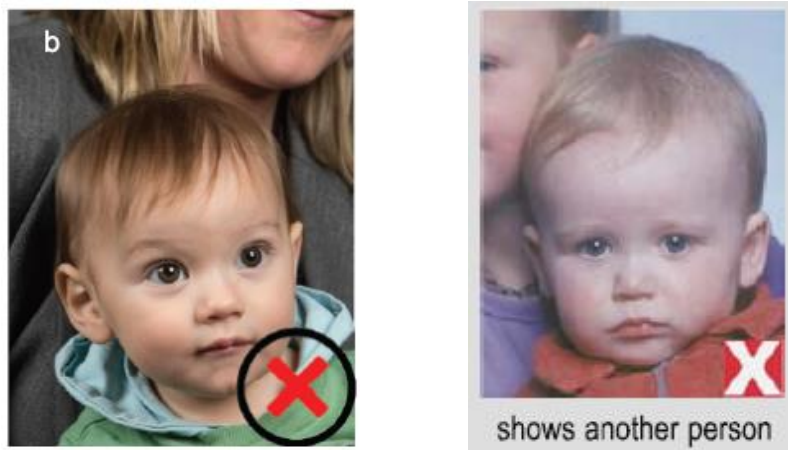

*Figure 54: Examples of images showing another person in the background, taken from ISO/IEC 39794-5:2019 (left) and ISO/IEC 19795-4:2011 (right)*

## 19.1   Methods and Algorithms

To check if any other faces are visible in the image one could use methods of face detection described in Section 2.1.

To analyse whether any objects like hands, arms or other body parts of assisting person are visible in the image the methods and algorithms proposed for background uniformity analysis could be used (see Section 4.3). It is actually not necessary to recognize what exact body parts of assisting person are visible, it would be sufficient to recognize that the background isn't uniform and this can be achieved with background uniformity analysis methods.

# 20    No Head Coverings

ISO/IEC 19794-5:2001 recommends that head coverings (e.g. hats, caps, headscarves) are absent, except in cases, where a covering cannot be removed, e.g. due to religious reasons. Similarly, ISO/IEC 39794-5:2019 requires that head coverings shall not be accepted except in circumstances specifically approved by the issuing state.

## 20.1    Impact on Face Recognition

There are no investigations on the influence of head coverings on face recognition.[293] Clearly, this impact depends on which parts of the head and face are covered by the head covering. If the head covering does only cover (parts of) the hair, e.g. parts of the face that would otherwise be covered by hair, it can hardly have any effect on face recognition. However, if the head covering occludes parts of essential face parts (e.g. eyebrows, cheek), it can result in a degradation of the recognition performance. The same holds, if a head covering casts dark shadows on parts of the face, so that the texture in these areas are not clearly visible.

## 20.2    Datasets

The following table summarizes the most eligible datasets containing face images and ground truth annotations on whether the subject is wearing a head covering.

| Name | Subjects | Images | Remark | Constrained? | License |
|------|----------|--------|--------|--------------|---------|
| UMB-DB[294] | 143 | 1473 | 3D scans and 2D colour images with labels for hats and scarves | Yes | Royalty-free license, for research only |
| CelebAMask-HQ[35] | unknown | 30.000 | Contains annotations for "hat" | No | Non-commercial use |

*Table 43: Summary of available datasets with labels for head coverings.*

## 20.3    Methods and Algorithms

There exist several publications proposing algorithms for detecting hard hats in images or video sequences, e.g. [255], [256], [257]. However, hard hats are quite unlikely to be worn in enrolment or border control scenarios. The face segmentation (face parsing) algorithm proposed in [34] also detects and segments hats or similar head coverings.

## 20.4    Software

The following face segmentation (also called *face parsing*, see Section 2.2.3) implementations also output the image area where a hat is visible, and can thus be used for the detection of head coverings. It needs to be verified, if all relevant kinds of head coverings (e.g. head scarves) are detected.

| Name | Model Size | License |
|------|-----------|---------|
| CelebAMask-HQ Face Parsing[41] | 7.5 MB | Non-commercial |
| face-parsing.PyTorch[42] | 51 MB | MIT |
| FaceParsing.PyTorch[43] | 96 MB | MIT |

*Figure 55: Software implementations of face segmentation algorithms that distinguish hats.*

---

[293] There are, however, studies on the impact of headscarves on face recognition by human observers, e,g, [235].

[294] http://www.ivl.disco.unimib.it/minisites/umbdb/description.html

# 21 Commercial Algorithms

Some commercial face recognition software products also include functions for assessing the quality of facial images. In this section, only those algorithms are described, for which detailed information is available.

There exist additional commercial algorithms that have been submitted to NIST FRVT QA (where, currently, only a quality scalar output is tested), but, for these, no further details were found:

| Commercial algorithm: |
| --- |
| China Electronics Import-Export Corp. |
| Guangzhou Pixel Solutions Co Ltd. |
| Rank One Computing |
| Tevian |

*Table 44: Overview of additional commercial algorithms submitted to NIST FRVT QA*

## 21.1 Cognitec Face VACS

The Face VACS SDK of Cognitec provides functions for face recognition, including quality assessment. The current version is 9.7.0, but the following information is based on Version 9.6.0, which is the latest version available to secunet Security Networks.

The SDK implements a class FRsdk::Portrait::Analyzer, by which the quality of a facial image according to ISO/IEC 19794-5:2005[295] can be estimated. Besides general information on the image (e.g. dimensions or colour space), the face (e.g. the position of the eyes and the face), and the subject (age, sex, ethnicity estimation), the output contains information on the following quality aspects:

| Quality aspect | Description |
| --- | --- |
| Face dimensions | Size of the face (width, height, distances between different feature points defined in ISO/IEC 19794-5:2005 section 5.6.3.) |
| Face mask | The confidence that the person is wearing a face mask |
| Eyeglasses | The confidence that the subject is wearing eyeglasses |
| Eyes open | The confidences that the eyes are open (for each eye separately) |
| Frontal gaze | The confidences that the person looks into the camera (for each eye separately) |
| Red eyes | A measure for the redness of the eyes (for each eye separately) |

---

[295] Even though the current version is ISO/IEC 19794-5:2011, the manual of the SDK refers exclusively to the previous version from 2005.

| Quality aspect | Description |
|---|---|
| Visibility of eyes regions | The amounts to which the areas surrounding the eyes are tinted (for each eye separately) |
| Exposure | The exposure measured by the average grey value within the facial region |
| Grey scale density | The number of different grey values within the facial region |
| Natural skin colour | A measure for how natural the skin colour is |
| Hot spots on skin | A measure for hot spots visible on the face |
| Background uniformity | A measure for the uniformity of the background |
| Pose | How frontal the pose is (joint measure for yaw and pitch) |
| Lighting uniformity | A measure for left-right-uniformity of the lighting of the face |
| Sharpness | A measure for focus and depth of field according to ISO/IEC 19794-5:2005, Section 7.3.3 |
| Closed mouth | If the mouth is closed |

*Table 45: Overview of VACS SDK Quality Aspects*

Except for the aspects measuring a size or distance, the returned values are between 0 and 1.

## 21.2   Dermalog Face SDK

The Face SDK of Dermalog provides functions for face recognition including quality assessment. The current version is 5.15.1.

The SDK provides various functions for quality assessment, which compute measures for the following aspects:

- **Uniform quality score:** The function FRCheckFaceQuality outputs a score for the overall quality between 0 and 100.
- **Neutral expression:** The function FXEstimateSmile estimates, if the person is smiling.
- **Pose:** The function FDFindFacePose estimates the pose angles pitch, yaw and roll.

In addition, the SDK contains a component Dermalog.Face.ICAOCheck, which estimates the compliance of the facial image to some of the requirements defined in ISO/IEC 19794-5. The output class contains the following elements:

| GeometricConstraintsPassportOk | GeometricConstraintsFullFrontalOk |
|---|---|
| IsPortraitImageInterpolated | FaceDetected |
| PoseOk | NonOccluded |
| SingleFace | BackgroundOk |
| FaceShadowOk | FocusOk |
| MouthClosed | NoHotspots |
| ResolutionOk | ExpressionNeutral |
| EyesOpen | ColourOk |

| GeometricConstraintsPassportOk | GeometricConstraintsFullFrontalOk |
|---|---|
| NoReflection | NoGlasses |
| GazeFrontal | ExposureOk |
| NoRedEyes | Yaw |
| Pitch | Rotation |

*Table 46: Overview of the quality aspects calculated by Dermalog Face SDK*

No details on the computation and exact meaning of these values are given in the user manual.

## 21.3   Neurotechnology Verilook SDK

The Verilook SDK of Neurotechnology provides functions for face recognition, including quality assessment. The current version is 12.3, but the following information is based on Version 12.1, which is the latest version available to secunet.

The class NLAttributes Class contains functions to compute various subject attributes (e.g. age, sex, ethnicity, emotion, beard, moustache, glasses), to assess the liveness of the subject, and to measure the following quality aspects.

| Quality aspect | Function name |
|---|---|
| Uniformity of Background | BackgroundUniformity |
| Eyes open | BlinkConfidence |
| Contrast and colour balance | Contrast, Saturation |
| Occlusion of eyes by sunglasses | DarkGlassesConfidence |
| Neutral Expression | Expression |
| Illumination | FaceDarknessConfidence, SkinReflectionConfidence |
| Face masks | FaceMaskConfidence |
| Greyscale Density | GrayscaleDensity |
| Eyes looking into the camera | LookingAwayConfidence |
| Mouth closed | MouthOpenConfidence |
| Noise | Noise |
| Pose | Pitch, Yaw, Roll |
| Resolution | PixelationConfidence |
| No red eyes | RedEyeConfidence |
| Sharpness | Sharpness |
| Natural skin colours | UnnaturalSkinToneConfidence, WashedOutConfidence |
| No head coverings | HatConfidence |

*Table 47: Verilook SDK quality aspects and corresponding function names*

Furthermore, the struct Neurotec.Biometrics.NBiometricStatus computed by the method NBiometricClient.Capture contains the following face image quality aspects.

Table 48: Entries of struct NBiometricStatus relating to quality aspects of facial images

| Quality aspect | Entry name |
|---|---|
| Unified Quality Score (threshold applied) | BadObject |
| No other faces in the background | TooManyObjects |
| Illumination | BadDynamicRange |
| | BadExposure |
| | BadContrast |
| | BadLighting |
| Sharpness | BadSharpness |
| No occlusion of the face | Occlusion |
| Pose | BadPose |
| Location and coverage of the head | BadPosition |
| | TooClose |
| | TooFar |

In addition, the function ANQualityMetric.Score computes an overall quality score.

## 21.4   Paravision Face Recognition SDKs

Paravision offers a variety of Face Recognition SDKs for different platforms providing functions for face recognition including QA assessment.  A documentation of the SDKs is publicly available.[296] The latest version 8.0 of the server SDK, which implements the generation 5 of the face recognition modules, contains the following functions to estimate the quality of the facial image:

- The function GetQualities estimates a uniform quality score of the facial image.
- The function GetMasks outputs a confidence that the subject is wearing a face mask.

## 21.5   eu-LISA USK6

The User Software Kit (USK) for EES Face Quality Assessment provided by eu-LISA can output warnings for the following quality aspects:

| Quality warning |
|---|
| Subject is wearing dark glasses |
| Subject is wearing a face mask |
| Pose (pitch/yaw/roll angles) |
| Mouth wide open |
| Inter-eye distance |
| Insufficient space around eyes |

Table 49: Overview over USK6 Warnings for the corresponding quality aspect

In addition, USK6 outputs a unified quality score computed by a CNN.

---

[296] https://docs.paravision.ai/paravision-navigation-v2/jxHZwah1vaWwCxiJJQL7/

# 22   Glossary

| Term | Description |
|---|---|
| Anchor Boxes | Commonly used technique in CNNs for object detection. Anchor boxes are pre-defined bounding boxes at various locations and scales marking potential locations of objects. For each anchor box, an object classifier is applied to the features restricted to the respective area. Furthermore, the location is fine-tuned by predicting the offset of the object's real bounding box to the anchor box. |
| Convolutional Neural Network (CNN) | Neural Network in which convolutional layers are used (typically the first layers). In convolutional layers, fixed-sized filters are shifted over the layer's inputs values to generate one output at each position. The outputs of the convolutional layers can be considered as features used for the classification or regression task performed by the following layers; thus, in a CNN, the features are not had-crafted but determined during training. |
| Face Detection | The task of finding and localising faces in images. |
| Facial Landmark | Key point specifying the location of a certain face part, e.g. a corner of the eye. |
| Face Segmentation | The task of separating the parts in the image where the face and, optionally, other parts of the subject are visible. |
| Feature Pyramid Networks (FPN) | Class of CNNs for object detection in images. FPNs feed features from several consecutive convolutional layers to the final layers (via lateral layer connections). Since the scale of the features extracted increases with each convolutional layer, this approach allows to detect objects at different scales using a common network (the final layers). |
| Regional CNNs (R-CNN) | Class of CNNs for object detection in images. In a first step, R-CNNs apply a so-called *selective search* to identify potential locations (*region proposals*) of objects in the image; for example, pre-defined anchor boxes can be used. For each region proposal, a base CNN is used to extract features, on which an *object classifier* and a *bounding box regression* is applied. The object classifier and the bounding box regression do not need to be implemented by a CNN but can also deploy classical machine learning algorithms.

An improved approach is the *Fast R-CNN*, where the features are computed only once for the entire image (using a base CNN) and, for each region proposal, the corresponding feature subsets are extracted from the image's features using a *region of interest (RoI) pooling layer*. Furthermore, before feeding the feature subsets corresponding to the region proposals to the object classifier and bounding box regression they are fed through a fully connected layer.

A further improvement of Fast R-CNN is *Faster R-CNN*, where the selective search is performed by a region proposal network, which takes as input the features extracted from the image by the base CNN and uses anchor boxes to compute region proposals. |

| Term | Description |
|------|-------------|
| Single Shot Multibox Detection (SSD) | Class of CNNs for object detection in images. An SSD consists of a *base CNN* extracting features from the input image followed by several consecutive *multiscale feature map blocks*, each of which reducing (e.g. by half) the dimensions of the feature maps from the previous block, allowing the SDD to detect objects at different scales. For each multiscale feature map block, the output features are restricted to pre-defined anchor boxes and then fed to an *object classification network* and a *bounding box prediction network*. 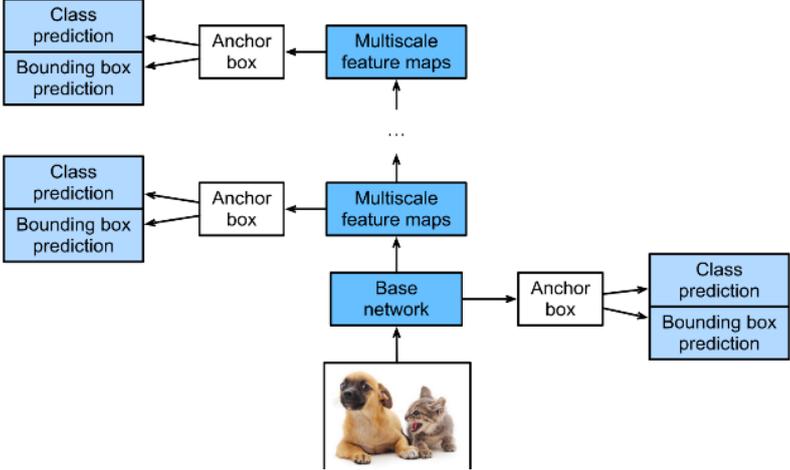 *Figure 56: General architecture of an SSD (image taken from https://d2l.ai/chapter_computer-vision/ssd.html)* |
| YOLO | Family of state of the art object detection networks ranging from YOLO1 to YOLOv5. YOLO stands for You Only Look Once. |